# Toolshed: Scale Tool-Equipped Agents with Advanced RAG-Tool Fusion and Tool Knowledge Bases

**Elias Lumer*, Vamse Kumar Subbiah, James A. Burke,**
**Pradeep Honaganahalli Basavaraju, Austin Huber**
*Innovation Hub, PricewaterhouseCoopers*

**ABSTRACT**

Recent advancements in tool-equipped Agents (LLMs) have enabled complex tasks like secure database interactions and multi-agent code development. However, scaling tool capacity beyond agent reasoning or model limits remains a challenge. In this paper, we address these challenges by introducing *Toolshed Knowledge Bases*, a tool knowledge base (vector database) designed to store enhanced tool representations and optimize tool selection for large-scale tool-equipped Agents. Additionally, we propose *Advanced RAG-Tool Fusion*, a novel ensemble of tool-applied advanced retrieval-augmented generation (RAG) techniques across the pre-retrieval, intra-retrieval, and post-retrieval phases, without requiring model fine-tuning. During pre-retrieval, tool documents are enhanced with key information and stored in the *Toolshed Knowledge Base*. Intra-retrieval focuses on query planning and transformation to increase retrieval accuracy. Post-retrieval refines the retrieved tool documents and enables self-reflection. Furthermore, by varying both the total number of tools (*tool-M*) an Agent has access to and the tool selection threshold (*top-k*), we address trade-offs between retrieval accuracy, agent performance, and token cost. Our approach achieves 46%, 56%, and 47% absolute improvements on the ToolE single-tool, ToolE multi-tool and Seal-Tools benchmark datasets, respectively (Recall@5).

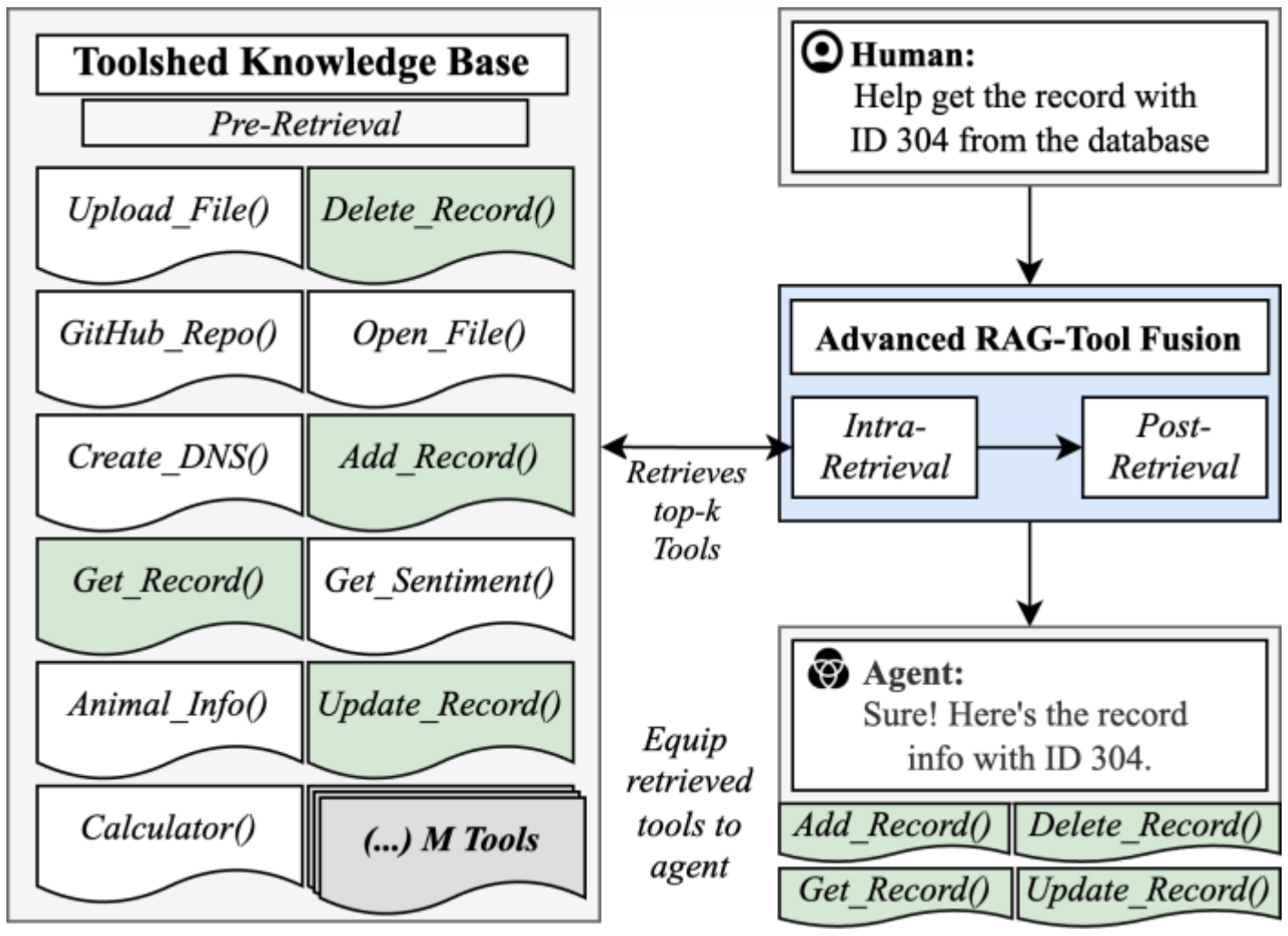

**Figure 1.** Toolshed Knowledge Base, optimized for selection threshold *top-k* and total tools *tool-M*, utilizing Advanced RAG-Tool Fusion, an ensemble of pre-retrieval, intra-retrieval, and post-retrieval retrieval-augmented-generation techniques applied to tool selection/retrieval, and planning.

## 1 Introduction

The latest advancements in Large Language Models (LLMs) have enabled LLM Agents to autonomously handle tasks and reliably interact with external functionalities and APIs. The addition of tool calling, also known as function calling, has allowed complicated user requests to be possible, shining light to new use-cases for Agents–such as 1) interacting with a secure domain-specific database, 2) multi-agent teams developing code features and pushing to GitHub, and 3) performing advanced QA tasks over domain-specific knowledge management articles. Moreover, tool calling permits reliable JSON outputs for pipelines [1], such as having an Agent extract information from unstructured text that conforms to a schema–which is a significant improvement on the previous method which prompted the LLM to produce a specific output and then parse the text response using regex [2]. In production-grade applications, tool calling not only enables Agents to take action beyond an unstructured response but ensures the agent output will follow the same schema on every response. Model providers like OpenAI, Anthropic, and Meta have fine-tuned their models to automatically determine when and how to invoke functions and generate the appropriate JSON schema based solely on the function definition provided in the API request [3]. This eliminates the need for developers to manually define tool parameters in prompts, allowing them to concentrate on scalability, tool selection (also referred to as tool retrieval), and agent orchestration. Both OpenAI (gpt-4o-2024-08-06), Anthropic (claude-3-5-sonnet-20240620), Google Gemini 1.5 Pro (gemini-1.5-pro-002) support up to 128 tool function definitions in the API request, though OpenAI recommends limiting it to around 20 tools per agent for complex user interactions to maintain accuracy [3], [4]. To overcome this limitation, multi-agent systems such as Autogen, Langgraph, and CrewAI have been introduced, allowing multiple Agents equipped with specialized toolsets to collaborate and distribute tasks effectively.

* elias.lumer@pwc.com

However, with multi-agent systems, each individual sub-agent is still constrained by the 128 tools, meaning that scaling beyond this limit—such as equipping a single agent with 1,000 specialized database operation tools—requires alternative approaches. Two primary methods have been proposed to address this: 1) tuning-based methods for tool calling and selection, where the LLM is fine-tuned for tool usage, and 2) retriever-based tool selection and planning, which is the focus of this paper [5]. Qin et al. (ToolLlama) [6] and Patil et al. (GorillaLLM) [7] are prominent tuning-based approaches where open-source LLMs are fine-tuned for tool calling to compete with accuracies of closed-source models. Huang et al. (PLUTO) [8], Yuan et al. (CRAFT) [9], Chen et al. (Re-Invoke) [10], Qu et al. (COLT) [11], and Anantha, et al., 2023 (ProTIP) [12] present prominent retriever-based (dense retrieval, if referring to SOTA non-term-based methods such as BM25 [13]) approaches that utilize neural networks to learn the semantic relationship between queries and tool descriptions to retrieve the most relevant tools from a vector database or tree-based architecture and fed to downstream tasks or directly in the prompt.

Despite the advancements in retriever-based tool selection systems, there is still a significant gap in approaches compared to that of the advanced retrieval-augmented generation (RAG) community [14]. One primary concern is that current state-of-the-art tool retrievers only rely on 1-2 key tool information to embed as vector representations of the tool: the tool name and tool description. As explained by Gao et al. [14], Advanced retrieval-augmented generation (RAG) methods have made significant progress in enhancing the quality of document chunks, such as appending document summaries, hypothetical questions, key topics, and key metadata to each chunk-level document. Additionally, there are inference-time solutions to increase the likelihood of gathering all the relevant documents based on a query: query decomposition and planning [15], [16], query rewriting [17], step-back prompting [18], document reranking and combining multiple knowledge sources with reranking as RAG-fusion [19], self-grading documents for relevancy [20], and adaptive routing of single vs. multi-step question-answering tasks [21]. Furthermore, there are configuration-level advancements to optimize parameters within a RAG system, such as the (document) selection threshold *top-k*, embedder type, LLM type, and knowledge source type [14]. There is limited research within the tool retrieval space on optimizing these configuration values such as *top-k*–the number of retrieved tools to equip to a single agent–which is further exacerbated by the lack of understanding in how an Agent will perform at any given number of *M* tools attached to it (since *M* will equal our *top-k* value).

The second primary concern is limited research on an Agents' performance (accuracy, token cost) equipped with any number of *M* tools (commonly referred as a Base LLM Agent, or Simple Agent attached to *M* available tools). As an example of a specialized agent with 100 static database operation tools (*tool-M* = 100), should we pass in all 100 useful tools to the API tools parameter (OpenAI, Anthropic) to answer a user question that only needs 1-4 tools to complete any given request? While current research examines much larger *tool-M* values, such as Qin et al.'s ToolBench's 16,000+ APIs [6], there is little research on the impact of *tool-M* values on a granular level. Furthermore, the same granular impact of the selection threshold for the tool retriever, *top-k*, is a primary concern which is tightly coupled to the *tool-M* value. Understanding how the *top-k* parameter impacts the retrieval system is crucial because it narrows the scope of tools an agent focuses on any given request, and without knowing how your agent performs with a given number of *M* tools (*tool-M*) independently, you cannot predict its performance when scaling with *top-k* (*tool-M* becomes *top-k*). And further, can we take advantage of cost benefits of limiting the number of tools an agent is equipped with, within the *tool-M* 1-128 range? These gaps highlight the critical scope of this paper and for future research in the development in LLM Agents tool selection, retrieval, and planning.

In this paper, we introduce *Toolshed Knowledge Bases*, a tool knowledge base designed to optimize the storage and retrieval of tools in a vector database for large-scale tool-equipped Agents. A *Toolshed Knowledge Base* is equipped to an agent and stores the enhanced tool document made up of the tool name, description and appended key information and tool-specific metadata (Fig. 1). *Toolshed Knowledge Bases* also address the optimization of the tool selection threshold (*top-k*) needed to achieve state-of-the-art accuracy while not adding more tools than necessary to drive up token costs and latency.

We also introduce *Advanced RAG-Tool Fusion*, a novel ensemble of advanced RAG patterns, modularly applied to the tool selection/retrieval and planning field. These patterns consist of 1) pre-retrieval techniques which directly utilize to the *Toolshed Knowledge Base* for an agent, 2) intra-retrieval techniques that transform and optimize the user query to retrieve a subset of tools, and 3) post-retrieval techniques that finalize the list of *top-k* tools to equip to the agent by various reranking and corrective strategies (Fig. 1). Pre-retrieval strategies optimize the vector embeddings of the tool documents to increase the likelihood that the correct tool will be retrieved for a given user query and correct tool pair. Intra-retrieval approaches aim to plan, expand, and transform the user query to effectively retrieve the correct tool(s), considering the different ways users phrase their questions and the diverse tools that may be relevant. Post-retrieval strategies refine and condense retrieved tools through reranking and self-reflection, using programmatic, embedding-based, and LLM-driven approaches for optimal retrieval. For this paper, we are solely interested in out-of-the-box embedders with no fine-tuning necessary–such as embedders from open/closed-source providers (ex. OpenAI, Mistral, etc.). Our ensemble-based Advanced RAG-Tool Fusion approach significantly advances tool retrieval, selection, and planning, achieving 46%, 56%, and 47% absolute improvements over BM25 on the ToolE single-tool, ToolE multi-tool, and Seal-Tools benchmarks, while also outperforming current state-of-the-art retrievers (Recall@5).

## 2 Related Works

### 2.1 Advanced RAG

Advanced Retrieval-Augmented Generation (RAG) builds upon naive RAG by incorporating more sophisticated retrieval and generation techniques, designed to improve the relevance, efficiency, and contextual depth of responses. The retrieval problem boils down to what is commonly known as the "needle-in-a-haystack" challenge, where a retriever must select the correct document from a list or database containing hundreds or thousands of documents for the LLM to reason through and provide the correct answer [22]. While naive RAG systems simply combine retrieval and generation in a linear process, advanced RAG introduces several key patterns that enhance its capabilities. Query rewriting, as shown by Ma et al. [17], refine user queries for better alignment with retrieval models, while Gao et al. [23] improve retrieval by generating hypothetical documents embeddings (HyDE) that enhance query understanding, even in out-of-domain cases. Query expansion, as shown by Jagerman et al. [24], Wang et al. (Query2doc) [25], and Peng et al. [26], further broadens queries by adding relevant terms or creating multiple diverse variations of the user query, improving retrieval accuracy. Setty et al. [27] enhance document chunks by appending key metadata about a document, summaries, and hypothetical questions (commonly known as reverse HyDe)–which optimizes relevance and brings the document closer to the query in semantic space [14]. Tang and Yang (MultiHop-RAG) [28], Trivedi et al. (IRCoT) [29], Yao et al. (ReAct) [30], Khattab et al. (DSP) [31], and Joshi et al. (REAPER) [16], and Xu et al. (ReWoo) [15] explore query decomposition and planning, which is focused on breaking down complex user questions into a series of logical retrieval steps or sub-questions for a significant boost in retrieval accuracy for questions with multi-reasoning traces. Step-back prompting [18] enables models to “take a step back” and abstracts key concepts before reasoning through detailed reasoning. RAG Fusion combines multiple sets of documents retrieved from different variations of the original query or the same query with different retrieval methods (blended or hybrid RAG, combining vector search with semantic search) [32], merging them using reciprocal rank fusion (RRF) to enhance the relevance and accuracy of the final output and reranking ensures relevance by reordering results [19]. Reranking algorithms reorders initially retrieved documents, using either typically cross-encoder embedding-based [33] or LLM-based methods [34], to prioritize the most contextually relevant results before passing them to the generation model [33]. Yan et al. (CRAG) [20] and Asai et al. (Self-RAG) [35] assesses the quality of retrieved documents and discards irrelevant ones with an option of retrieving additional documents through critique and self-reflection. Agentic RAG, similar to Self-RAG, equips an agent with a retriever tool to allow it the autonomy to rewrite the query and retrieve documents, answer without a tool from its previous knowledge, query specific knowledge bases, or query the same knowledge base while inputting metadata filtering [36]. Finally, Jeong et al. (Adaptive-RAG) [21] dynamically chooses the appropriate retrieval strategy (single-retrieval, multi-retrieval, etc.) based on query complexity. Our approach, *Advanced RAG-Tool Fusion*, instead of focusing on a document-based knowledge base for Q&A, employs an ensemble of the aforementioned techniques, creatively applied to the selection, retrieval, and planning of tool-equipped Agents within a tool knowledge base (*Toolshed Knowledge Base*).

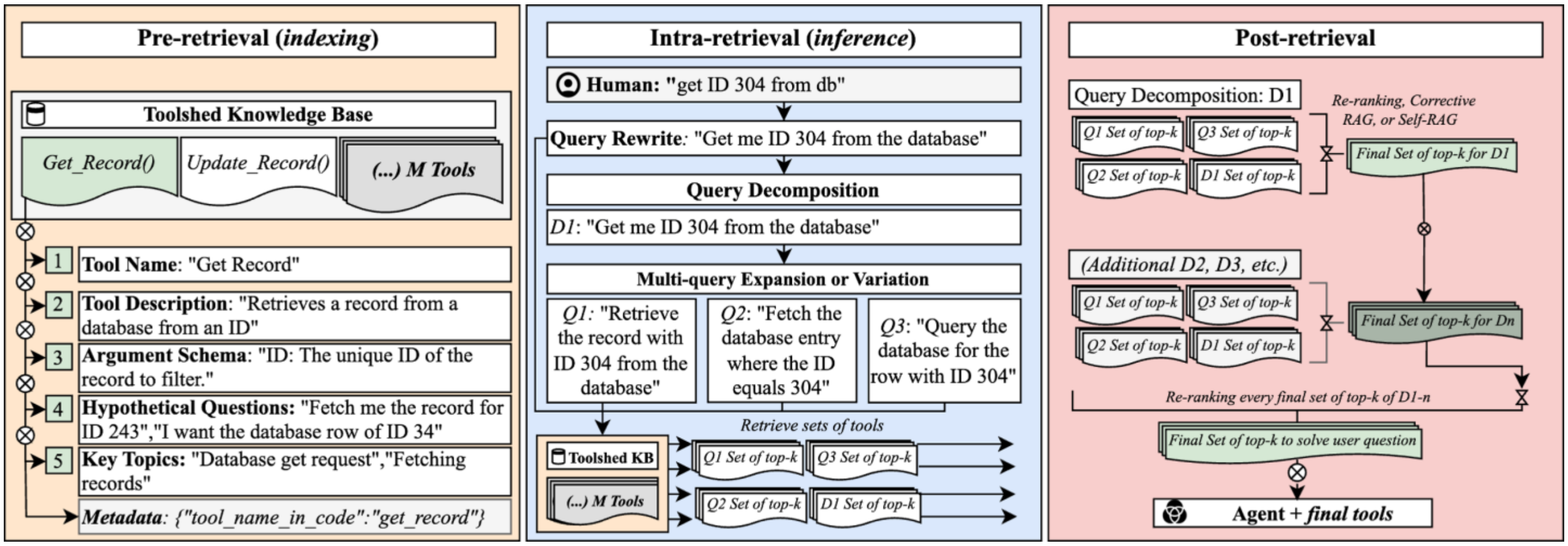

**Figure 2.** Advanced RAG-Tool Fusion within three phases. In the pre-retrieval phase, we focus on optimizing the tool document by appending a high-quality name, description, argument schema, and key information, hypothetical questions that the tool can answer, and additional metadata. The intra-retrieval phase performs subsequent modules to transform the user question into a combination of decomposed queries and expanded or varied queries to query the Toolshed Knowledge Base and retrieve the top-k relevant tools.:

### 2.2 Task Planning for Tools

Similar to that of advanced RAG, task planning is essential for breaking down complex user queries into manageable sub-tasks– to which tools can be retrieved for each sub-task. Wei et al. (Chain-of-Thought) [37] and Yao et al. (ReAct) [30] lay the foundation by enabling LLM Agents to systematically decompose tasks, improving both interpretability and problem-solving. Yuan et al. (EasyTool) [38] and Huang et al. (PLUTO) [8] use Agents to decompose queries into independent sub-tasks and retrieve the relevant tools for those tasks. Our approach within *Advanced RAG-Tool Fusion* also leverages an agent for query decomposition. However, our ensemble-based method differs by utilizing multiple other pre-retrieval, intra-retrieval, and post-retrieval strategies to increase retrieval accuracy from those decomposed sub-tasks within a tool knowledge base (Fig. 2).

### 2.3 Tool Selection or Retrieval

Tool selection or retrieval addresses the fundamental "needle-in-a-haystack" [39], [22] problem in RAG for tools. Researchers adopt two distinct approaches: Retriever-based and LLM-based.

### 2.3.1 Retriever-based Tool Selection

Tool retrieval is tightly coupled to task planning for tools. Early retriever-based works such as TF-IDF [40] and BM25 [13] have been used for information retrieval (and more recently tool retrieval) by relying on exact term matching to align queries with documents or tools. Anantha et al. (ProTIP) [12] train a BERT-base-uncased model as a retriever to match decomposed queries to tool descriptions. Yuan et al. (CRAFT) [9] utilize a multi-view matching and ranking approach to retrieve tools by aligning queries with function names and docstrings, generating potential function names and docstrings at inference time, and leveraging SimCSE embeddings to optimize retrieval based on the original query, tool name, and documentation. Zheng et al. (ToolRerank) [41] combine Adaptive Truncation and Hierarchy-Aware Reranking to refine tool retrieval by adjusting to seen and unseen tools, while leveraging both a dual-encoder and cross-encoder to return fine-grained results for single and multi-tool queries. Chen et al. (Re-Invoke) [10] utilize synthetic queries (for each tool) along with tool descriptions for embeddings, intent extraction from the user query, and ranks the retrieved tools for each intent. Moon et al. (Tool2Vec) [42] also use synthetic query generation (but not tool descriptions) for embeddings, while training a multi-label classifier for a 2-step retrieval. Our approach differs from ProTip [12], CRAFT [9], ToolRerank [41] and Tool2Vec [42] in which we do not rely on labeled data to train retrievers–we utilize out-of-the-box embedders from model providers such as OpenAI, Databricks, Mistral to showcase the possibility of zero-shot usage without fine-tuning. While Re-Invoke [10] and ToolRerank [41] use generated queries to enhance the vector representations of the tool documents (reverse HyDe application of advanced RAG), our approach not only generates synthetic queries but also generates key-topics/themes/intents of each tool, generates high quality tool descriptions and clear argument schema parameter names and descriptions to embed along with the tool name. Furthermore, compared to Re-Invoke [10] solely generating user intents, *Advanced RAG-Tool Fusion*'s (Fig. 2) ensemble modules include initial query rewriting, query decomposition into sub-tasks (user intents) and multi-query expansion or variation on each sub-intent–focusing on capturing diverse ways of describing an intent to better match the correct tool with a sub-intent.

### 2.3.1 LLM-based Tool Selection

In addition to solely retriever-based tool selection, researchers have relied on LLMs for tool retrieval. Li et al. (API-Bank) [43] equip an LLM Agent with a tool to search APIs ("Plan+Retrieve+Call" paradigm) by passing in keywords it thinks it will retrieve a relevant tool–identical to the Agentic RAG approach.[9] The authors have noted a limitation that GPT-4 often will not call the search API tool [43]. Du et al. (AnyTool) [44] use function calling to retrieve tools in a hierarchy-based tool-category-API structure and incorporates a self-reflective mechanism if the LLM Agent deems the retrieved tool is not able to solve the user question. While our approach primarily relies on a retriever-based tool selection method, we also leverage an Agent for self-reflection and reranking of the retrieved tools from each subquery or expanded query, resulting in a final list of tools to equip the Agent. However, our approach does not use solely Agentic RAG [36] in the sense that we equip an agent with a retriever tool and relying on it to use it. Instead, we prompt the LLM to break down queries and always query the *Toolshed Knowledge Base*. In the post-retrieval phase, there is an option to implement Agentic RAG through Self-RAG, if after retrieval not all tools are found to solve the user question (See Appendix A for a detailed case study). Furthermore, *Advanced RAG-Tool Fusion* is not dependent on a hierarchical-based tool-category-API system, highlighting its plug-and-play capability for any set of tools. If there is a need to group tools or utilize hierarchical groups, *Advanced RAG-Tool Fusion* can utilize metadata filtering for the *Toolshed Knowledge Base* (Appendix A, Fig. 6).

### 2.4 Tool Calling

Past work also focuses on the pure tool invocation of the LLM by extracting the parameters and fine-tuning LLMs for tool-calling. Patil et al. (GorillaLLM) [7] enhance API calling by fine-tuning a LLaMA-7B model using retriever-aware training, allowing it to dynamically retrieve up-to-date API documentation and generate accurate function calls based on user queries, with additional verification through AST matching to prevent hallucinations. Qin et al. (ToolLLM) [6] fine-tune LLaMA on a dataset of over 16,000 real-world APIs, using a depth-first search decision tree (DFSDT) for enhanced reasoning and integrating an API retriever, achieving strong generalization to unseen APIs and performance comparable to ChatGPT. Liu et al. (ToolACE) [45] generate high-quality, diverse tool-learning data and finetune an LLM through a self-evolution process, ensuring high accuracy in tool-calling tasks. Hao et al. (ToolkenGPT) [46] represent tools as "toolken embeddings," which allow LLM Agents to invoke tools by generating "toolkens" and not require expensive fine-tuning with long tool definitions. Hao et al. (CITI) [47] use Mixture-Of-LoRA (MOLoRA) to fine-tune key components of LLMs, enhancing tool utilization while preserving general performance. The key difference between our approach and the tool calling prior work is that our focus is not to finetune an LLM for improved tool calling. Our approach uses out-of-the-box function-calling LLM Agents [3] from model providers such as OpenAI and Anthropic, who already finetuned their models for advanced function calling, and focuses on tool selection/retrieval and planning through the lens of *Advanced RAG-Tool Fusion*. Furthermore, we study the impact of varying the number of tool/function definitions passed into the "tools" parameter in OpenAI or Anthropic models, analyzing how different tool selection thresholds (*top-k*) affect retrieval accuracy and tool-calling performance in *Advanced RAG-Tool Fusion*.

## 3 Method

### 3.1 Tool Datasets

There are several notable datasets in the tool calling community, such as ToolBench [6], ToolAlpaca [48], ToolE [49], τ -bench [50] Seal-Tools [51]. However, upon reviewing the tool dataset and golden query-tool pairings (which link each user question with the correct tools and their parameters/arguments), we found several issues: some tool descriptions were unclear, certain tools lacked parameter details, many tools overlapped, and some queries could be solved by multiple overlapping tools. For this study, we selected Seal-Tools and ToolE as the primary datasets. Wu et al. (Seal-Tools) [51] created a tool dataset featuring single- and multi-reasoning trace queries, maintaining a high tool count (~4,000) while strategically preventing tool overlap, thereby minimizing unintended retrieval errors. Additionally, the ToolE dataset by Huang et al. [49], which is highly regarded in previous research, also features minimal tool overlap (~200 tools) and high-quality single- and multi-reasoning trace queries.

### 3.2 Models

Our approach is a plug-and-play method which does not require any pre-training of embedder retrievers or LLMs. This is because we want to adapt to future models and eliminate the start-up time and cost for pretraining of tool-labeled data. However, exploring finetuning embedders to increase retriever accuracy is a viable route of research–but the not the scope of this paper.

#### 3.2.1 LLM Models

In our approach, we primarily utilize the Azure OpenAI GPT-4o (2024-05-13) LLM model. Additionally, we conducted tests using the gpt-35-turbo-16k (0613) and gpt-4 (0613) models. Notably, OpenAI has stated that their gpt-4o-2024-08-06 model offers 100% reliability in generating tool-calling JSON schema outputs [1]. This ensures that, moving forward, outputs involving tool calls from the LLM will consistently generate valid JSON schemas without errors, such as missing required parameters or tool names. And, having reliable JSON outputs for tool calling further emphasizes the focus for a robust system for selecting/retrieving and planning tools effectively. Future work can explore additional model providers that contain function-calling abilities such as Anthropic and Meta.

#### 3.2.2 Embedder Models

We primarily utilize Azure OpenAI models for embedding tools within *Toolshed Knowledge Bases*: text-embedding-3-large, text-embedding-3-small, and text-embedding-ada-002 [52]. Future work could explore embedding models from other providers, such as Databricks BGE, Mistral, Jina, and Cohere (for reranking), to compare retrieval accuracy and performance.

### 3.3 Toolshed Knowledge Bases (Our Approach)

Efficient and intelligent storage of tools is a key component of the *Advanced RAG-Tool Fusion* framework. The *Toolshed Knowledge Base* serves as the vector database for storing tools that will be retrieved during inference-time and equipped to a Single Agent. The strategy we use to represent tool documents stems from the pre-retrieval phase of *Advanced RAG-Tool Fusion*, but the knowledge base itself is referred to as the *Toolshed Knowledge Base*.

In Fig. 2, we concatenate the tool name, description, argument schema, synthetic queries, and key topics of each tool as the vector representation of that tool and store it along with the other tools in the *Toolshed Knowledge Base*. Since tool names are required to have no spaces when using OpenAI function definitions, the tool name used in the enhanced concatenated tool document adds a space between words if needed to better represent them in the vector space–such as "GetRecord" being renamed to "Get Record" along with other concatenated features.

A crucial aspect of the *Toolshed Knowledge Base* is the inclusion of a metadata dictionary (Fig. 2) in each tool document, which contains the true tool name as it appears in the code repository. In the codebase, we initialize a dictionary (Appendix A) where the keys are the actual tool names (e.g., "get_record"), and the values are the corresponding Python tools or functions. During inference, the *top-k* relevant tool documents are retrieved, and the metadata dictionary is used to map the tool document's true name to its associated Python tool or function for further use.

In practice, a *Toolshed Knowledge Base* would contain a collection of tools for a Single Agent. Optimized for high quality retrieval, during inference-time (explained in more detail in 3.4) user queries or decomposed queries that aim to retrieve relevant tools will be well-represented across the vector space by these enhanced tool documents. In summary, the *Toolshed Knowledge Base* stores enhanced tool documents, with their representation strategy originating from our pre-retrieval approach in *Advanced RAG-Tool Fusion*.

### 3.4 Advanced RAG-Tool Fusion (Our Approach)

Since managing large scale tools and having Agents use the correct tool is fundamentally the same problem as the Retrieval-Augmented Generation "needle-in-a-haystack" [22], we can apply advanced RAG principles creatively to the tool retrieval/selection/planning field. While previous work touched on few, if any, individual components of advanced RAG within tool scaling, our approach–*Advanced RAG-Tool Fusion*–introduces an ensemble of state-of-the-art advanced RAG patterns applied to tool retrieval and planning across three clear phases. These phases (Fig. 2) are pre-retrieval, intra-retrieval, and post-retrieval. The rest of this section will give detailed descriptions of each phase of *Advanced RAG-Tool Fusion*. And the end of each phase section, we highlight the prescription for keys to success for applying the principles to production-grade agentic systems in your own tool-retrieval use cases. These keys to success are for the individual phases, if you want to know the broader considerations to use *Advanced RAG-Tool Fusion*, see Section 5. Furthermore, see Appendix A for a detailed hypothetical case study application of *Advanced RAG-Tool Fusion*.

#### 3.4.1 Pre-retrieval (*indexing*)

In the pre-retrieval or indexing phase of *Advanced RAG-Tool Fusion*, our goal is to enhance the quality of the tool document to be retrieved at a higher accuracy rate in the retrieval stage. The assumption is that storing a tool's name and description in a vector database does not yield satisfactory results when querying the vector database with a user question [10], [42]. Our approach in *Advanced RAG-Tool Fusion* enhances the tool documents with high quality descriptions (name, description, argument schema) along with key metadata and essential information before being indexed in the vector database. The method in which enhancing tool documents stems from the advanced RAG where researchers append document chunks with metadata, document summaries, key topics, and hypothetical questions [16].

As seen in Fig. 2, we apply this same logic to tools and understand which components make up a tool: 1) tool name, 2) tool description, 3) argument schema (parameters/description) (Appendix G). Then, we append strategic application of advanced RAG: 4) hypothetical questions each tool can answer (synthetically generated), 5) key topics/intents (synthetically generated). These 5 components are what makes up our tool document in the state-of-the-art *Advanced RAG-Tool Fusion* pre-retrieval phase. Once the vector embedding is created for these five components, it is stored in the *Toolshed Knowledge Base*, a vector database for tools. As pointed out in Section 3.3, tool names are attempted to add a space between them to enhance their representation in the vector space. We recommend exploring different combinations of these 5 components–different tool datasets may perform differently. Furthermore, adding new components to the tool, changing how the metadata is attached (e.g. metadata filtering) through hierarchical groupings of tools (Appendix A, Fig. 6), or using knowledge graphs are additional areas of research to increase the intra-retrieval phase accuracy in *Advanced RAG-Tool Fusion*.

**Leading practices for pre-retrieval (indexing).**

If opting to enhance the tool document with any 5 components:

1) *Tools* should not be identical/overlap, and *tool names* should be unique, and have a "embedded version" with spaces instead of underscore, dashes, or no spaces.

2) *Tool description* should be long, high quality and descriptive (such as when to use the tool and when not to).

3) Appending the function's *argument schema* can help retrieval. If appending, ensure parameter name and descriptions are clear and descriptive with no abbreviations.

4) Appending *synthetic questions* can increase retrieval. If appending, ensure questions are diverse, mirror future user questions about the tool, and utilize parameters in the question.

5) Appending *key topics/intents* can increase retrieval. If appending, generate the key topics based on not only the tool name and description, but also any synthetic questions.

### 3.4.2 Intra-retrieval (*inference-time*)

In the retrieval stage of *Advanced RAG-Tool Fusion*, our goal is to retrieve the correct tool(s) needed for a user question. The assumption is that simply querying the word-for-word user query does not capture the full essence of which tool should be retrieved–as users can use short-hand and not be very clear [17]. Therefore, our approach in *Advanced RAG-Tool Fusion* (Fig. 2) initially rewrites the query to fix any typos, errors, unclear pronouns (with available chat history), and overall conciseness. Additionally, a single user question may consist of multiple parts, often requiring several tools. Directly embedding and querying the entire question leads to poor retrieval results for tools [28]. *Advanced RAG-Tool Fusion* breaks the query into logical, independent steps, then rewrites and expands each step with relevant keywords to capture the essence of retrieving the correct tool from the *Toolshed Knowledge Base*. The corresponding advanced RAG methods applied to tool retrieval are query planning/decomposition, query rewriting, multi-query expansion or variation, and step-back prompting [16].

In Appendix A, Fig. 7, a complex user question is first decomposed into logical independent steps–this serves as the plan to solve the user question step-by-step. Then, for each step, it is rewritten and expanded into three (or any number of) variations queries, to attempt to capture the underlying meaning behind the user question. Finally, each of the three variations within each of the decomposed independent steps are used to query the *Toolshed Knowledge Base* to retrieve the *top-k* (in our example in Appendix A, Fig. 7, *top-k* = 5) most relevant tools. Along with query planning, decomposition, and initial query rewriting, multi-query expansion or variation is a critical component of *Advanced RAG-Tool Fusion*, as each a simple decomposed query or user intent might not be able to capture different ways to solve the decomposed query or intent. In the previous example in Appendix D, Fig. 18, the decomposed user question/intent "What is a neural network" can be solved in many different avenues: a research tool, a web search tool, an educational course tool, a mentor/tutor tool, a tool specific for neural networks, etc. Since this module has no awareness of the available tools in the *Toolshed Knowledge Base*, query rewriting and multi-query expansion or variation allows for the query "What is a neural network" to be rewritten in various ways to try to capture different meanings and retrieve the correct tool(s).

**Leading practices for intra-retrieval.**

If opting to add query decomposition, transformation, or more:

1) *Query planning* or decomposition is a "must-have" to separate out multiple tools for multiple steps in a complex query.

2) If user questions are short-hand and use pronouns, *rewriting* them initially (if applicable, utilize previous chat history) increases retrieval

3) If there are multiple tools that can solve the same question (but are not identical, see the "neural network" example in Appendix D, Fig. 18), *multi-query expansion or variation* can help identify diverse pathways to solve the question, which can help retrieval by broadening the search scope.

4) *Step-back query rewriting*, which is not a "must-have" can help answer abstract questions and can be used within the planning module.

5) Experiment the retrieval accuracy for *various top-k* (1, 5, 10, 15, 20, ≤ 128), and adjust the threshold as needed based on the complexity of your tool dataset.

### 3.4.3 Post-retrieval

In the post-retrieval phase of *Advanced RAG-Tool Fusion*, the goal is to finalize the list of tools for the agent (Fig. 2). Some irrelevant tools may pass through the intra-retrieval phase because they are similar enough to be retrieved but not useful for answering the user's question. To address this, we either discard these tools or keep them, selecting only the *top-k* most relevant tools. Using self-reflection, if the Agent believes it does not have all tools necessary to solve the user request, they can autonomously search the *Toolshed Knowledge Base* (Appendix A, Fig. 8). The associated advanced RAG patterns include reranking [34], [33], corrective RAG [20], and self-RAG [35].

Our approach in *Advanced RAG-Tool Fusion* uses an LLM-based reranker to rerank and indirectly correct tools in two key areas: 1) the rewritten/expanded queries/intents, and 2) the original decomposed query steps. As shown in Appendix D, Fig. 8, after each rewritten/expanded intent retrieves its own *top-k* (in this case, *top-k* = 5) most relevant tools to capture diverse tool solutions, a reranker condenses the total tools for the decomposed query down to *top-k*. An additional LLM-based reranker, if needed, is applied to each decomposed query step, further refining the tools to *top-k* for the original user question.

An LLM reranker can be replaced by an embedder reranker (cross-encoder) [33] to perform the identical reranking. However, the LLM can perform additional reasoning such as discarding irrelevant tools and condensing duplicate tool retrievals.

Finally, the final tools are provided to the agent, which can solve the user's question. The focus of this paper is retrieving relevant tools to equip to the agent, not the solving process [30].

| Retriever | Recall @ 1 | Recall @ 5 | Recall @ 10 | Dataset |
|---|---|---|---|---|
| BM25 | | 0.410 | 0.550 | Seal-Tools |
| Seal-Tools DPR | | 0.480 | 0.680 | Seal-Tools |
| Ours: Advanced RAG-Tool Fusion with Toolshed Knowledge Base | | **0.876** | **0.965** | Seal-Tools |
| | | | | |
| BM25 | 0.272 | 0.462 | | ToolE - Single Tool |
| Re-Invoke's | 0.672 | 0.871 | | ToolE - Single Tool |
| Ours: Advanced RAG-Tool Fusion with Toolshed Knowledge Base | **0.726** | **0.928** | | ToolE - Single Tool |
| | | | | |
| BM25 | 0.093 | 0.335 | | ToolE - Multi Tool |
| Re-Invoke's | 0.333 | 0.801 | | ToolE - Multi Tool |
| Ours: Advanced RAG-Tool Fusion with Toolshed Knowledge Base | **0.400** | **0.894** | | ToolE - Multi Tool |

**Table I.** Retriever results comparison on the Seal-Tools and ToolE datasets. We compared our *Advanced RAG-Tool Fusion* approach against a BM25 baseline, and the state-of-the-art retrieval accuracies of Seal-Tool's retriever and Re-Invoke for ToolE. For the metrics BM25, Seal-Tools [51], and Re-Invoke [10], we report the numbers available in the papers. The metrics are reported as recall@k, some *k* values were not calculated or clearly defined in the original benchmarks, thus not reproduced in our approach. Specifically, Recall @1 for ToolE – Multi Tool is irrelevant due to more than one tool needed, hence there will always be 50% or less accuracy. The best-performing method is highlighted in boldface.

**Leading practices for post-retrieval.**

If opting to add reranking, corrective or self-RAG:

1) As a post-requisite to intra-retrieval *Keys 1 and 3*, query decomposition and multi-query expansion, *reranking* the N sets of retrieved tools from the decomposed or expanded queries to the condensed final *top-k* is a "must-have."

2) If *reranking*, you can explore the retrieval accuracy vs. cost/latency trade-off of an embedder vs. LLM-based reranker.

3) Using an LLM as a *tool-grader* to better align the final *top-k* with relevant tools can yield satisfactory results.

4) Using *self-RAG* is not a "must-have," but if not all the correct tools were not retrieved after the aforementioned steps, adding *self-reflection* to fetch additional tools can be a solution.

5) There may be duplicate tools in each of decomposed query tool sets or query expansion tools; removing duplicates is a must.

## 4 Experiments

In this section, we describe two experiments and results to gauge the effectiveness of *Advanced RAG-Tool Fusion* and the impact of varying 1) the number of tools (*tool-M*) an agent has on Simple (Base) Agent accuracy and 2) the tool selection threshold (*top-k*) has on retrieval accuracy. While these experiments are different, the former is a direct study on our *Advanced Rag-Tool Fusion* approach and the latter dictates the configuration and optimization of values such as *top-k* within the *Advanced RAG-Tool Fusion* framework.

### 4.1 Scaling Tool Selection, Retrieval and Planning with Advanced RAG-Tool Fusion

In this experiment, we test the retrieval performance of *Advanced RAG-Tool Fusion* on available benchmark datasets and compare results with recall@k.

#### 4.1.1 Experiment Settings

As stated in 3.1, the benchmark dataset used to test retrieval performance is both Seal-Tools and ToolE. The total LLMs and embedders are stated Section 3.2 and total results are featured in Appendix I, but all final results in Section 4.1.2 utilize Azure OpenAI GPT-4o (2024-05-13) and Azure OpenAI text-embedding-3-large. For metrics, we report recall@k at available benchmark *k* values (1, 5, 10). When scaling the number of tools in 4.2 using *Advanced RAG-Tool Fusion*, recall@k values are varied to visualize practical application of the system.

#### 4.1.2 Results Analysis

Table I compares the retrieval results on the Seal-Tools and ToolE datasets. Each dataset is represented as a new section in the table. The first row of each section provides the baseline BM25 lexical search results, while the subsequent rows compare our *Advanced RAG-Tool Fusion* approach against state-of-the-art benchmarks, including Seal-Tools DPR and Re-Invoke.

Across all dataset benchmarks, our Advanced RAG-Tool Fusion methods outperform the BM25 baseline by approximately 46%. On the Seal-Tools dataset (single and multi-tool evaluation), Advanced RAG-Tool Fusion surpasses the Seal-Tools method by 41%. On the ToolE single-tool dataset, it outperforms Re-Invoke by 5%, and on the ToolE multi-tool dataset, by 9%. These metrics represent absolute improvements in Recall@5. Overall, these results strongly indicate that the ensemble of methods within *Advanced RAG-Tool Fusion* outperforms any single approach.

### 4.2 Varying Number of Tools (*tool-M*) and Selection Threshold (*top-k*) within Toolshed Knowledge Base

In this experiment, we first test the tool calling ability of an LLM Agent by incrementally equipping it *M* number of tools from 1-128. The evaluation framework we use to assess the tool calling ability is our own *Toolshed Evaluation Framework*, which granularly checks that the Agent correctly outputs the tool name, the parameter keys (tool/function parameter names), and the parameter values (arguments values). Details on the *Toolshed Evaluation Framework* are found in Appendix B. The reason behind testing the tool calling accuracy of a Simple Agent with *M* varied tools from 1-128 is to eventually plug in our *top-k* value to one of these *M* values to scale tool selection. In other words, when using *Advanced RAG-Tool Fusion* at any *top-k* value, the agent will have the tool calling ability of a Simple Agent (also known as a Base LLM Agent) where *tool-M = top-k*. This equation is explained in Section 4.2.2.

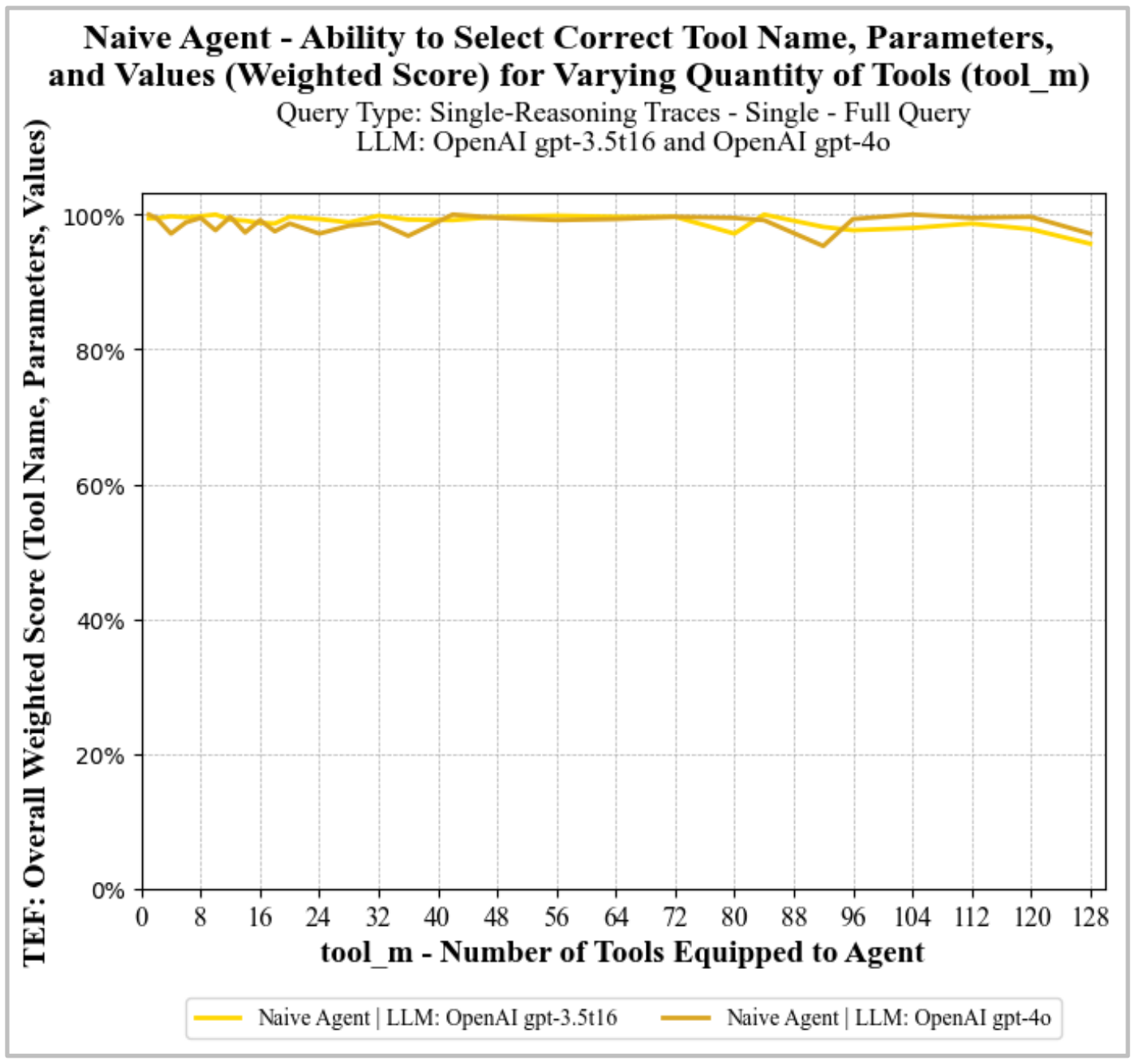

**Figure 3.** Impact of Varying the Number of Tools (*M*) Equipped to a Simple Agent (1-128) on its tool calling accuracy, for single reasoning trace queries. Full results including multi-reasoning traces are in Appendix C. The accuracy is a weighted score from the *Toolshed Evaluation Framework*, combining three metrics of a tool call: tool name, tool parameter keys, and tool parameter values (Appendix B).

Next, after a detailed analysis of the number of tools an agent can accurately call at any given *tool-M* level from 1-128, we test the retrieval accuracy (*Advanced RAG-Tool Fusion*) by varying both the selection threshold *top-k* (from 1-128) and the total tools in the *Toolshed Knowledge Base* (from 1-4000). In specific, for any given *tool-M* level, we vary *top-k* from 1-128, where *top-k tool-M* ≤. *Tool-M* is capped at ~4000 due to Seal-Tools [51] tool dataset. *Top-k* is capped at 128 since OpenAI (*gpt-4o-2024-08-06*) and other model providers allow up to 128 tool function definitions in the API request (within the *tools* parameter) [3], [4]. Finally, we vary a number of configurations within *Advanced RAG-Tool Fusion*: the 1) number of components that make up the tool embedding (name, description, argument schema, hypothetical questions, and key topics, 2) the embedder for the retriever, and 3) the Large Language Model.

### 4.2.1 Experiment Settings

As stated in 3.1, the tool dataset used to vary the total number of tools in the *Toolshed Knowledge Base* (*tool-M*) is Seal-Tools—which has roughly 4,000 unique tools, compared to ToolE's 200 tools. We vary the LLMs and embedders as stated in Section 3.2. The Simple Agent variation results are described in Appendix I. All final results for Section 4.2.3 utilize Azure OpenAI GPT-4o (2024-05-13) and Azure OpenAI text-embedding-3-large. We vary the *Toolshed Knowledge Base* configurations as stated in Section 4.2 (Appendix I). We vary *tool-M* from 1-4000. We vary *top-k* from 1-128 for each *tool-M* value, where *top-k* ≤ *tool-M*.

### 4.2.2 Modeling Advanced RAG-Tool Fusion at Different *tool-M* and Tool Selection Threshold *top-k* Values

In Appendix E, Fig. 19, the *Advanced RAG-Tool Fusion* equation is modeled for any given *tool-M* and *top-k* value. Simply put, since *Advanced RAG-Tool Fusion* allows the agent to only be equipped with a subset (*top-k*) of tools, when it actually has a large database of tools (*tool-M*), the agent's performance can then be measured by the retrieval accuracy (Fig. 4) multiplied by the agent's performance where *tool-M* = *top-k* (Fig. 3). This is important when choosing a *top-k* value/range for *Advanced RAG-Tool Fusion* (from 1-128). If a specific tool dataset is particularly troublesome for the agent (Appendix C, Fig. 16 and Fig. 17), choosing a lower *top-k* value where the retrieval accuracy is a satisfactory value is ideal. If an agent performs well across all *tool-M* values, then *top-k* becomes an optimization problem including retrieval accuracy, token cost, and latency. See Appendix A for a detailed case study.

### 4.2.3 Results Analysis

First, Fig. 3 compares the Simple (Base) Agent's tool calling ability while varying the number of tools equipped to the Agent. This tests the ability of a Simple Agent to sift through its total set of *M* tools and measures its capacity to reason and select the correct tool(s) based on the user query. The full results for parallel and sequential multi-reasoning trace queries are detailed in Appendix C. In other words, it is the function-calling version of document needle-in-a-haystack [22]. The weighted accuracy is measured by the *Toolshed Evaluation Framework* (Appendix B) which is the combination of whether the agent chose the correct tool name, correct tool parameter keys, and correct function arguments (parameter values). On every *tool-M* value (1-128), the Simple Agent's tool calling accuracy is around 97–100%. Thus, for the Seal-Tools dataset specifically, the number of tools (1 to 128) equipped to an agent (also the known as the variable *tool-M*) is not statistically significant in predicting the Simple Agent's accuracy [51]. However, for other tool datasets that are either confined to one specific domain (e.g. 1,000 tools for only finance) [50], require complex sequential multi-reasoning (Appendix F), involve multi-turn tool dependencies or state tracking [53], [50], or include human-in-the-loop elements [53], the number of tools may play a factor and should be studied further (see Appendix C, Fig. 16 and Fig. 17 for a complex hypothetical tool dataset and Simple or Base LLM Agent Accuracy).

Next, Fig. 4 and Fig. 5 compare the retrieval accuracy of multi-hop queries for *Advanced RAG-Tool Fusion* (Fig. 5) to the Seal-Tools DPR (Fig. 4), by varying the number of total tools (*M*) in the vector database and the selection threshold *top-k* of the retriever. In Fig. 4, for the Seal-Tools benchmark DPR [51], as *tool-M* increases, the retriever accuracy decreases significantly. Additionally, as *top-k* increases, the retriever accuracy improves. At the maximum *tool-M* level for the Seal-Tools dataset, the retrieval accuracy for any *top-k* level is significantly lower than for lower *tool-M* levels. However, in Fig. 5, for *Advanced RAG-Tool Fusion*, for any *tool-M* and *top-k* level, the retrieval accuracy is very strong around 95-100%. Once the *tool-M* level is greater than 500, the retrieval accuracy falls between 80-100% for *top-k* values below 6. See Appendix I for detailed results of this experiment with variations of the *Toolshed Knowledge Base* configurations, LLMs, and embedders.

## 5 Discussion

### 5.1 Advanced RAG-Tool Fusion Performance Analysis

We conduct several studies on the retrieval performance of *Advanced RAG-Tool Fusion*. Our findings indicate our ensembled-based approach significantly improves the baseline as well as previous state-of-the-art retrievers. We show that *Advanced RAG-Tool Fusion* can scale LLM Agents to hundreds or thousands of tools, without a significant drop in accuracy. Importantly, our approach uses no fine-tuning of an embedder or LLM, allowing researchers or enterprise practitioners to plug-and-play with *Advanced RAG-Tool Fusion*. See Appendix A for a detailed case study application of our approach. To summarize, in the pre-retrieval phase, we enhance the vector representations of tools by adding the tool name, description, argument schema, synthetical questions, and key intents/topics. We encourage researchers or industry practitioners to perform simple benchmarks on their own dataset to identify if their dataset requires or favors certain tool components. Moreover, if the tool description itself is very short, non-descriptive, or overlaps with other tool descriptions, we suggest having an LLM re-generate it to be more detailed and unique, such as Yuan et al. describes (EasyTool) [38]. Comparing results in Table I, the key differentiator on the Seal-Tools benchmark [51] is the query decomposition or planner module. This retrieval comparison across different *tool-M* and *top-k* values can be seen in Fig. 4 and Fig. 5. Additionally, the key differentiator between our approach on the ToolE benchmark [10] is the pre-retrieval, multi-query expansion or variation, query rewriting, and reranking modules. While our work moves the needle and bridges the gap between the tool retrieval, selection, planning and advanced RAG communities, more research is needed–particularly on the task planning side–to increase retrieval accuracy further. While the retrieval accuracy may not be perfect at Recall@5 or Recall@10, this is where we can optimize *top-k* for datasets that are more difficult for the retriever (Appendix C, Fig. 16 and Fig. 17).

### 5.2 Retrieval Impact of Number of Tools in Toolshed Knowledge Base and Choosing Selection Threshold

The consideration on how the 1) number of tools in a *Toolshed Knowledge Base* (*tool-M*) and 2) the selection threshold *top-k* affects retrieval accuracy has not been studied in this level of detail previously. Our work sheds light on the relationship between *tool-M* and *top-k* when building a retriever and agent system. We demonstrate how these factors enable scalable tool usage to any number of tools (*M*) and reveal the precise impact on accuracy (Fig. 3, Fig. 4, Fig. 5). In Appendix C, Fig. 14 and Fig. 15, we note that increasing the tool selection threshold will increase token costs under the hood for OpenAI and Anthropic API calls. Therefore, optimizing token cost has a trade-off with retrieval accuracy as well as Simple (Base) Agent accuracy. In a hypothetical scenario in Appendix C, Fig. 16 and Fig. 17, a Simple (Base) Agent (with a complex tool dataset [50], [53] or multi-hop sequential queries [28] that need human feedback and additional reasoning) struggles with selecting the correct tools to answer a user question. We demonstrate setting the *top-k* value lower (which decreases token cost too) to help the agent reason through less tools would be advised. However, we note that a low *top-k* may hinder retrieval accuracy. We show the trade-off between retrieval accuracy, agent performance, and token cost when selecting the *top-k* value and its impact on the number of tools in a *Toolshed Knowledge Base* (*tool-M*). *Advanced RAG-Tool Fusion* is critical for production-grade agentic applications; by increasing retrieval accuracy, we can confidently lower *top-k*, which consistently saves on token costs and increases agent performance when dealing with complex tool datasets.

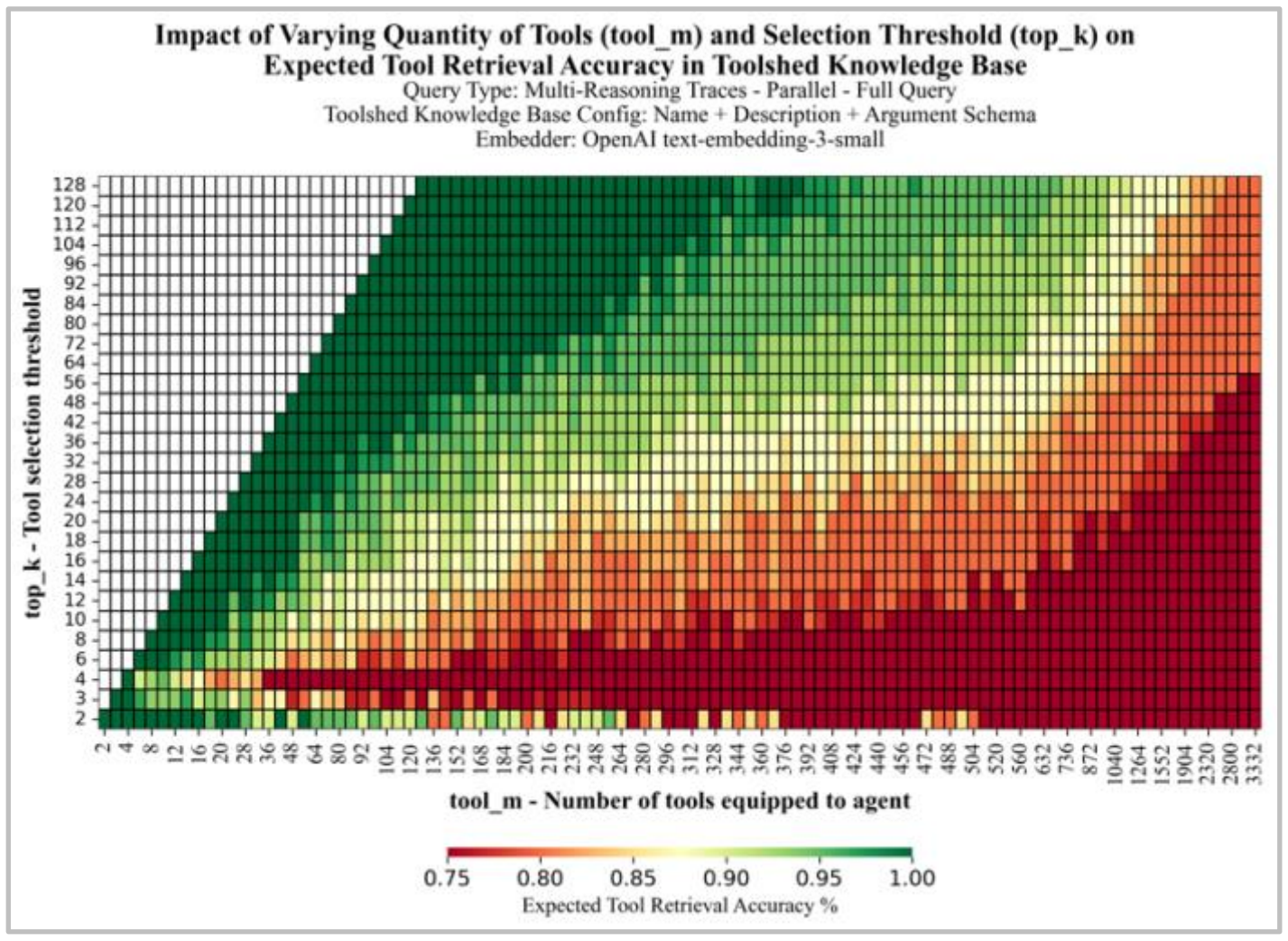

**Figure 4.** Impact of varying the selection threshold (*top-k*) and number of total tools (*tool-M*) from 1–4000 on retrieval accuracy of Seal-Tools DPR benchmark, without query decomposition.

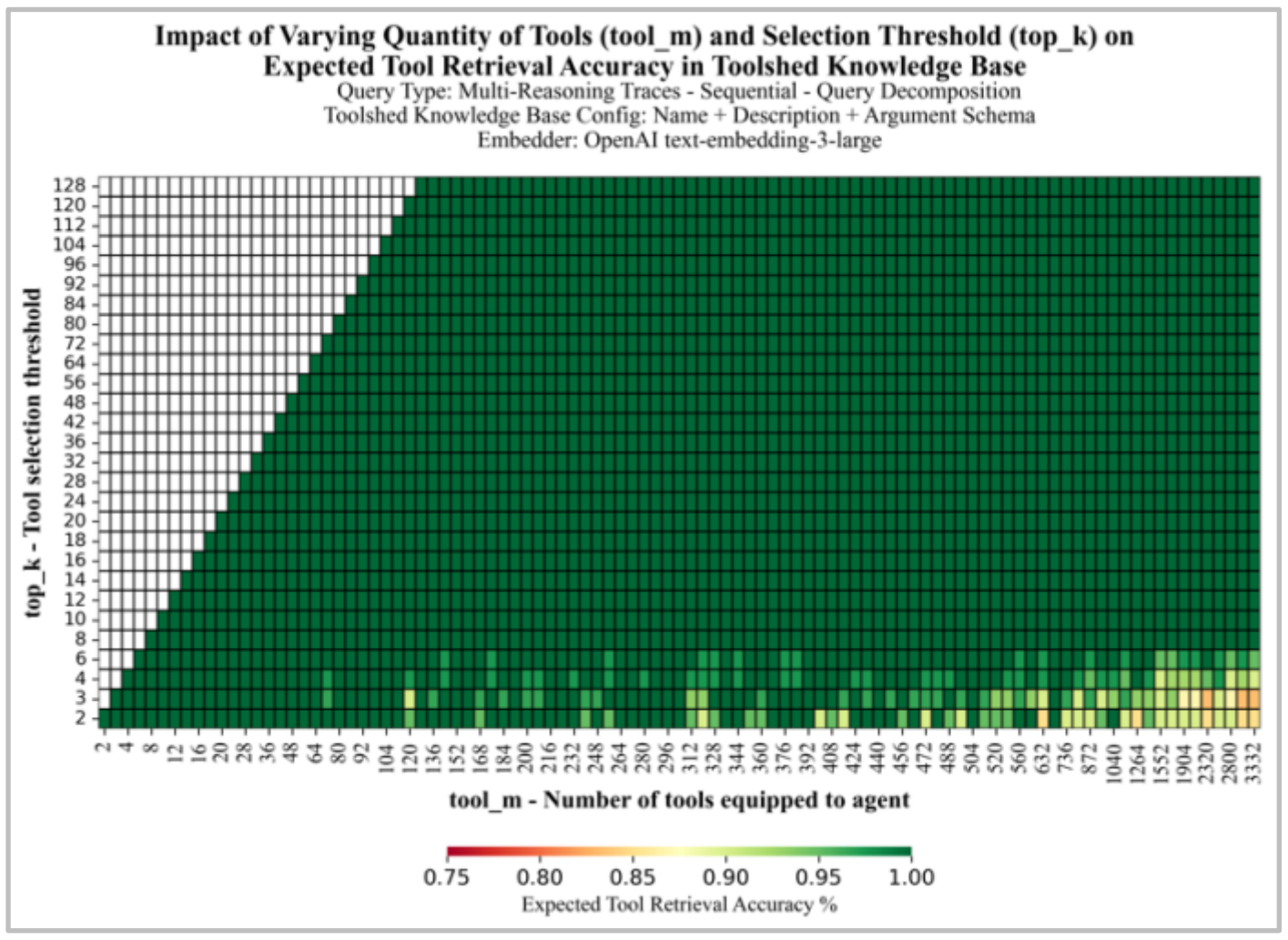

**Figure 5.** Impact of varying the selection threshold (*top-k*) and number of total tools (*tool-M*) from 1–4000 on retrieval accuracy of *Advanced RAG-Tool Fusion*. This approach uses query decomposition, significantly improving retrieval accuracy.

#### 5.2.1 Simple Agent Accuracy

As stated in 5.2, understanding the Simple Agent accuracy for a given tool dataset is critical when optimizing *Advanced RAG-Tool Fusion*. In our study, as seen in Fig. 3 and Appendix D, the number of tools equipped to a Simple Agent does not affect its accuracy for the Seal-Tools dataset. The reason is mostly likely two-fold. Firstly, the Seal-Tools [51] dataset contains a very diverse set of tools that do not overlap. That is, these tools span across different sectors such as finance, education, weather, healthcare, and more. Specifically, a single query's corresponding golden correct tool is clearly unique from others in its category. Whereas other tool datasets may have a user query that can be answered by multiple overlapping tools, causing an artificial decrease in accuracy. Another reason is the tool dataset only requires a single-turn tool call with no additional reasoning from the user or result of the tool. When there is additional reasoning from the tool call to call an additional tool or human-in-the-loop feedback needed, the Agent needs to reason further and may not choose correctly. For example, if the user is asking to schedule an appointment, the agent first asks the user what day/time their appointment should be and calls a "CheckSchedule" tool to see if its available. If it is, the agent calls the "CreateAppointment" tool. If not, the agent re-asks the user and loops through the previous steps until they successfully create the appointment. These types of questions are much more complex than the questions in the Seal-Tools dataset and need to be explored further. We showcase in Appendix C, Fig. 16 and Fig. 17 what the Simple Agent accuracy may look like for these complex questions, along with corresponding retrieval accuracy and token cost. Furthermore, we demonstrate that *Advanced RAG-Tool Fusion* not only enables scalable tool-calling agents, but also increases agent accuracy by limiting the scope of tools that an agent will have to reason with (for complex tool datasets). Recent work by Yao et al. (τ-bench) [50] and J. Lu et al. (ToolSandbox) [53] explores complex multi-turn human-in-the-loop tool datasets, but further benchmarking is needed with higher tool counts to evaluate the impact of limiting *top-k*, allowing the agent to reason with fewer tools while maintaining high accuracy (see Appendix D). These limitations are described in Section 7.

## 6 Conclusion

As agent applications become more complex and scale to hundreds or thousands of tools, there is a need to consistently retrieve the correct tools to answer a user question. In this work, we present *Advanced RAG-Tool Fusion*, an ensemble of advanced RAG patterns novelly applied to the tool selection/retrieval, and planning field. Our framework consists of strategies within three phases: pre-retrieval, intra-retrieval, and post-retrieval. We have demonstrated that this ensemble of methods enables scalable tool-equipped agents and significantly outperforms both the baseline and the independent application of any single method, without fine-tuning an LLM or retriever. Furthermore, we present *Toolshed Knowledge Bases*, the vector database to efficiently store the collection of tools during the pre-retrieval stage. Lastly, we study the impact of varying both 1) the total tools (*tool-M*) in the *Toolshed Knowledge Base* and 2) the tool selection threshold (*top-k*) on retrieval accuracy and Simple Agent tool calling ability. *Advanced RAG-Tool Fusion* moves the needle for scaling tool-equipped Agents and sheds light on the retrieval-threshold-cost-accuracy trade-off.

## 7 Limitations and Future Study

While we have presented *Advanced RAG-Tool Fusion* and pushed the needle forward in the field, there is more work to be done for production-grade agentic applications. We have identified a couple limitations that can influence future areas of research. First, the planning mechanism for breaking down user questions may require additional context about available tools, as well as human involvement, to achieve 100% accuracy. Currently, the planning module is tool-agnostic. However, adding context on the types of tools in the Toolshed Knowledge Base could help decompose queries into sub-queries. Additionally, if the planning agent can interact with users by asking clarifying multi-round questions, such as "to confirm, you want to do X and Y, correct?" it would improve the breakdown of sub-intents and enhance retrieval accuracy. While this would not be a zero-shot tool calling application, this is a trade-off that can be assessed if necessary to achieve higher retrieval accuracy.

The second limitation concerns optimizing the tool selection threshold when breaking down sub-queries and intents. Currently, we set a fixed selection threshold for the final agent, such as 24. Currently, we set a fixed threshold, such as 24, where three identified sub-intents each retrieve 8 tools. However, if one sub-intent is more complex, this fixed threshold may hinder accuracy. Future research could explore dynamic thresholds based on the complexity of retrieving tools for each sub-intent, capping at the overall selection threshold (e.g., 24).

The third limitation, or rather an unexplored aspect, involves applying our approach in production environments with multi-turn chat history. For example, assume we have a chatbot that starts the session with zero chat history. The user asks to calculate the net present value (NPV) of their finances, to which our chatbot responds using *Advanced RAG-Tool Fusion* to correctly identify the NPV() tool. However, the user asks a follow-up question: "what if the initial cost was $500 more?" Since this is a direct follow-up question that uses the same NPV() tool, should we use the 1) initialized agent with *top-k* tools from the first chat, the 2) single NPV() tool the agent previously used, or 3) re-run the modules again to attain the NPV() tool. For complex multi-round, human-in-the-loop questions, further research is needed [50], [53]. Nonetheless, bridging the gap between advanced RAG and tool selection, retrieval, and planning will allow for more advancements in the field. We hope that future contributions build on *Advanced RAG-Tool Fusion* with a common goal to maximize the potential tool-calling ability of LLM Agents through planning, selection, and retrieval.

## References

[1] OpenAI, "Introduction to structured outputs." 2024. [Online]. Available: https://platform.openai.com/docs/guides/structured-outputs/introduction

[2] L. Nguyen, "LangChain 101 — Lesson 3: Output parser." 2024. [Online]. Available: https://medium.com/@larry_nguyen/langchain-101-lesson-3-output-parser-406591b094d7

[3] OpenAI, "Function calling." 2024. [Online]. Available: https://platform.openai.com/docs/guides/function-calling

[4] Google Cloud, "Function declarations." 2024. [Online]. Available: https://cloud.google.com/vertex-ai/generative-ai/docs/multimodal/function-calling\#function-declarations

[5] C. Qu *et al.*, "Tool Learning with Large Language Models: A Survey." 2024. [Online]. Available: https://arxiv.org/abs/2405.17935

[6] Y. Qin *et al.*, "ToolLLM: Facilitating Large Language Models to Master 16000+ Real-world APIs." 2023. [Online]. Available: https://arxiv.org/abs/2307.16789
[7] S. G. Patil, T. Zhang, X. Wang, and J. E. Gonzalez, "Gorilla: Large Language Model Connected with Massive APIs." 2023. [Online]. Available: https://arxiv.org/abs/2305.15334
[8] T. Huang, D. Jung, and M. Chen, "Planning and Editing What You Retrieve for Enhanced Tool Learning." 2024. [Online]. Available: https://arxiv.org/abs/2404.00450
[9] L. Yuan, Y. Chen, X. Wang, Y. R. Fung, H. Peng, and H. Ji, "CRAFT: Customizing LLMs by Creating and Retrieving from Specialized Toolsets." 2024. [Online]. Available: https://arxiv.org/abs/2309.17428
[10] Y. Chen *et al.*, "Re-Invoke: Tool Invocation Rewriting for Zero-Shot Tool Retrieval." 2024. [Online]. Available: https://arxiv.org/abs/2408.01875
[11] C. Qu *et al.*, "Towards Completeness-Oriented Tool Retrieval for Large Language Models." 2024. doi: https://doi.org/10.1145/3627673.3679847.
[12] R. Anantha, B. Bandyopadhyay, A. Kashi, S. Mahinder, A. W. Hill, and S. Chappidi, "ProTIP: Progressive Tool Retrieval Improves Planning." 2023. [Online]. Available: https://arxiv.org/abs/2312.10332
[13] S. Robertson and H. Zaragoza, "The Probabilistic Relevance Framework: BM25 and Beyond," *Foundations and Trends® in Information Retrieval*, vol. 3, no. 4. pp. 333–389, 2009. doi: 10.1561/1500000019.
[14] Y. Gao *et al.*, "Retrieval-Augmented Generation for Large Language Models: A Survey." 2024. [Online]. Available: https://arxiv.org/abs/2312.10997
[15] B. Xu, Z. Peng, B. Lei, S. Mukherjee, Y. Liu, and D. Xu, "ReWOO: Decoupling Reasoning from Observations for Efficient Augmented Language Models." 2023. [Online]. Available: https://arxiv.org/abs/2305.18323
[16] A. Joshi, S. M. Sarwar, S. Varshney, S. Nag, S. Agrawal, and J. Naik, "REAPER: Reasoning based Retrieval Planning for Complex RAG Systems." 2024. [Online]. Available: https://arxiv.org/abs/2407.18553
[17] X. Ma, Y. Gong, P. He, H. Zhao, and N. Duan, "Query Rewriting for Retrieval-Augmented Large Language Models." 2023. [Online]. Available: https://arxiv.org/abs/2305.14283
[18] H. S. Zheng *et al.*, "Take a Step Back: Evoking Reasoning via Abstraction in Large Language Models." 2024. [Online]. Available: https://arxiv.org/abs/2310.06117
[19] A. H. Raudaschl, "Forget RAG, the future is RAG-Fusion: The next frontier of search: Retrieval Augmented Generation meets Reciprocal Rank Fusion and generated queries." 2023. [Online]. Available: https://towardsdatascience.com/forget-rag-the-future-is-rag-fusion-1147298d8ad1
[20] S.-Q. Yan, J.-C. Gu, Y. Zhu, and Z.-H. Ling, "Corrective Retrieval Augmented Generation," 2024, *arXiv*. doi: 10.48550/ARXIV.2401.15884.
[21] S. Jeong, J. Baek, S. Cho, S. J. Hwang, and J. C. Park, "Adaptive-RAG: Learning to Adapt Retrieval-Augmented Large Language Models through Question Complexity." 2024. [Online]. Available: https://arxiv.org/abs/2403.14403
[22] G. Kamradt, "Needle in a haystack - pressure testing LLMs: A simple 'needle in a haystack' analysis to test in-context retrieval ability of long context LLMs." 2023. [Online]. Available: https://github.com/gkamradt/LLMTest_Needle_in_a_Haystack
[23] L. Gao, X. Ma, J. Lin, and J. Callan, "Precise Zero-Shot Dense Retrieval without Relevance Labels." 2022. [Online]. Available: https://arxiv.org/abs/2212.10496
[24] R. Jagerman, H. Zhuang, Z. Qin, X. Wang, and M. Bendersky, "Query Expansion by Prompting Large Language Models." 2023. [Online]. Available: https://arxiv.org/abs/2305.03653
[25] L. Wang, N. Yang, and F. Wei, "Query2doc: Query Expansion with Large Language Models." 2023. [Online]. Available: https://arxiv.org/abs/2303.07678
[26] W. Peng *et al.*, "Large Language Model based Long-tail Query Rewriting in Taobao Search." 2024. [Online]. Available: https://arxiv.org/abs/2311.03758
[27] S. Setty, H. Thakkar, A. Lee, E. Chung, and N. Vidra, "Improving Retrieval for RAG based Question Answering Models on Financial Documents." 2024. [Online]. Available: https://arxiv.org/abs/2404.07221
[28] Y. Tang and Y. Yang, "MultiHop-RAG: Benchmarking Retrieval-Augmented Generation for Multi-Hop Queries." 2024. [Online]. Available: https://arxiv.org/abs/2401.15391
[29] H. Trivedi, N. Balasubramanian, T. Khot, and A. Sabharwal, "Interleaving Retrieval with Chain-of-Thought Reasoning for Knowledge-Intensive Multi-Step Questions." 2023. [Online]. Available: https://arxiv.org/abs/2212.10509
[30] S. Yao *et al.*, "ReAct: Synergizing Reasoning and Acting in Language Models." 2023. [Online]. Available: https://arxiv.org/abs/2210.03629
[31] O. Khattab *et al.*, "Demonstrate-Search-Predict: Composing retrieval and language models for knowledge-intensive NLP." 2023. [Online]. Available: https://arxiv.org/abs/2212.14024
[32] K. Sawarkar, A. Mangal, and S. R. Solanki, "Blended RAG: Improving RAG (Retriever-Augmented Generation) Accuracy with Semantic Search and Hybrid Query-Based Retrievers." 2024. [Online]. Available: https://arxiv.org/abs/2404.07220
[33] R. Theja, "Boosting RAG: Picking the best embedding & reranker models." 2023. [Online]. Available: https://www.llamaindex.ai/blog/boosting-rag-picking-the-best-embedding-reranker-models-42d079022e83
[34] W. Sun *et al.*, "Is ChatGPT Good at Search? Investigating Large Language Models as Re-Ranking Agents." 2023. [Online]. Available: https://arxiv.org/abs/2304.09542
[35] A. Asai, Z. Wu, Y. Wang, A. Sil, and H. Hajishirzi, "Self-RAG: Learning to Retrieve, Generate, and Critique through Self-Reflection." 2023. [Online]. Available: https://arxiv.org/abs/2310.11511
[36] A. Roucher, "Agentic RAG: Turbocharge your RAG with query reformulation and self-query!" 2023. [Online]. Available: https://huggingface.co/learn/cookbook/en/agent_rag

[37] J. Wei *et al.*, "Chain-of-Thought Prompting Elicits Reasoning in Large Language Models." 2023. [Online]. Available: https://arxiv.org/abs/2201.11903
[38] S. Yuan *et al.*, "EASYTOOL: Enhancing LLM-based Agents with Concise Tool Instruction." 2024. [Online]. Available: https://arxiv.org/abs/2401.06201
[39] N. F. Liu *et al.*, "Lost in the Middle: How Language Models Use Long Contexts." 2023. [Online]. Available: https://arxiv.org/abs/2307.03172
[40] K. Papineni, "Why Inverse Document Frequency?," in *Second Meeting of the North American Chapter of the Association for Computational Linguistics*, 2001. [Online]. Available: https://aclanthology.org/N01-1004
[41] Y. Zheng, P. Li, W. Liu, Y. Liu, J. Luan, and B. Wang, "ToolRerank: Adaptive and Hierarchy-Aware Reranking for Tool Retrieval." 2024. [Online]. Available: https://arxiv.org/abs/2403.06551
[42] S. Moon *et al.*, "Efficient and Scalable Estimation of Tool Representations in Vector Space." 2024. [Online]. Available: https://arxiv.org/abs/2409.02141
[43] M. Li *et al.*, "API-Bank: A Comprehensive Benchmark for Tool-Augmented LLMs." 2023. [Online]. Available: https://arxiv.org/abs/2304.08244
[44] Y. Du, F. Wei, and H. Zhang, "AnyTool: Self-Reflective, Hierarchical Agents for Large-Scale API Calls." 2024. [Online]. Available: https://arxiv.org/abs/2402.04253
[45] W. Liu *et al.*, "ToolACE: Winning the Points of LLM Function Calling." 2024. [Online]. Available: https://arxiv.org/abs/2409.00920
[46] S. Hao, T. Liu, Z. Wang, and Z. Hu, "ToolkenGPT: Augmenting Frozen Language Models with Massive Tools via Tool Embeddings." 2024. [Online]. Available: https://arxiv.org/abs/2305.11554
[47] Y. Hao *et al.*, "CITI: Enhancing Tool Utilizing Ability in Large Language Models without Sacrificing General Performance." 2024. [Online]. Available: https://arxiv.org/abs/2409.13202
[48] Q. Tang *et al.*, "ToolAlpaca: Generalized Tool Learning for Language Models with 3000 Simulated Cases." 2023. [Online]. Available: https://arxiv.org/abs/2306.05301
[49] Y. Huang *et al.*, "MetaTool Benchmark for Large Language Models: Deciding Whether to Use Tools and Which to Use." 2024. [Online]. Available: https://arxiv.org/abs/2310.03128
[50] S. Yao, N. Shinn, P. Razavi, and K. Narasimhan, "τ-bench: A Benchmark for Tool-Agent-User Interaction in Real-World Domains." 2024. [Online]. Available: https://arxiv.org/abs/2406.12045
[51] M. Wu, T. Zhu, H. Han, C. Tan, X. Zhang, and W. Chen, "Seal-Tools: Self-Instruct Tool Learning Dataset for Agent Tuning and Detailed Benchmark." 2024. [Online]. Available: https://arxiv.org/abs/2405.08355
[52] OpenAI, "OpenAI Embeddings Documentation." 2024. [Online]. Available: https://platform.openai.com/docs/guides/embeddings
[53] J. Lu *et al.*, "ToolSandbox: A Stateful, Conversational, Interactive Evaluation Benchmark for LLM Tool Use Capabilities," Aug. 08, 2024, *arXiv*: arXiv:2408.04682. Accessed: Oct. 17, 2024. [Online]. Available: http://arxiv.org/abs/2408.04682

## Appendix A: Hypothetical Case Study Application of Advanced RAG-Tool Fusion in all Three Phases

The following is a hypothetical case study application of *Advanced RAG-Tool Fusion*. The three phases outline necessary considerations, modules to build, and step-by-step examples. Lastly, we show logistics and maintenance practices for production.

**Hypothetical Situation**

You have 3,000 tools or functions and aim to create a multi-agent system. Each agent uses a dedicated *Toolshed Knowledge Base* and implements *Advanced RAG-Tool Fusion* for optimized retrieval and execution. The tools are divided among the following sub-agents:

- **1,000 Finance Tools**: Focused on financial operations and analytics.
- **1,000 Database Operation Tools:** Designed for database management and query execution.
- **1,000 Healthcare Tools: Tailored for healthcare-related tasks and insights.**

**Configuration (per each *TSKB*):**

- LLM: *AOI gpt-4o-2024-08-06*
- Embedder: *AOI text-embedding-3-small*
- Tool-M: *1000 (per each TSKB)*
- Top-k: 10 *(final agent)*

**Pre-retrieval (indexing)**

The following steps outline the pre-processing pipeline needed for each set of tools. After step 6, store the tool documents in a vector database.

1. Create Clear, Descriptive Names for Each Tool
   - Each tool should have a name that clearly describes its function.

   > e.g. "get_net_present_value"

2. Develop Clear Descriptions for Each Tool
   - Provide a detailed description of each tool, including specific scenarios where it would be used and where it wouldn't.

   > *"Calculates the net present value (NPV) of a series of cash inflows and outflows over a specified period, discounted to present value based on a given rate. Useful for determining the value of future cash flows, particularly in investment scenarios, when provided with initial investment, discount rate, and time period."*

3. Define Clear Argument Schema with Parameter Names and Descriptions for Each Tool
   - List and define the input arguments for each tool.

   > initial_value: *"The initial cash flow at the start of the period, which could be an investment, cost, or inflow."*
   > start_date: *"The beginning date of the cash flow period."*
   > end_date: *"The end date of the cash flow period."*
   > discount_rate: *"The rate used to discount future cash flows to their present value."*
   > scrap_value: *"The final residual value of the asset at the end of the period."*
   > cash_flows: *"A series of inflows and outflows over the specified period."*

4. Optional: Generate Hypothetical Questions for Each Tool *(may help with retrieval)*
   - Add 1-10 diverse user questions related to the tool. Use existing questions if available. Try to include parameters.

   > 1. "What is the NPV for a project starting on January 1, 2025, with an initial outflow of $100,000, annual cash flows of $15,000, and a discount rate of 8%?"
   > 2. "Calculate the net present value for cash flows of $20,000 per year over 10 years, with a 7% discount rate."
   > 3. "What is the NPV if my project ends in December 2030, with an initial cost of $50,000 and a scrap value of $5,000?"

5. Optional: Generate Key Topics, Themes, or Intents for Each Tool *(may help with retrieval)*
   - Add 1–10 key topics or intents associated with the tool. Be concise and differentiable from other tools. Can generate from the hypothetical questions and tool name, description, and argument schema.

   > 1. "Investment Valuation"
   > 2. "Cash Flow Analysis"

6. Metadata for Tool Name in Code Repository

   > {"tool_name":"get_net_present_value"}

**Figure 6.** Hypothetical Case Study Application of *Advanced RAG-Tool Fusion*, pre-retrieval (indexing) phase.

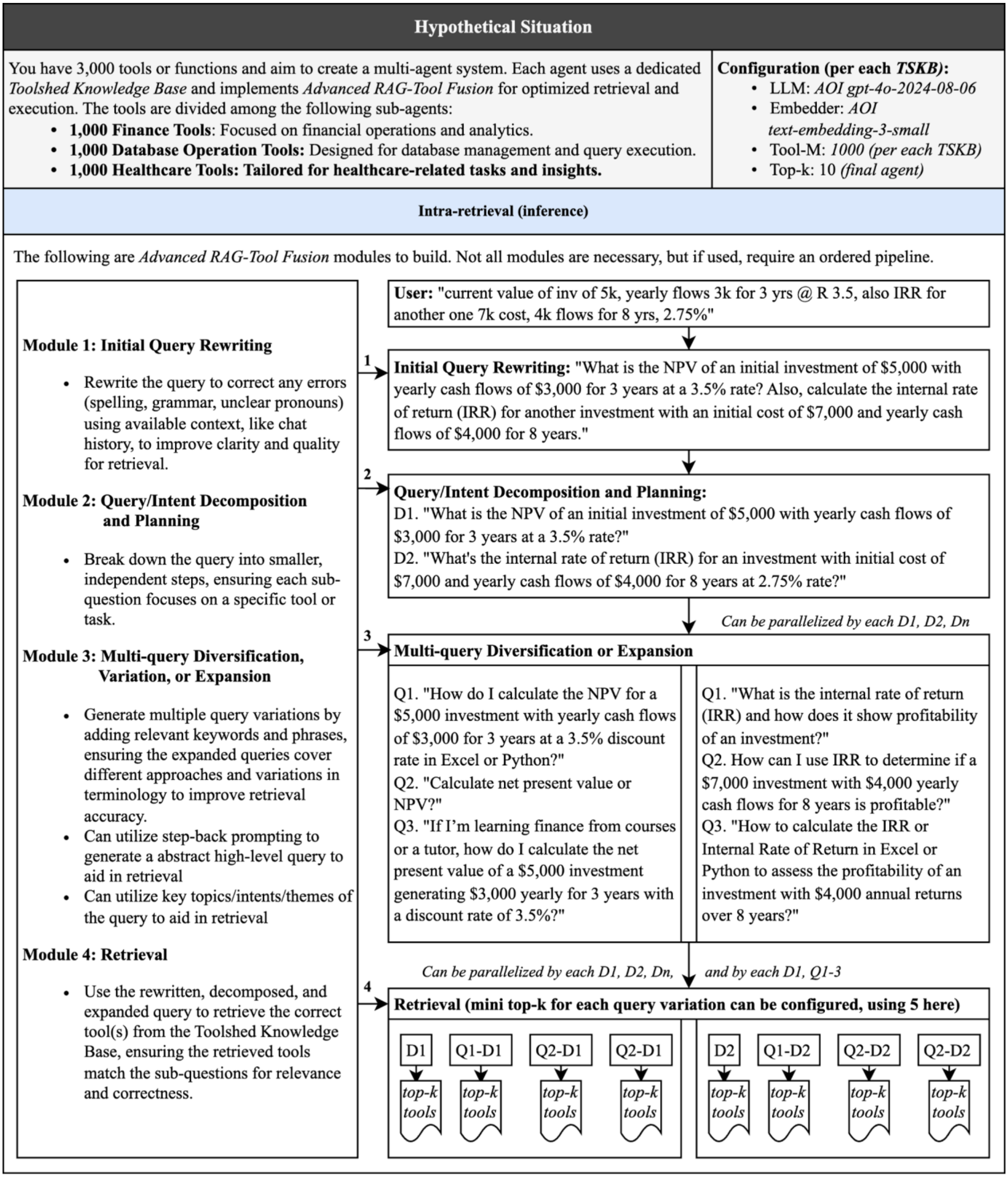

**Figure 7.** Hypothetical Case Study Application of *Advanced RAG-Tool Fusion*, intra-retrieval (inference) phase.

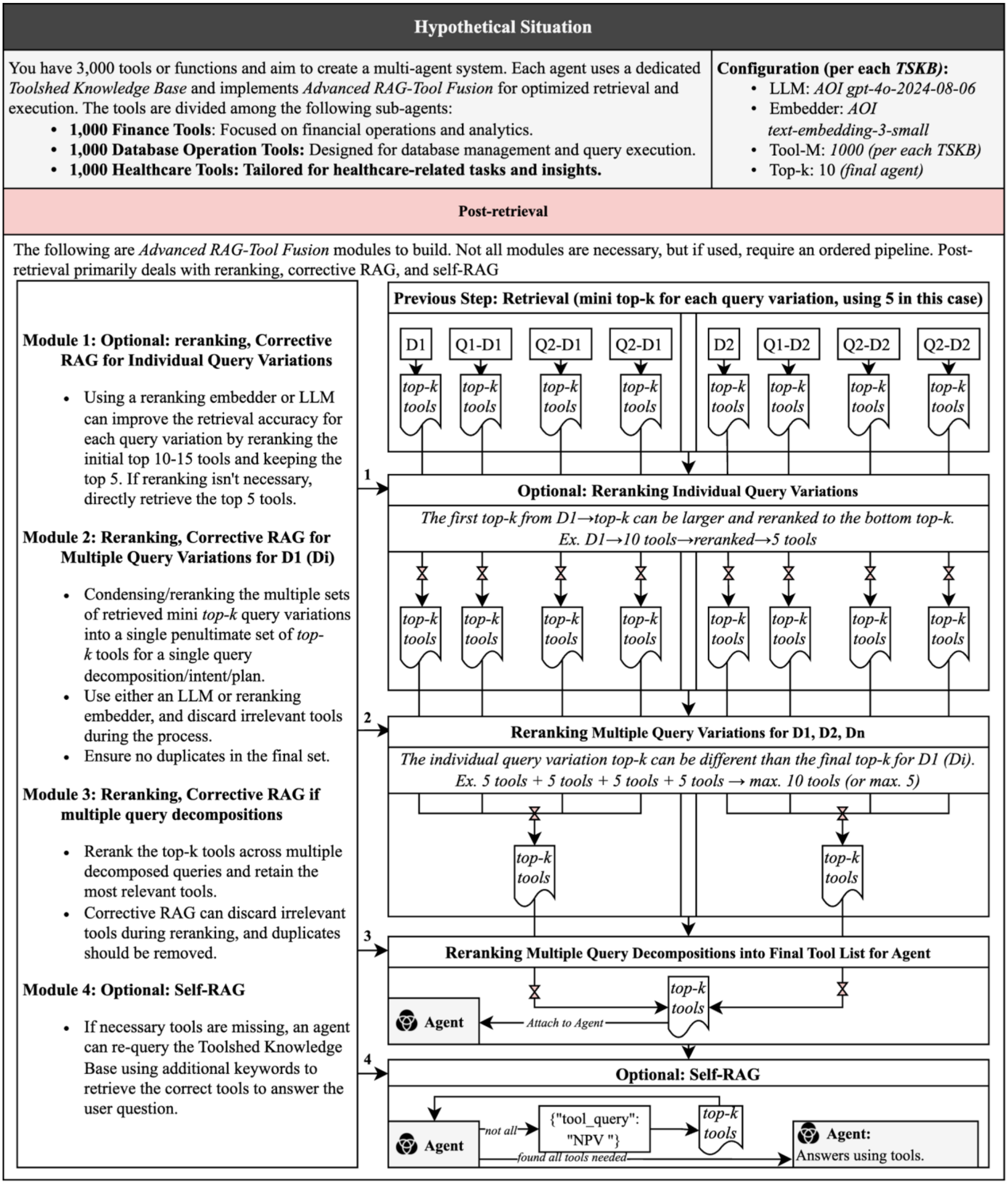

**Figure 8.** Hypothetical Case Study Application of *Advanced RAG-Tool Fusion*, post-retrieval phase.

**Hypothetical Situation**

You have 3,000 tools or functions and aim to create a multi-agent system. Each agent uses a dedicated *Toolshed Knowledge Base* and implements *Advanced RAG-Tool Fusion* for optimized retrieval and execution. The tools are divided among the following sub-agents:

- **1,000 Finance Tools**: Focused on financial operations and analytics.
- **1,000 Database Operation Tools:** Designed for database management and query execution.
- **1,000 Healthcare Tools: Tailored for healthcare-related tasks and insights.**

**Configuration (per each *TSKB)*:**

- LLM: *AOI gpt-4o-2024-08-06*
- Embedder: *AOI text-embedding-3-small*
- Tool-M: *1000 (per each TSKB)*
- Top-k: 10 *(final agent)*

**Logistics and Maintenance of Toolshed Knowledge Base with Tool Calling**

The following steps outline best practice considerations when using *Advanced RAG-Tool Fusion* and *Toolshed Knowledge Bases in production.*

1. Unified tools.py for each *Toolshed Knowledge Base*
   - To aid in maintaining the tools or functions, each set of tools used for a *Toolshed Knowledge Base* should be separated.

   finance_tools.py -- the collection of 1,000 finance tools in whatever tool creation framework of your choice.

2. Ability to add a new tool to the *Toolshed Knowledge Base*
   - Automated systems in place to add a new tool/function to the *Toolshed Knowledge Base.*
   - *Recommended solution:* Use hashes to track changes in tool name, description, argument schema, and any appended questions or key topics/intents. Steps: generate a unique hash for each tool and compare it to the previously stored hash to identify when re-indexing or updates are necessary.

3. Ability to delete a tool to the *Toolshed Knowledge Base*
   - Automated systems in place to add a new tool/function to the *Toolshed Knowledge Base.*
   - *Recommended solution:* Use hashes to track changes in tool name, description, argument schema, and any appended questions or key topics/intents. Steps: generate a unique hash for each tool and compare it to the previously stored hash to identify when re-indexing or updates are necessary.

4. Ability to update a tool to the Toolshed Knowledge Base
   - Automated systems in place to update an existing tool/function to the *Toolshed Knowledge Base.*
   - *Recommended solution:* Use hashes to track changes in tool name, description, argument schema, and any appended questions or key topics/intents. Steps: generate a unique hash for each tool and compare it to the previously stored hash to identify when re-indexing or updates are necessary.

5. Generate a Toolshed Dictionary from the tools.py file for actual agent-tool execution
   - The Toolshed dictionary will serve as a in-app, inference-time look-up, where each key is the tool name in tools.py, and the value is the actual python tool or function.
   - After *Advanced RAG-Tool Fusion* retrieves the *top-k* relevant tools to equip to an agent, these tools are actually the documents represented by the vector database. For each retrieved tool document, use the tool document metadata key "tool_name" that we set up in step 6 of phase 1 pre-retrieval/indexing to access the key-value pair of the Toolshed Dictionary. You can then attach these functions to the agent in the framework of your choice.

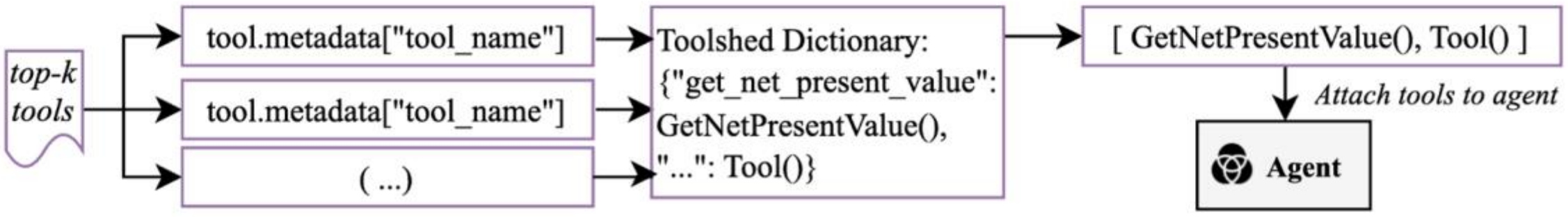

**Figure 9.** Hypothetical Case Study Application of *Advanced RAG-Tool Fusion*, logistics and maintenance of *Toolshed Knowledge Base*.

## Appendix B: Toolshed Evaluation Framework

The Toolshed Evaluation Framework builds upon previous work such as ToolEval [6] by adding granular metrics within a tool call: 1) tool name, 2) tool parameter keys, and 3) tool parameter values. We added granular metrics to identify whether an agent's poor performance is due to its inability to reason through available tools and output the correct tool name, understand the parameter keys, or correctly input parameter values. As seen below, based on the golden dataset and agent response, recall at the sub-metric level is computed. Furthermore, weighted score is a combination of all three sub-metrics (50% for tool name, 25% for parameter keys and values, respectively. If there are multiple tool calls in a single golden QA set, these sub-metrics are averaged.

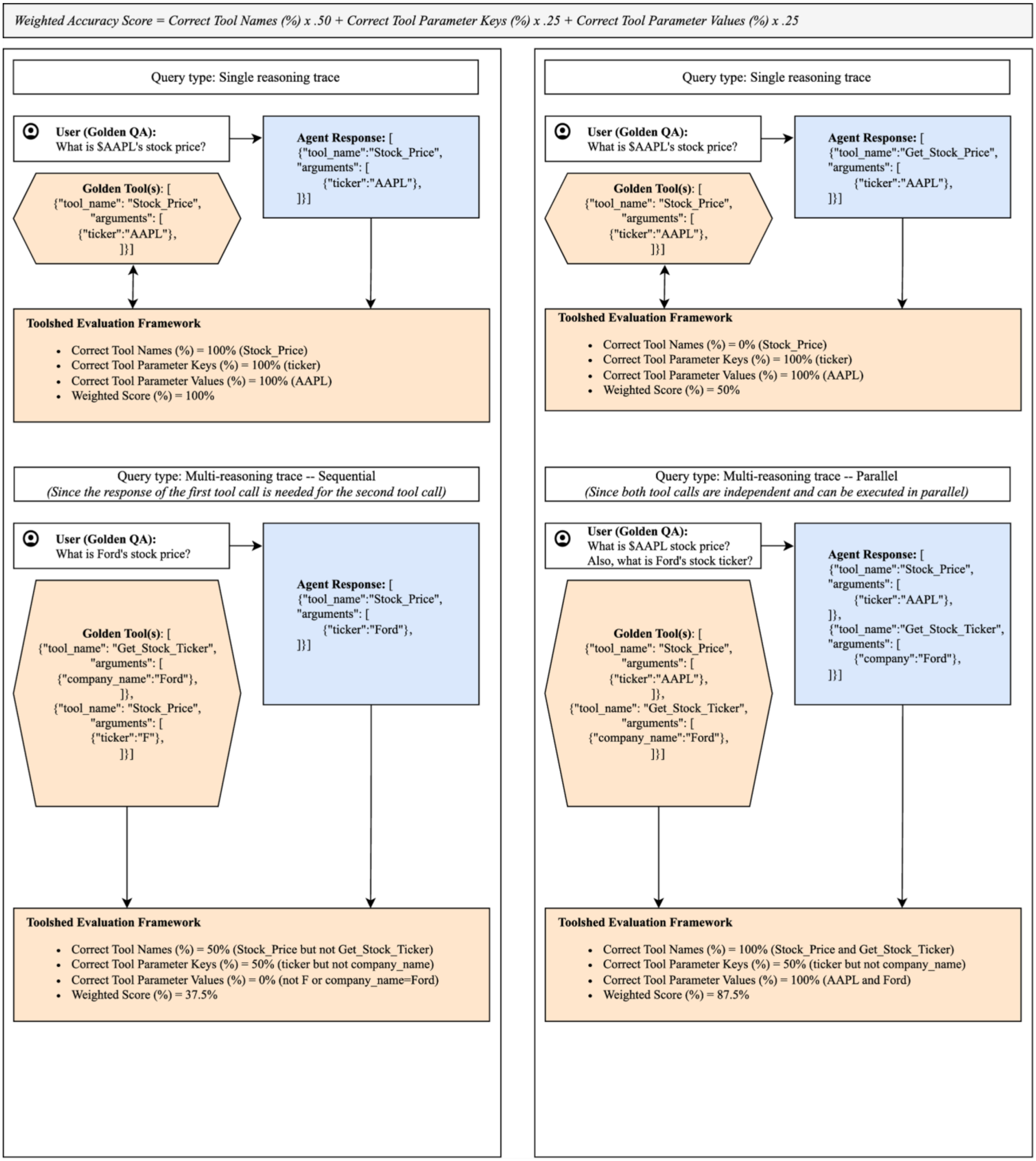

**Figure 10.** *Toolshed Evaluation Framework.*

# Appendix C: Comparing Simple Agent to Advanced RAG-Tool Fusion by Varying the Number of Tools (*tool-M*) and Selection Threshold (*top-k*) and measuring Simple Agent Accuracy, Token Count (Cost)

## 1. Measuring Accuracy

The following graphs compare the Seal-Tools results for tool calling weighted accuracy of a Simple Agent and *Advanced RAG-Tool Fusion* at fixed selection threshold (*top-k*) at increasing *tool-M* levels from 1-4000. The reason why the Simple Agent drops to 0% at *tool-M* = 129 is because model providers such as OpenAI (gpt-4o-2024-08-06), Anthropic (claude-3-5-sonnet-20240620), Google Gemini 1.5 Pro (gemini-1.5-pro-002) support up to 128 tool function definitions in the API request [3], [4]. Thus, *Advanced RAG-Tool Fusion* with a fixed *top-k* less than or equal to 128 will always be a viable option versus the Simple Agent.

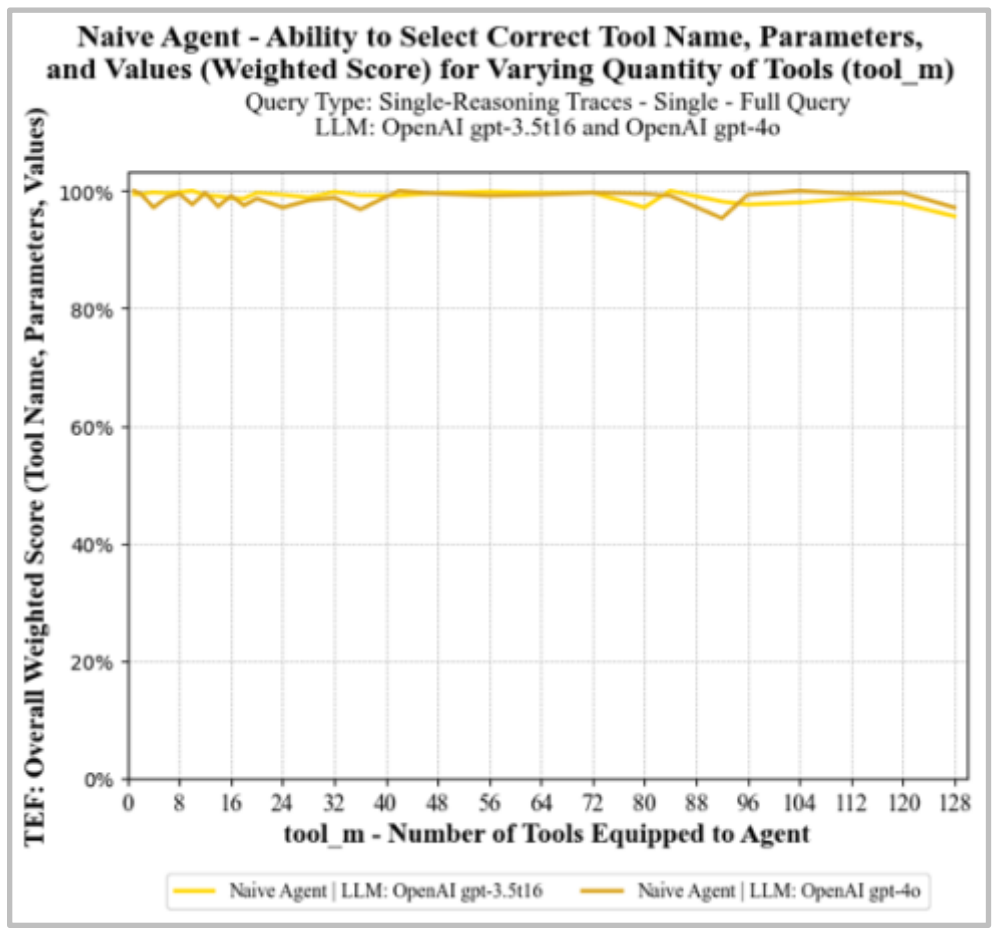

**Figure 10.** Simple Agent Accuracy for Single Reasoning Traces, varying Number of Tools (*M*) from 1-128.

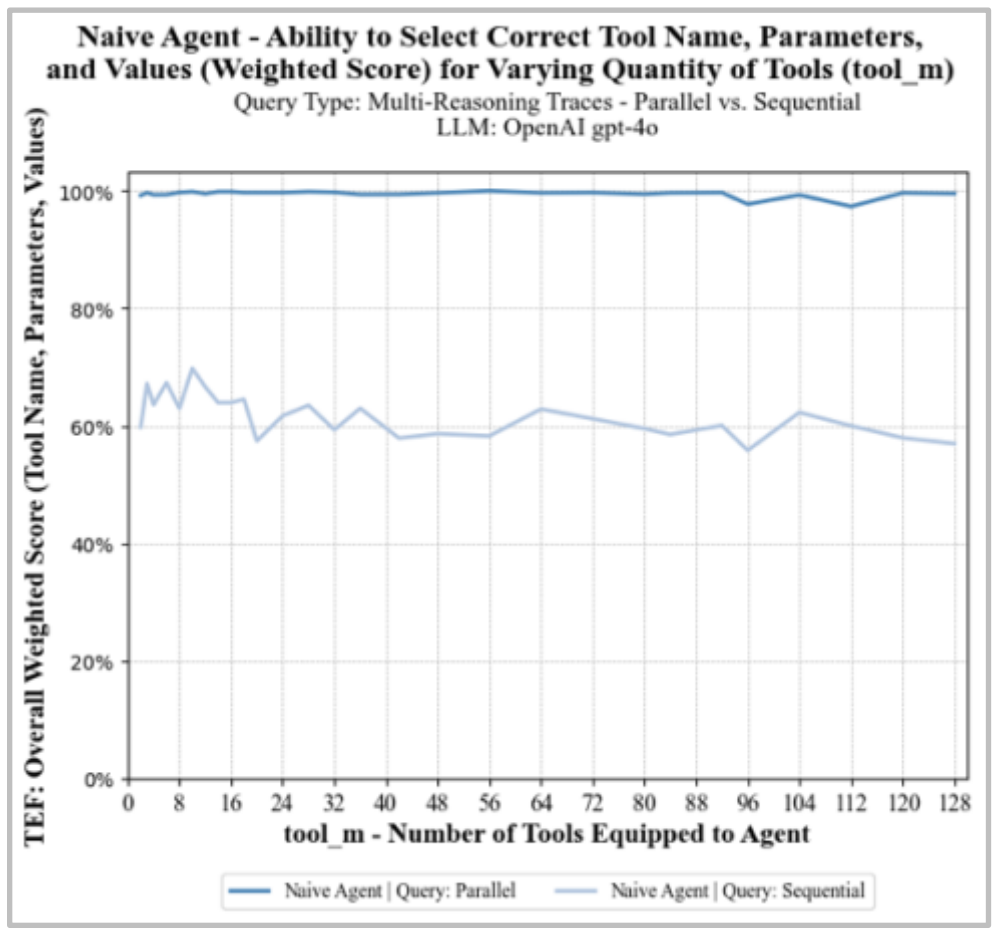

**Figure 11.** Simple Agent Accuracy for Multi-Reasoning Traces, varying Number of Tools (*M*) from 1-128.

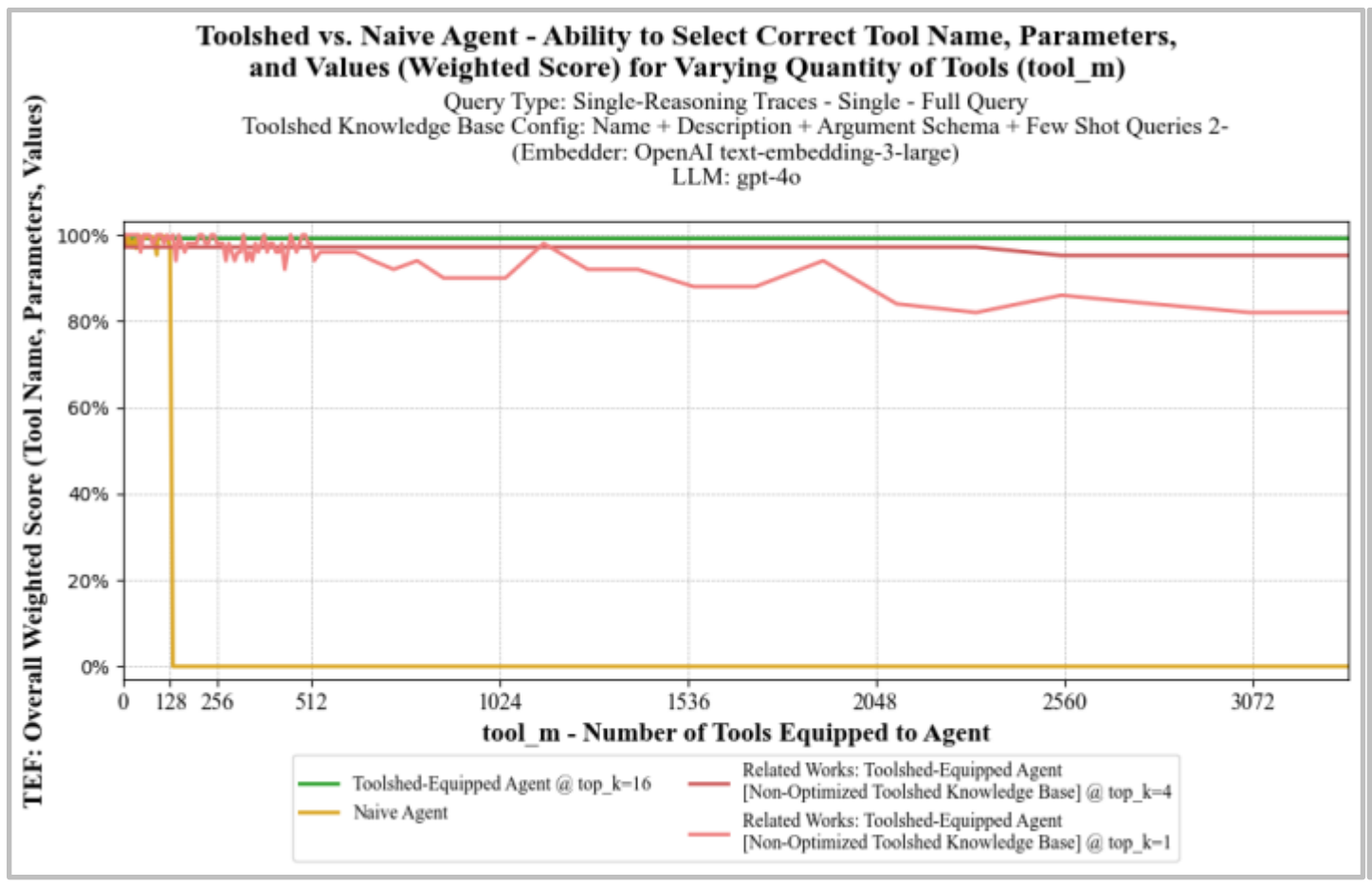

**Figure 12.** Comparing Simple Agent Accuracy vs. Agent using *Advanced RAG-Tool Fusion* (*Toolshed*) for Single-Reasoning Traces, varying Number of Tools (*M*) from 1-4000 and Tool Selection Threshold (*top-k*) for Advanced RAG-Tool Fusion retrieval. Additionally, we compare a retriever (in red) that does consider varying *top-k* when utilizing Agents.

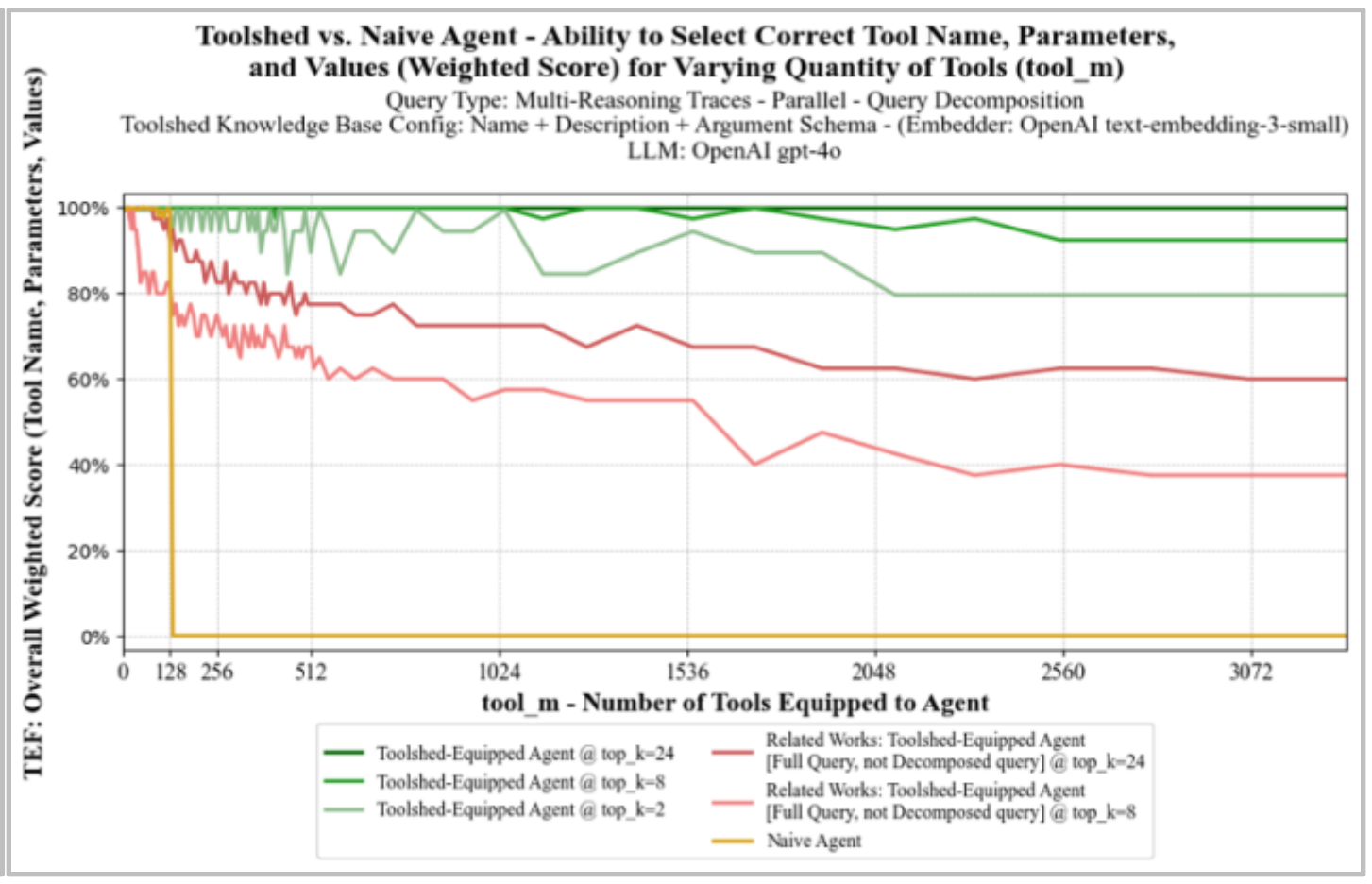

**Figure 13.** Comparing Simple Agent Accuracy vs. Agent using *Advanced RAG-Tool Fusion* (*Toolshed*) for Multi-Reasoning Traces, varying Number of Tools (*M*) from 1-4000 and Tool Selection Threshold (*top-k*) for Advanced RAG-Tool Fusion retrieval. Additionally, we compare a retriever (in red) that does not utilize Advanced RAG-Tool Fusion methods such as query decomposition to break down multi-reasoning traces. For our methods in green, we utilize a larger *top-k* since each sub-decomposed-query will effectively utilize a sub-*top-k* of the larger *top-k* divided by the N number of reasoning traces. See Appendix X for this sub-*top-k* breakdown.

**2. Measuring Token Cost**

The following graphs compare the Seal-Tools results for tool calling token counts for Simple Agent and *Advanced RAG-Tool Fusion* at fixed selection threshold (*top-k*) at increasing *tool-M* levels from 1-4000. These graphs highlight the increasing token count for more tools equipped to an Agent. Additionally, fixing the tools at a given *top-k* value will prevent unnecessary token costs.

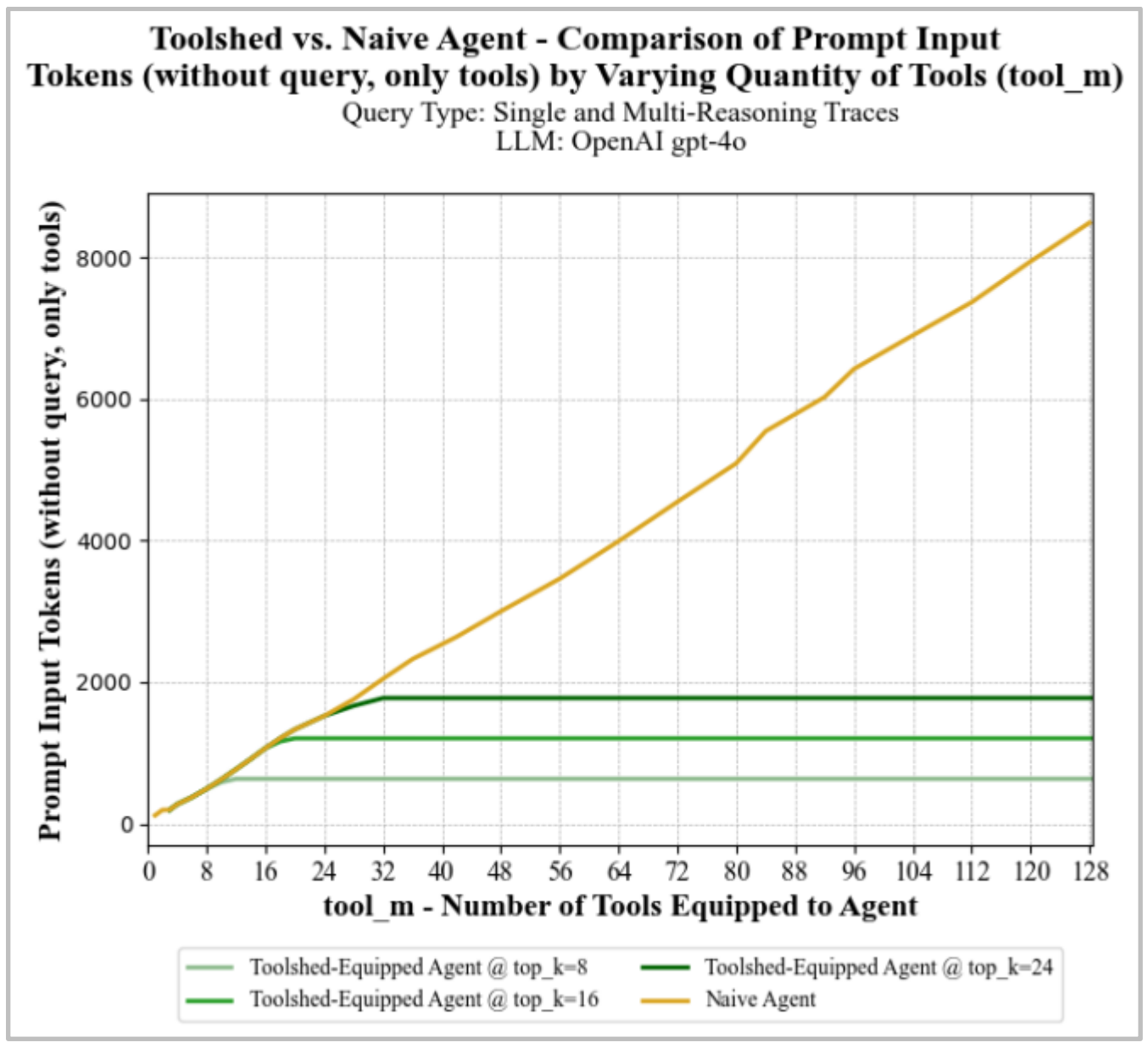

**Figure 14.** Comparing Prompt Input Tokens (less user query, only tools) for a Simple Agent with *M* number of tools equipped from 1-128 to an Agent using *Advanced RAG-Tool Fusion* and limiting *top-k* where the retrieval accuracy is satisfactory (90%+) to save on token costs.

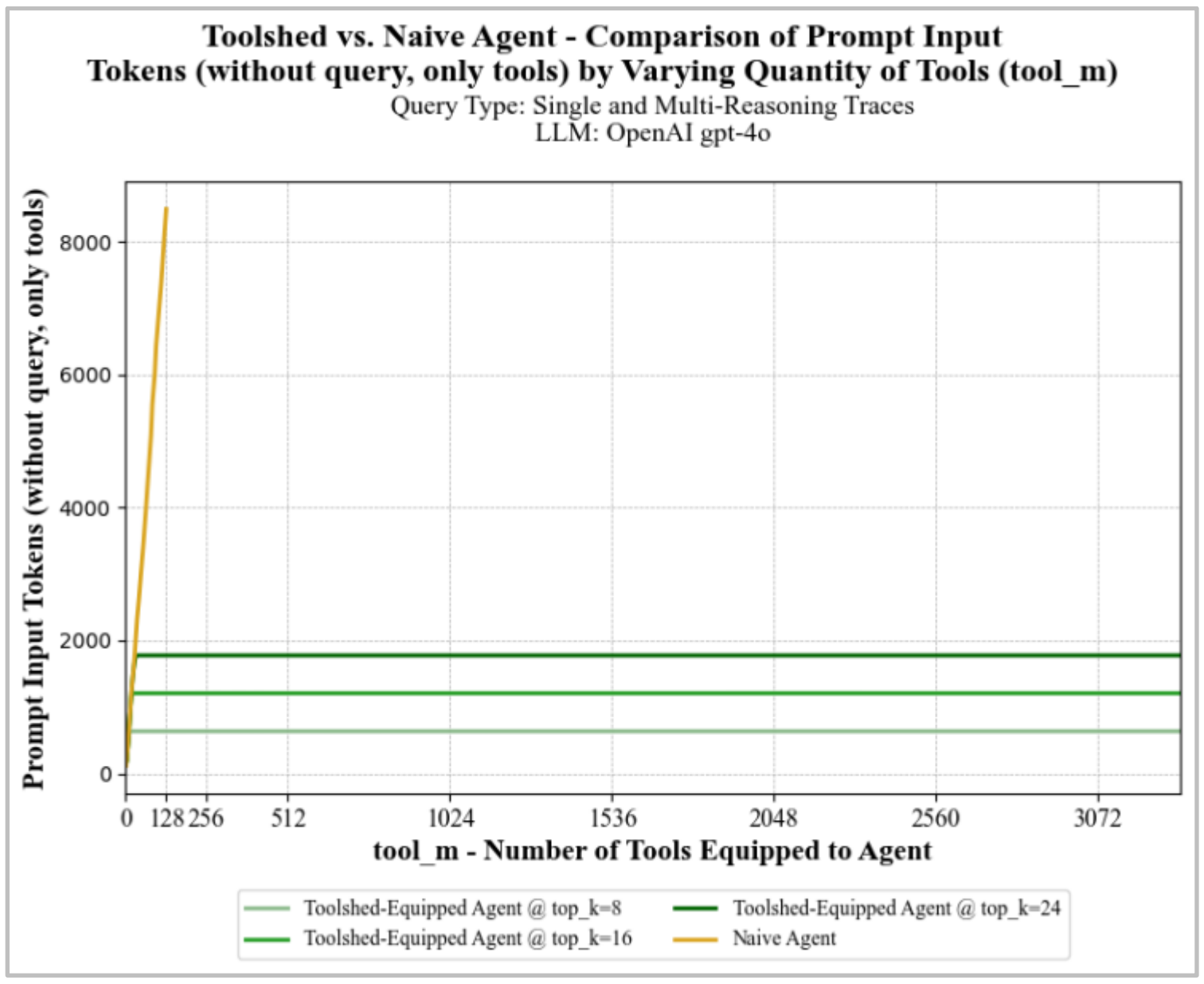

**Figure 15.** Comparing Prompt Input Tokens (less user query, only tools) for a Simple Agent with *M* number of tools equipped from 1-4,000 to an Agent using *Advanced RAG-Tool Fusion* and limiting *top-k* where the retrieval accuracy is satisfactory (90%+) to save on token costs.

**3. Hypothetical Example of Optimizing the Trade-off between Retrieval Accuracy, Agent Performance, Token Count for any Number of Tools (*tool-M*) Equipped to an Agent**

In this hypothetical data example, we examine a Simple Agent that struggles with solving user questions when equipping it more than 20 tools. In Fig. 16, we set up *Advanced RAG-Tool Fusion*'s retrieval accuracy for any given *tool-M* value. Given the equation in Appendix E, we model the Agent accuracy of *Advanced RAG-Tool Fusion* in the second graph. Furthermore, we highlight the token count trade-off, which increases linearly with more tools, unless fixed to a selection threshold (*top-k*) in *Advanced RAG-Tool Fusion*. Since Simple Agent Accuracy performs very strong (~100%) until the *tool-M* = 20, we can fix the *Advanced RAG-Tool Fusion top-k* value at any value between 1 and 20. Then, in Fig. 17, we assess the retrieval accuracy of *Advanced RAG-Tool Fusion* at *top-k* = 5, 10 and 20. While the *top-k* = 20 retrieval accuracy performs higher than the *top-k* = 10 and *top-k* = 5 retrievers, we can consider if the retrieval accuracy gain justifies the token cost for the additional tools. For whichever *top-k* we choose, we model the *Advanced RAG-Tool Fusion* agent accuracy along with the token count at each level. Overall, we demonstrate that in addition to scaling to thousands of tools, *Advanced RAG-Tool Fusion* can benefit small-scale tool-equipped Agents when they struggle with a certain number of tools, by only giving them a subset (*top-k*) of tools that we know they can reliably reason with (Simple Agent accuracy where *tool-m*=*top-k*).

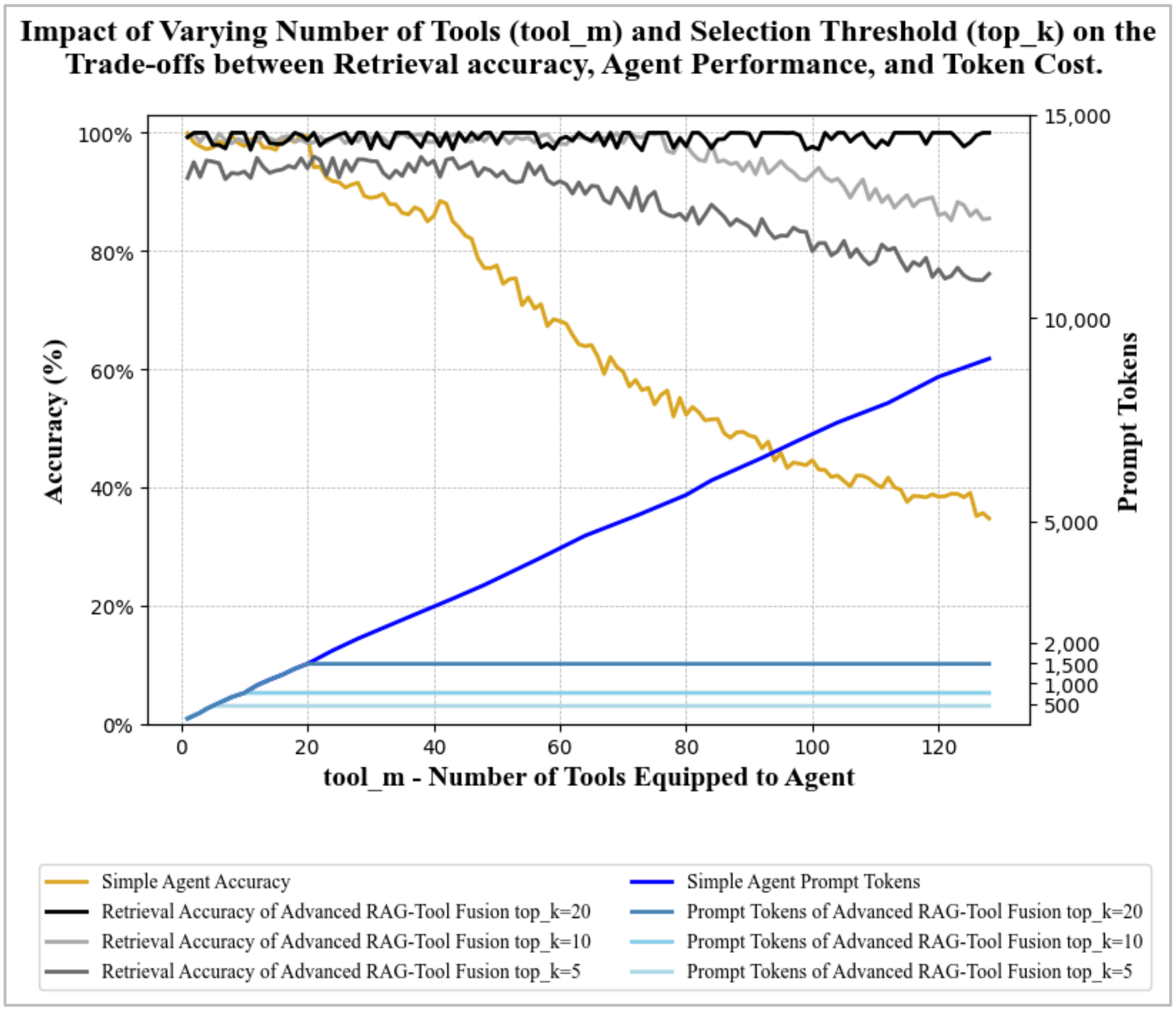

**Figure 16.** Hypothetical example to showcase the trade-off between retrieval performance, agent accuracy, and token cost, across an increasing number of tools equipped to an agent (*Tool-M*). In this figure, the *Advanced RAG-Tool Fusion* retrieval at various *top-k* values is compared to the Simple (Base) Agent Accuracy. Also, the token count (cost) is shown at various levels for a given *top-k*.

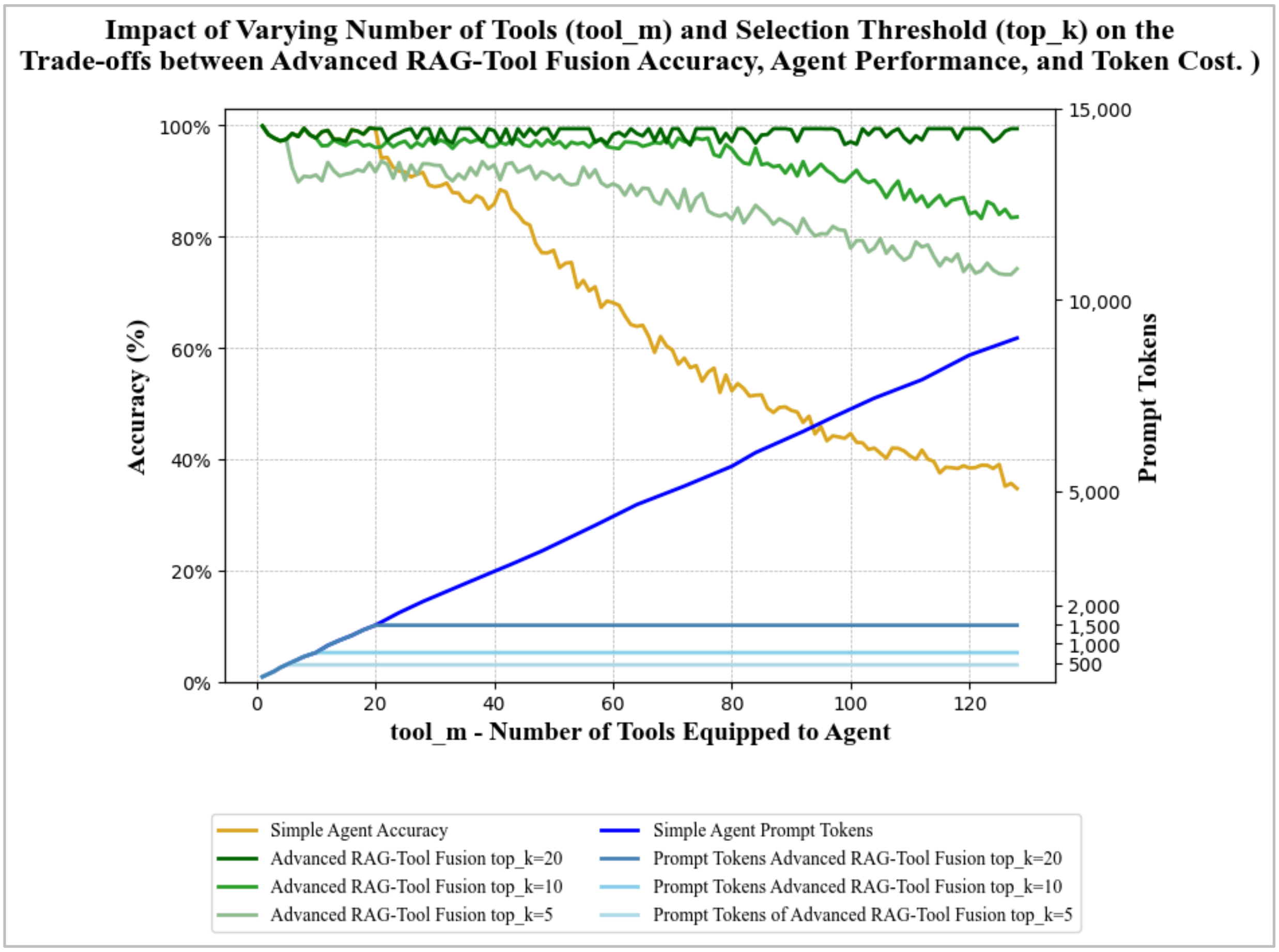

**Figure 17.** Hypothetical example to showcase the trade-off between retrieval performance, agent accuracy, and token cost, across an increasing number of tools equipped to an agent (*Tool-M*). In this figure, the *Advanced RAG-Tool Fusion* Agent Accuracy equation (Appendix E) is modeled by the given retrieval accuracy in Fig. 16 and the Simple Agent Accuracy. Additionally, the token count (cost) is shown at various levels for a given *top-k*.

## Appendix D: Multi-Query Expansion or Variation Example

In the following example, we show the application of multi-query diversification or expansion from a user question. Starting from the user question “what is a neural network,” if the Agent does not have any prior knowledge what type of tools may be in the *Toolshed Knowledge Base*, querying just the intent or user query may not result in the correct tool being retrieved. This is due to the diverse ways to answer the question. In the following example, “what is a neural network” can be answered with many possible tools, such as a web search, a Python documentation tool for neural networks, a YouTube video tool, a mentorship or tutor tool, or an online courses tool. From a retrieval standpoint, we need to ensure these diverse ways to solve the user question are accounted for. That is where multi-query diversification or expansion generates N number of variations of the user question or intent that aim to reword the question in different ways to solve it. For example, in the third example, we access tools relevant to education and tutoring. In the first example, we correctly retrieve the tools relevant to online courses. Without multi-query diversification or expansion, if we sent in the user query or intent “What is a neural network,” this would not yield the correct tool, unless the OnlineCourseTool()’s description or appended hypothetical questions or key words had keywords such as neural network.

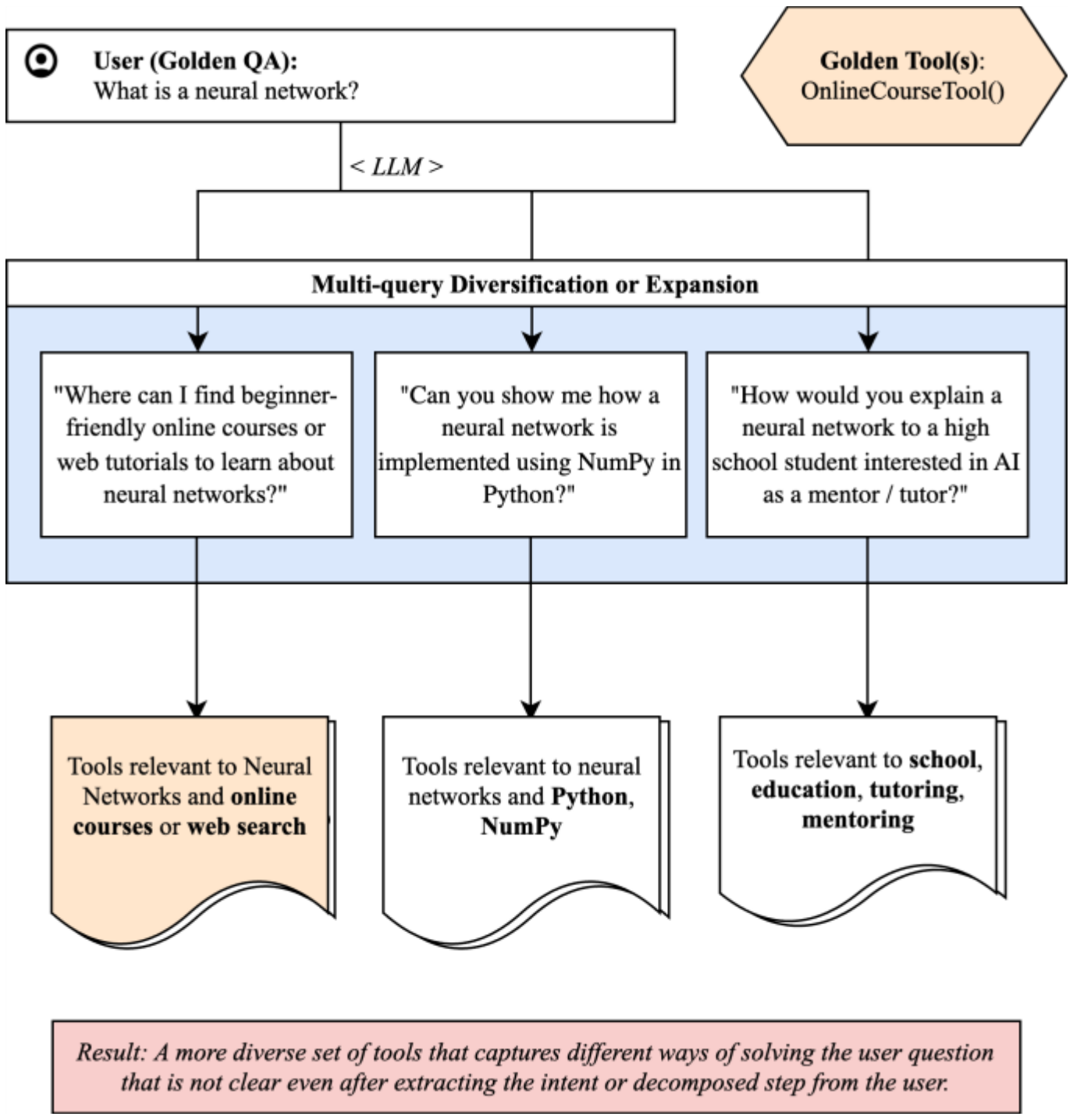

**Figure 18.** Multi-query diversification or expansion example that generates three variations of the user question or intent to tackle diverse ways of solving the user question.

## Appendix E: Advanced RAG-Tool Fusion Modeled Equation

The expected accuracy of *Advanced RAG-Tool Fusion* at any *tool-M* or *top-k* value can be modeled by the expected accuracy of a Simple Agent where its *tool-M* is equal to the *Advanced RAG-Tool Fusion*'s *top-k* value, multiplied by the expected retrieval accuracy of *Advanced RAG-Tool Fusion* at the given *tool-M* and *top-k* level. As seen in Appendix C, we are simply limiting the Simple Agent's *tool-M* by a fixed (or dynamic) *top-k* of our choosing. Doing this allows us to not only scale to thousands or millions of tools, but also allow our Simple Agent at *tool-m* = *top-k* to only focus on a small number of tools to increase its accuracy if needed, along with save on token costs.

$$\begin{aligned}\mathbf{E}[\text{Advanced RAG-Tool Fusion Agent Accuracy}(tool_m, \text{top-}k)] \quad &= \quad \mathbf{E}[\text{Simple Agent Accuracy}(tool_m = top_k)] \\ &\quad \times \mathbf{E}[\text{Advanced RAG-Tool Fusion Retrieval Accuracy}(tool_M, \text{top-}k)]\end{aligned}$$

**Figure 19.** The expected accuracy model of *Advanced RAG-Tool Fusion*, showing the relationship between *tool-M*, *top-k*, and retrieval accuracy in comparison to a Simple Agent.

## Appendix F: Types of Queries for Tool Datasets

There are two distinct types of queries: single reasoning trace queries and multi-reasoning trace queries. Simply put, single reasoning traces only require a single tool call, where multi-reasoning traces require more than a single tool call. To break down multi-reasoning traces even further, there are two distinct types: parallel queries and sequential queries. Parallel multi-reasoning traces occur when the multiple tool calls are independent from each other and can be executed at the same time. Model providers such as OpenAI have fine-tuned their models to output parallel tool calls when it deems necessary [3]. The other type of multi-reasoning trace query is sequential queries. These scenarios occur when the output of the first tool call is needed before it can execute the second tool. As seen below in the sequential example, the weather tool must be executed first. After the LLM Agent reasons that it is in fact a sunny day, it will use the reservation tool to make a reservation. This requires more advanced reasoning capabilities of the LLM, such as ReAct or CoT.

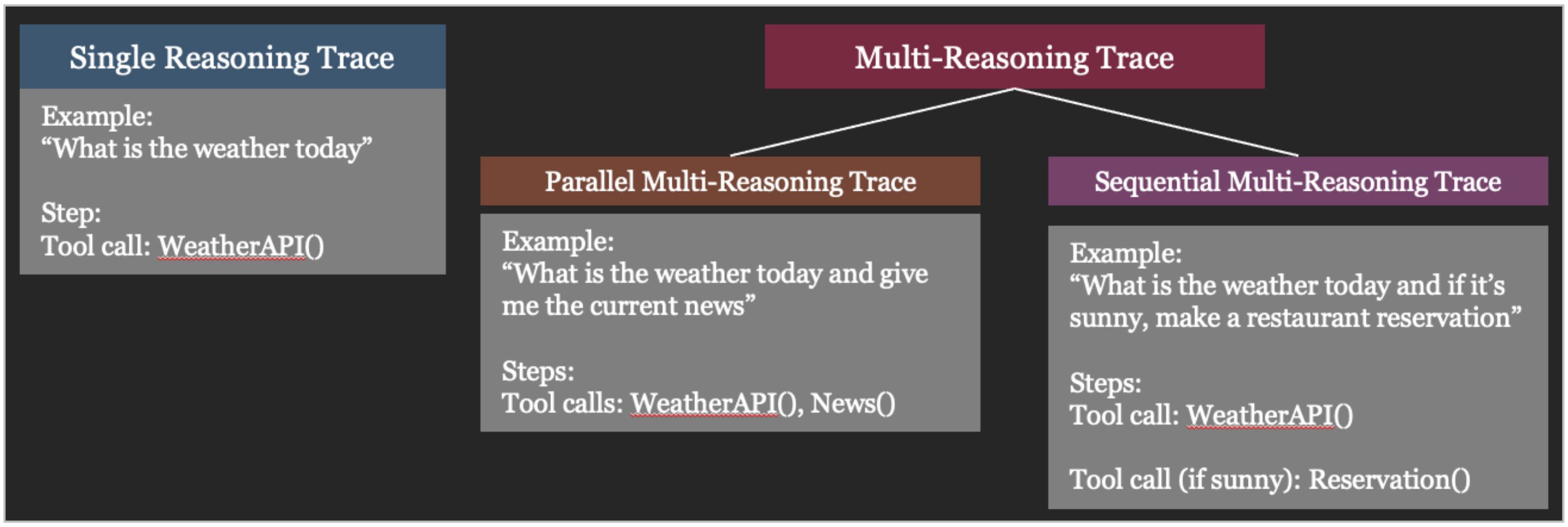

**Figure 20.** Types of queries in a tool dataset. Single-reasoning traces (single tool call) and multi-reasoning traces (multiple reasoning traces). Multi-reasoning traces can be further broken into parallel and sequential queries. While parallel multi-reasoning traces are independent and can be executed at the same time, sequential multi-reasoning traces need to be executed and reasoned with step-by-step.

## Appendix G: OpenAI Python Tool or Function Definition Documentation Example Schema

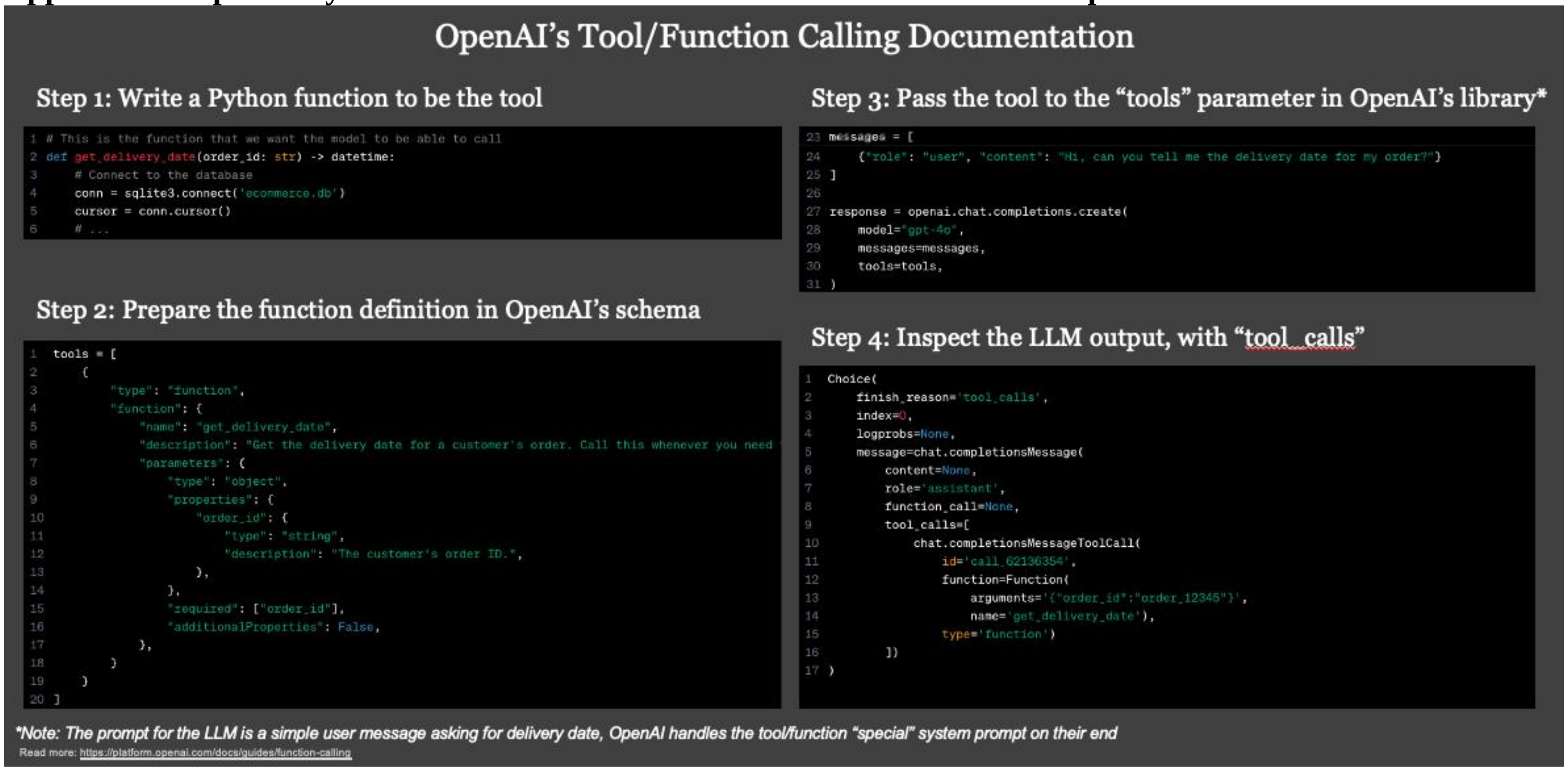

**Figure 21.** OpenAI Tool calling documentation

## Appendix H: Example Prompts Utilized for Advanced RAG-Tool Fusion

The following LLM prompts are examples of how to orchestrate prompting for each module in *Advanced RAG-Tool Fusion*. These prompts provide a high-level idea on how to formalize each module as an input and output. However, in real-world production tool datasets, it is encouraged to give additional instructions and backstory about the tool dataset as well as few-shot examples of the desired input and output. For example, when building an *Advanced RAG-Tool Fusion* pipeline for 1,000 financial tools, in the system prompts, you can give the LLM context on being an expert in financial tools and ensure each output is unique. In the same example, adding few-shot examples specific to the expected financial questions (for decomposing queries or multi-query expansion) will be more effective than a simple prompt with no context. Thus, the following example prompts can be fine-tuned for the specific use-case in production. Furthermore, when utilizing structured outputs to parse JSON schemas from the LLM output, we recommend first having the LLM generate its thoughts, approach, and an unstructured output for the question, followed by an additional LLM call to transform the unstructured output into a JSON structured format. This two-step approach is known to increase reasoning accuracy by allowing the LLM to fully lay its thoughts and answer out before worrying about being confined to a JSON schema.

reverse_hyde

```
system_message="""You are an expert at generating hypothetical questions that a given python function can answer.
You are given a name of a function description of the function along with various example questions that can be answered by using the function.
Your job is to do the following:
2. Generate 10 additional example questions that the function can answer. Do not repeat any of the given example questions, but they can be similar.

Take a deep breath and think through the problem step by step."""

human_message="FUNCTION NAME: `{tool_name}`\nFUNCTION DESCRIPTION: `{tool_description}`\n\nEXAMPLE QUESTIONS:\n```{example_questions}```\n\n10 ADDITIONAL QUESTIONS:"

```

**Figure 20.** Example prompt to generate hypothetical questions to append to each tool document in the *Toolshed Knowledge Base*, during the pre-retrieval phase in *Advanced RAG-Tool Fusion*. As stated previously, this base prompt can be augmented with few-shot examples of expected hypothetical questions based on a specific domain.

key_topics

````
system_message="""You are an expert at generating key topics from a list of questions that a python function can answer.
You are given a name of a function description of the function along with various example questions that can be answered by using the function.
Your job is to do the following:
1. Generate a list of five key topics, each 1-5 words long, that capture the overarching theme and topic of the following questions.

As an example:
EXAMPLE QUESTIONS:
```
"I'm attending a summer wedding, and I'd love to wear something floral and elegant. Any recommendations?"
"I have a job interview in the tech industry. What outfit should I wear to look both professional and modern?"
"Can you suggest an outfit for a winter date night that is both warm and stylish?"
"I need an outfit idea for a brunch with friends this weekend. Any suggestions?"
```
EXAMPLE KEY TOPICS:
```
[
'Event-Specific Fashion',
'Seasonal Outfit Ideas',
'Elegant and Glamorous',
'Casual Chic',
'Professional Attire'
]
```

Take a deep breath and think through the problem step by step."""

human_message="FUNCTION NAME: `{tool_name}`\nFUNCTION DESCRIPTION: `{tool_description}`\n\nEXAMPLE QUESTIONS:\n```{example_questions}```\n\n5 KEY TOPICS THAT CAPTURE THE OVERARCHING THEME AND TOPIC:"
````

**Figure 21.** Example prompt to generate key topics, intents, or themes to append to each tool document in the *Toolshed Knowledge Base*, during the pre-retrieval phase in *Advanced RAG-Tool Fusion*. As stated previously, this base prompt can be augmented with few-shot examples of expected key topics, intents, or themes based on a specific domain.

decomposition

```
system_message="""You are an expert at breaking down user questions into clearly defined step(s).
You will be given a user question that can be answered by a single action or multiple actions.
However, you do not know the actions ahead of time.
For some questions, the user may be asking for a single thing, which can be broken down into one step. You can remove redundant information such as 'can you help me with that,' 'please,' etc. Just keep the main question.
For other questions, the user may be asking for multiple things, which can be broken down into multiple steps. These will be clearly separate questions, often marked by two question marks '?' or conjunctions like 'and,' 'Additionally,' 'Also,' etc.

Your job is to do the following:
1. Break down the following question into clearly defined steps.
2. Always be as clear as possible. If it is a multi-step question, each step should contain a clear independent action.
3. For multi-step questions, you can break them down into 2-4+ reasoning steps, depending on the complexity of the request.

Example of single-step questions:
---------
EX. QUESTION:
What's the weather forecast for tomorrow in New York City?
EX. STEP(S):
['What's the weather forecast for tomorrow in New York City?']
---------
EX. QUESTION:
Can you calculate the net present value (NPV) of an investment with a 5% discount rate and projected cash flows of $1000 per year for 5 years?
EX. STEP(S):
['Calculate the net present value (NPV) of an investment with a 5% discount rate and projected cash flows of $1000 per year for 5 years.']
---------
EX. QUESTION:
Can you find me a restaurant reservation for tonight in downtown Boston?
EX. STEP(S):
['Find me a restaurant reservation for tonight in downtown Boston.']
---------
Example of multi-step questions (2+ steps):
---------
EX. QUESTION:
What's the net present value (NPV) and internal rate of return (IRR) for a project with an initial investment of $5000, projected annual cash inflows of $1200 for 7 years, and a 6% discount rate?
EX. STEP(S):
['Calculate the net present value (NPV) for the project with the given details.', 'Calculate the internal rate of return (IRR) for the project.']
---------
EX. QUESTION:
Can you summarize the Federal Reserve's latest policy, analyze its impact on inflation, and provide insights into stock market trends?
EX. STEP(S):
['Summarize the Federal Reserve's latest policy.', 'Analyze the impact of the policy on inflation.', 'Provide insights into stock market trends.']
---------
EX. QUESTION:
I'm planning a trip to Italy. Can you suggest the best cities to visit, create an itinerary, help me book flights, and recommend good places to stay?
EX. STEP(S):
['Suggest the best cities to visit in Italy.', 'Create an itinerary for my trip.', 'Help me book flights to Italy.', 'Recommend good places to stay in those cities.']
---------
EX. QUESTION:
Can you research the latest trends in electric vehicle (EV) manufacturing, identify the top 5 companies, provide their stock prices, and suggest investment strategies for EV stocks?
EX. STEP(S):
['Research the latest trends in electric vehicle (EV) manufacturing.', 'Identify the top 5 companies in the EV industry.', 'Provide the stock prices of these companies.', 'Suggest investment strategies for EV stocks.']
---------

Take a deep breath and think through the problem step by step."""

human_message="USER QUESTION: {query}\nSTEPS FOR THAT QUERY:"
```

**Figure 20.** Example prompt to decompose the user question into decomposed independent steps or user intents, during the intra-retrieval phase in *Advanced RAG-Tool Fusion*. To handle multi-hop queries, it is critical to break down the multi-step user question into independent sub-tasks or user intents. If the application only requires single-reasoning trace questions, the query decomposition module is not necessary. As stated previously, for query decomposition or user intent decomposition, adding few-shot examples into the system prompt related to the specific tool dataset's expected user questions (e.g. financial tools and questions) will increase decomposition accuracy. If the module does not successfully break down the user question into correct sub-tasks, other modules—such as multi-query expansion and variation, post-retrieval reranking, corrective RAG, and self-RAG—will increase the likelihood of correcting the failed decomposition. For complex sequential user questions, prompting frameworks such as ReAct can sequentially solve each sub-task in a loop, prompting frameworks that handle the decomposition in a single step will save latency and enable parallel tool invocations. The limitation, as stated in Section 7, is that for inherently sequential user questions—where the output of the first tool call feeds into subsequent tool inputs—prompt tuning or ReAct prompting may improve overall end-to-end accuracy.

prompt.sty

````
You are an expert at converting user questions to 3 sentence variations that target different keywords and nuanced approaches with the goal to embed this query in a
vector database to retrieve relevant tools across various industries.
Your goal is to craft 3 nuanced sentence variations that target different aspects of understanding or solving the query.
For example, one sentence could focus on a detailed aspect of the user query, while another is more broad to cover more ground when embedding
these sentences to retrieve the most relevant tools for the user query.
As an example, if the user is "Could you explain how quantum computing works?", you could craft the following sentences to deal with specific technical aspects of quantum computing, general research or study resources, and then another sentence for learning or tutorials. These 3 sentences will aim to retrieve a different scope of tools to cover more ground.
Example user question: "Could you explain how quantum computing works?"
```
Example 3 sentences:
1. "Conduct a detailed analysis of quantum computing, combining scholarly articles, tutorials, and expert insights to provide a well-rounded understanding of its core principles and algorithms."
2. "Offer an interactive learning experience to explain the fundamentals of quantum computing, using guided tutorials, quizzes, and simulations to help users grasp its key concepts."
3. "Search for and compile web-based resources, videos, and articles that break down the complexities of quantum computing into easy-to-understand explanations for beginners."
```
Before you start, understand this from a practical standpoint: The user question can be matched to a range of tools or solutions within the system, and your crafted variations should optimize for breadth and specificity.
Write out your approach and plan for tackling this, then provide the 3 sentences you would craft for the user question.
Think through your approach step by step, be intelligent, take a deep breath.
USER QUESTION: {user_question}
YOUR APPROACH , REASONING , AND 3 SENTENCES:
````

**Figure 21.** Example prompt to generate multi-query expansion or variation, during the intra-retrieval phase in *Advanced RAG-Tool Fusion*. This base prompt can be fine-tuned to a specific dataset in a domain, such as finance, healthcare, or education. The few-shot examples can be updated to match the use-case to.

reranker

```
finalized_list_thoughts = """OK here are the results:
USER QUESTION EMBEDDED AND RETRIEVED TOOLS:
{user_question_results}
SENTENCE 1 EMBEDDED AND RETRIEVED TOOLS:
{sentence_1_results}
SENTENCE 2 EMBEDDED AND RETRIEVED TOOLS:
{sentence_2_results}
SENTENCE 3 EMBEDDED AND RETRIEVED TOOLS:
{sentence_3_results}
=========
Based on these results, rank the top 5 most relevant tools to solve the user question. Just return the 5 tool names for each relevant tool.
"""
```

**Figure 23.** Example prompt to rerank the tool documents from each expanded or varied sub-query, during the intra-retrieval phase in *Advanced RAG-Tool Fusion*. A reranking embedder can also be utilized instead of an LLM. This prompt is an example of how to rerank tools from an LLM. The pipeline that uses this prompt is more complex with structured outputs and tool name validations. The assumed variables are the retrieved tools from each of the multi-query expansion or variation module.

decomposed_reranker

```
multiple_decomposed_queries_reranker="""You are an expert at combining and narrowing down the top tools from each user intent to a single unique list of 5 tools that solve the user question.
You will be given a user query that has been broken down into 2-4 distinct user intents.
You are also given the 5 most relevant tools for each intent that can solve that particular intent, which ARE IN ORDER OF RELEVANCE!
Your task is to combine the top tools from each intent into a single unique list of 5 tools that are most relevant to the user question, which can solve the entire multi-step process.

The rules for combining tools:
1. Start by selecting the top tool from each intent.
2. If fewer than 5 tools are selected after adding the top tool from each intent, add the second most relevant tool from each intent, one at a time, until the list reaches 5 tools.
3. If there is an overlap of tools between the intents, count the tool only once and move to the next relevant tool in that intent's list.

Here are some examples with no overlapping tools:
---------
USER QUESTION: 'I need to evaluate different real estate markets, assess financial risk, and identify eco-friendly property investments.'
INTENT 1: 'Evaluate the top real estate markets.'
LIST OF TOOLS FOR INTENT 1: ['RealEstateAnalyzer', 'MarketTrendsExplorer', 'GeoDataSearch', 'InvestmentFinder', 'DataScanner']
INTENT 2: 'Assess the financial risks of investing in these markets.'
LIST OF TOOLS FOR INTENT 2: ['RiskAssessor', 'FinancialTracker', 'RiskMapAnalyzer', 'FinanceExpert', 'DataWizard']
INTENT 3: 'Identify eco-friendly property investments.'
LIST OF TOOLS FOR INTENT 3: ['GreenInvestmentTool', 'EcoFriendlyFinder', 'SustainableInvestments', 'ClimateIndex', 'EnvironmentalAnalyzer']
THE APPROACH TO TAKE:
`First, select the top tool from each intent: 'RealEstateAnalyzer', 'RiskAssessor', 'GreenInvestmentTool'. This gives us 3 tools. Next, go back to Intent 1 and add 'MarketTrendsExplorer', and to Intent 2 and add 'FinancialTracker'. Now we have a unique list of 5 tools.`

---------
USER QUESTION: 'Can you find data on electric vehicle growth, compare manufacturing costs, recommend top companies, and suggest potential partnerships?'
INTENT 1: 'Retrieve data on electric vehicle market growth.'
LIST OF TOOLS FOR INTENT 1: ['EVDataAnalyzer', 'IndustryMonitor', 'GrowthTrends', 'MarketInsightTool', 'DataCollector']
INTENT 2: 'Compare manufacturing costs for electric vehicles.'
LIST OF TOOLS FOR INTENT 2: ['CostCalculator', 'PriceComparisonTool', 'EVManufacturingMetrics', 'FinancialInsight', 'CostFinder']
INTENT 3: 'Recommend the top electric vehicle companies.'
LIST OF TOOLS FOR INTENT 3: ['TopEVCompanies', 'EVIndustryReport', 'CompanyRanker', 'StockWatcher', 'BusinessIntelligenceTool']
INTENT 4: 'Suggest potential partnerships in the electric vehicle sector.'
LIST OF TOOLS FOR INTENT 4: ['PartnerSearch', 'CollaborationFinder', 'IndustryConnections', 'EVPartnerTracker', 'OpportunityScanner']
THE APPROACH TO TAKE:
`First, take the top tool from each intent: 'EVDataAnalyzer', 'CostCalculator', 'TopEVCompanies', 'PartnerSearch'. This gives us 4 tools. Next, go back to Intent 1 and add 'IndustryMonitor'. Now we have a unique list of 5 tools.`

---------
Here is an example with overlapping tools:
---------
USER QUESTION: 'I need to forecast market trends, analyze financial health, visualize data, and create a report.'
INTENT 1: 'Forecast upcoming market trends.'
LIST OF TOOLS FOR INTENT 1: ['MarketTrendsForecast', 'StockInsightTool', 'FinancialPrediction', 'DataTracker', 'MarketScout']
INTENT 2: 'Analyze the financial health of key companies.'
LIST OF TOOLS FOR INTENT 2: ['FinancialHealthAnalyzer', 'CompanyPerformanceTool', 'RiskEvaluator', 'DataInsightPro', 'ProfitTracker']
INTENT 3: 'Visualize market data trends over time.'
LIST OF TOOLS FOR INTENT 3: ['DataVisualizationTool', 'MarketTrendsDashboard', 'GraphBuilder', 'ChartMaster', 'TrendMapper']
INTENT 4: 'Create a report summarizing the findings.'
LIST OF TOOLS FOR INTENT 4: ['ReportBuilder', 'DocumentCreator', 'DataSummaryTool', 'ReportWriter', 'AnalyticsPublisher']
THE APPROACH TO TAKE:
`First, take the top tool from each intent: 'MarketTrendsForecast', 'FinancialHealthAnalyzer', 'DataVisualizationTool', 'ReportBuilder'. This gives us 4 tools. Next, go back to Intent 2 and add 'CompanyPerformanceTool'. Now we have a unique list of 5 tools.`

---------
YOUR TURN:
"""

human_combiner_prompt = """USER QUESTION: '{user_question}'
{''.join([f"INTENT {i+1}: '{final_list_of_intents[i]}'\\nLIST OF TOOLS FOR INTENT {i+1}: {final_list_of_list_of_tools[i]}\\n" for i in range(len(final_list_of_intents))])}
THE APPROACH TO TAKE AND 5 FINAL UNIQUE TOOLS:"""
```

**Figure 24.** Example prompt to rerank the tool documents from each decomposed query's final *top-k* documents in Figure 21, during the intra-retrieval phase in *Advanced RAG-Tool Fusion*. A reranking embedder can also be utilized here. This step is not needed for single-reasoning traces where there is only a single decomposed query. This prompt is an example of how to rerank tools from an LLM. The pipeline that uses this prompt is more complex with structured outputs and tool name validations. The assumed variables are the retrieved tools from each of the multi-query expansion or variation module.

## Appendix I: Different Embedders and Toolshed Knowledge Base Configurations Impact Advanced RAG-Tool Fusion Retrieval Accuracy of Varying Total Tools (*tool-M*) and Tool Selection Threshold (*top-k)*

The following graphs detail the impact on retrieval for variations in Azure OpenAI embedders and Toolshed Knowledge Base Configurations. These configurations are how the tool document is stored: 1) tool name and tool description, 2) tool name, tool description, and argument schema, 3) tool name, tool description, 1 few shot query, 4) tool name, tool description, argument schema, 1 few shot query, 5) tool name, tool description, 2 few shot queries, and 6) tool name, tool description, argument schema, 2 few shot queries. Additionally, we vary two modular approaches within *Advanced RAG-Tool Fusion*: 1) With query decomposition and 2) Without query decomposition. Furthermore, we inspect the differences of query type: 1) single reasoning trace, 2) multi-reasoning trace (parallel), and 3) multi-reasoning trace (sequential). For the results, since parallel and sequential reasoning traces performed similarly (both multi-reasoning traces need query decomposition for high retrieval accuracy), we only showcase parallel results. While there are additional variations upon these configurations, such as varying the number of few shot questions from 1-10 or appending 1-10 key topics/intents/themes of the tool, the general trend is that the more enhanced the tool document is, the higher the retrieval accuracy. However, for this dataset, we noticed that the argument schema was a critical element to include, whereas the hypothetical queries did not add much value. Nevertheless, with a different dataset such as the ToolE dataset, the hypothetical questions were critical in increasing accuracy. Therefore, we highly encourage researchers and industry practitioners to test different tool document configurations on a small sample golden dataset for their use case. Then, implement the Toolshed Knowledge Base with the optimal configuration for your tool dataset.

**Configuration 1A:** Embedder: OpenAI text-embedding-3-large
Toolshed Knowledge Base Configuration: Tool Name, Tool Description

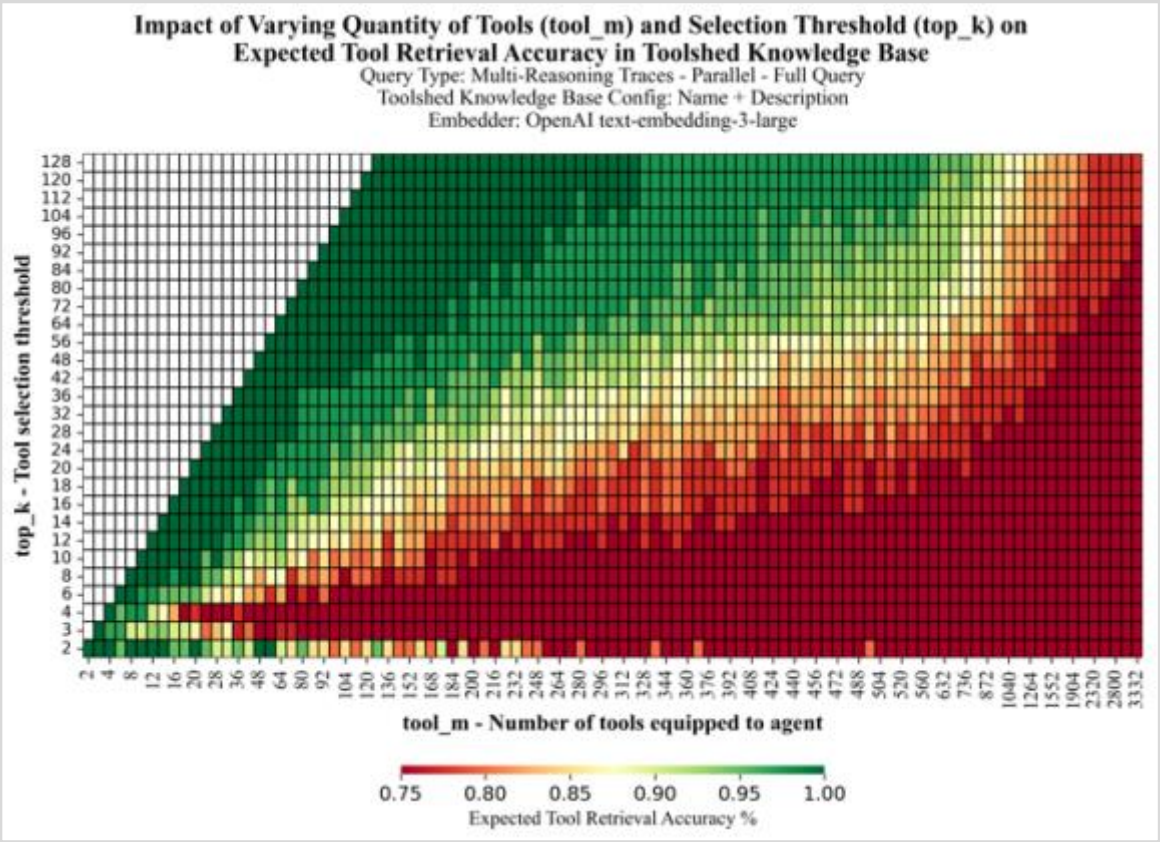

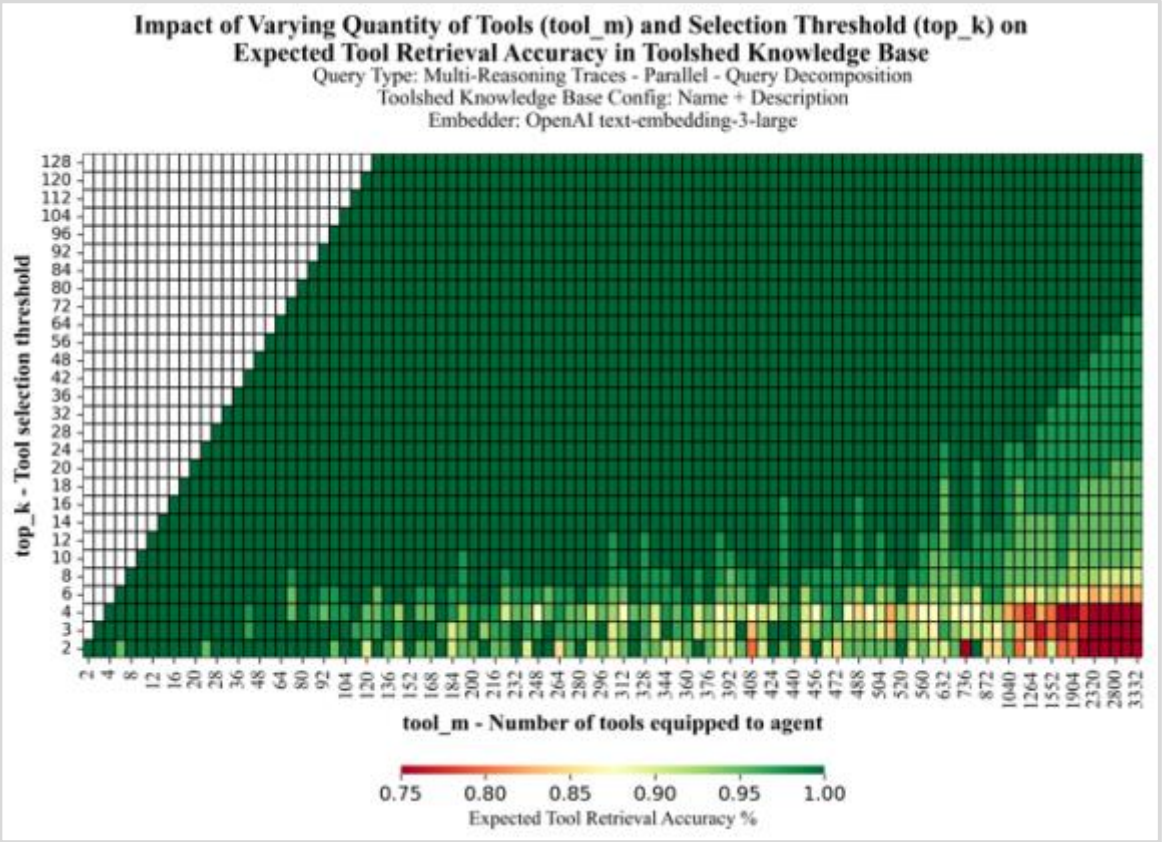

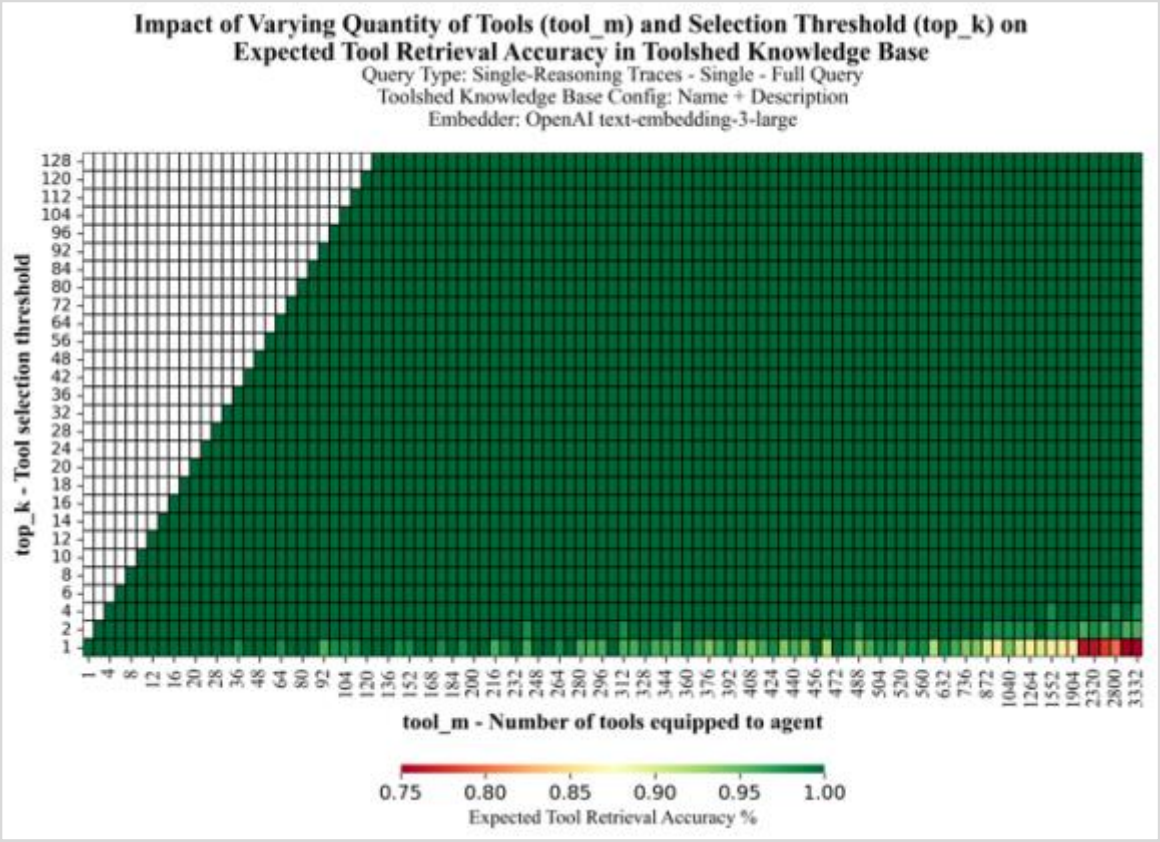

**Configuration 1B:** Embedder: OpenAI text-embedding-3-small
Toolshed Knowledge Base Configuration: Tool Name, Tool Description

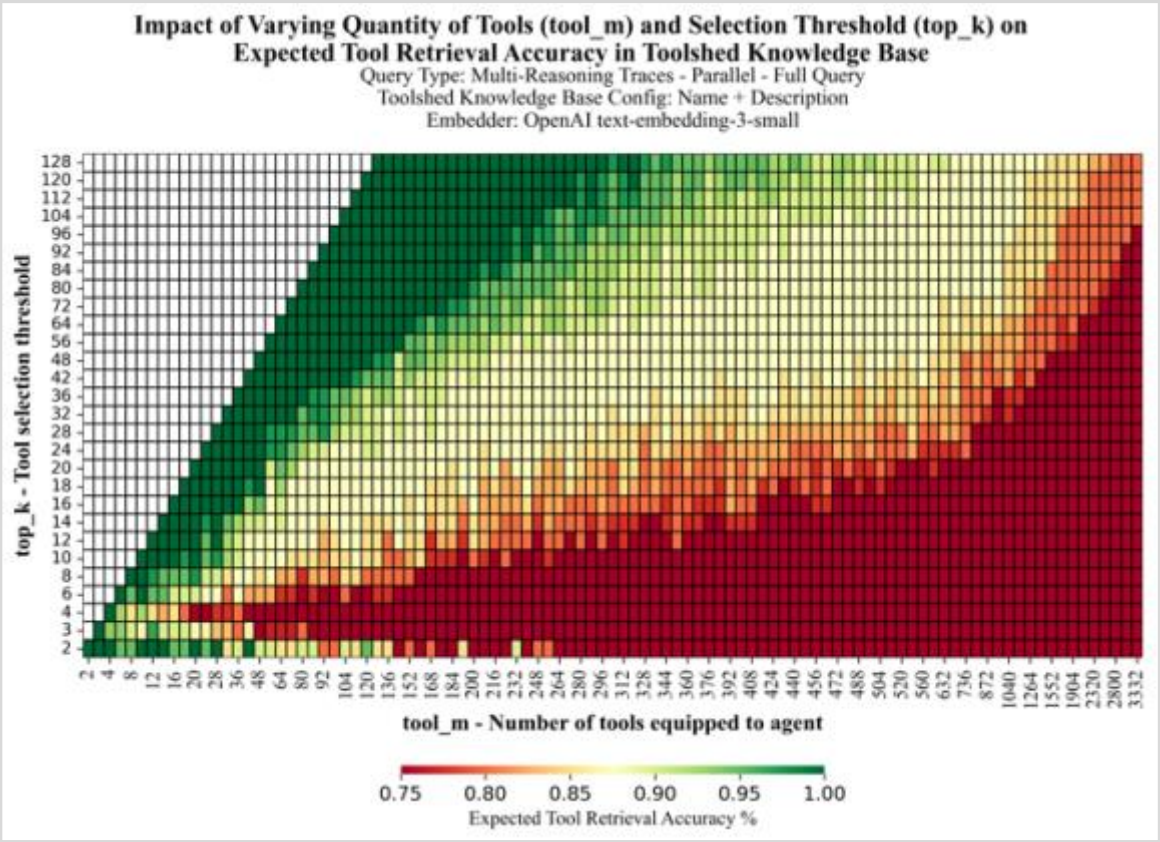

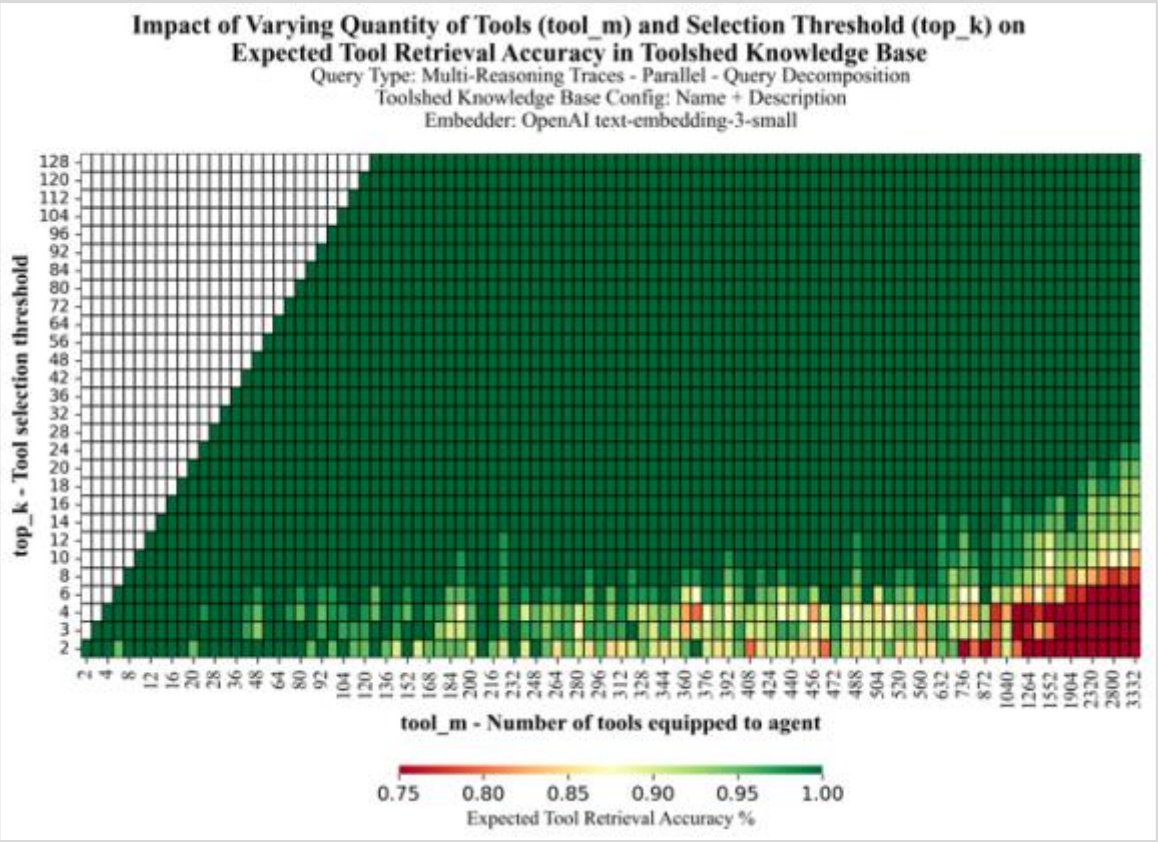

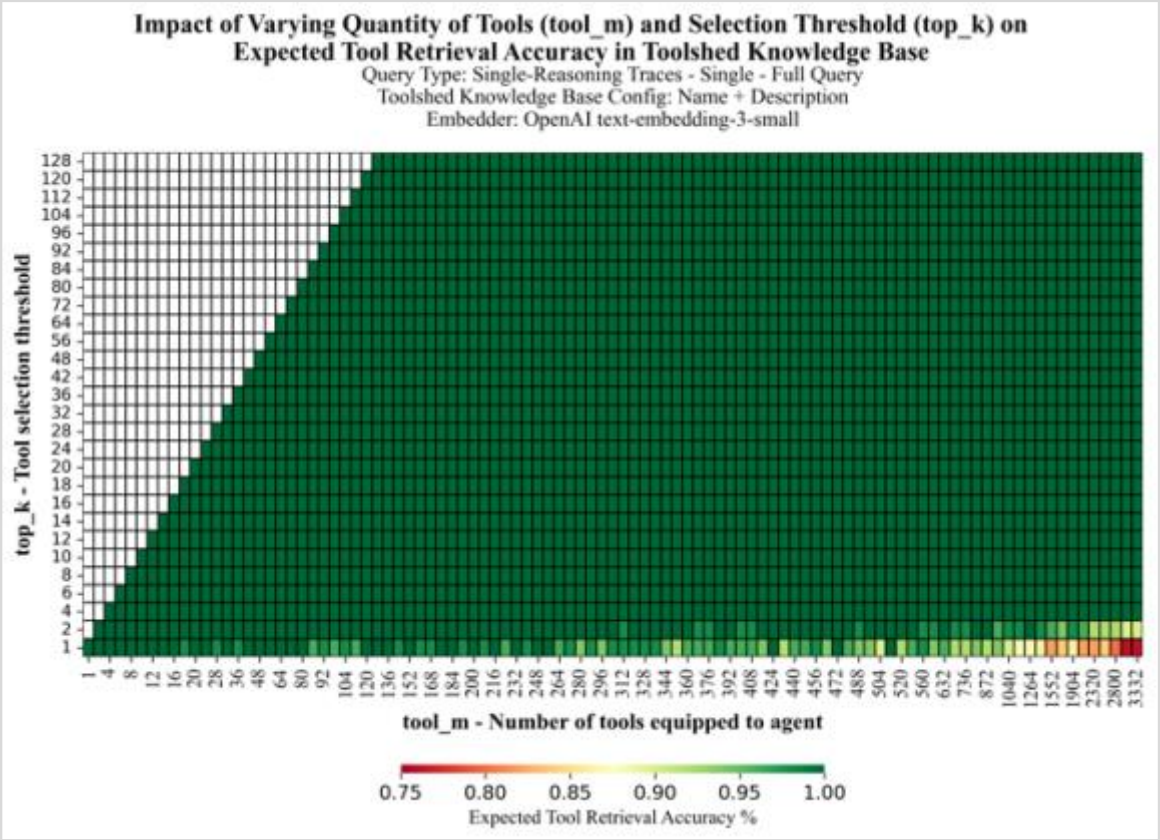

**Configuration 1C:** Embedder: OpenAI text-embedding-ada-002
Toolshed Knowledge Base Configuration: Tool Name, Tool Description

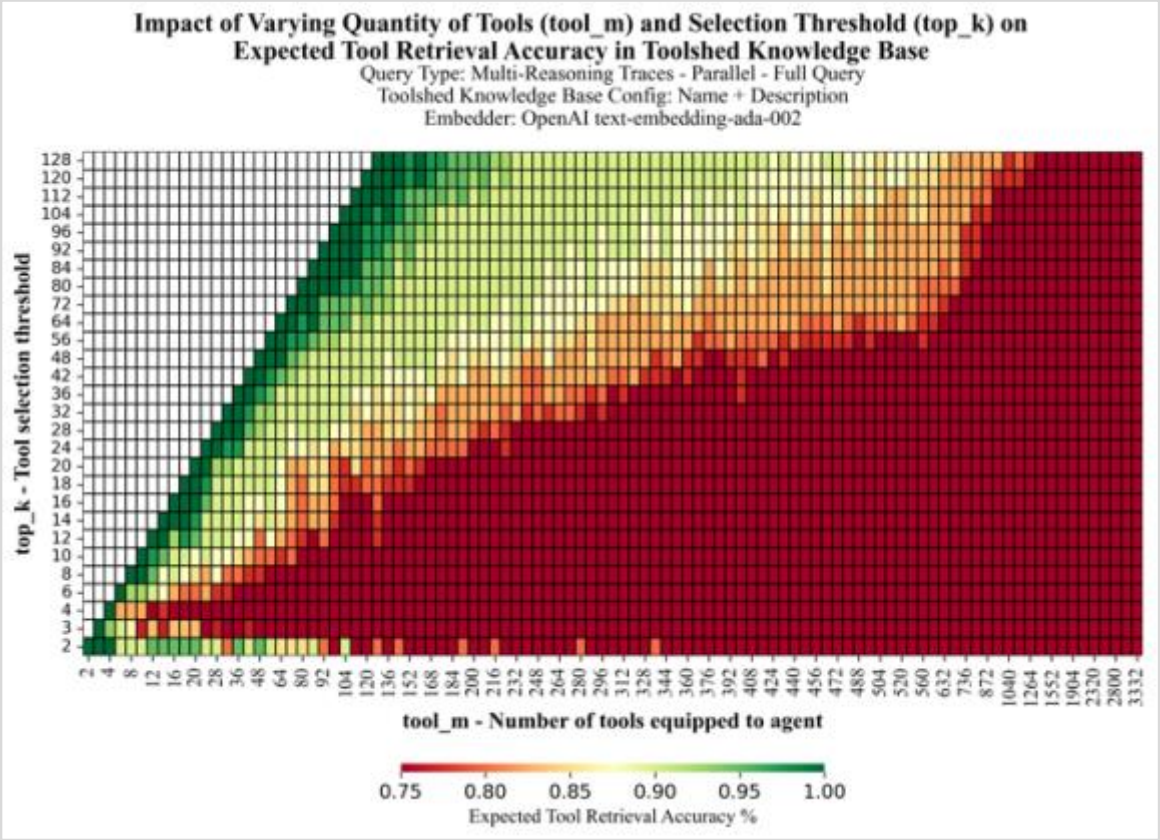

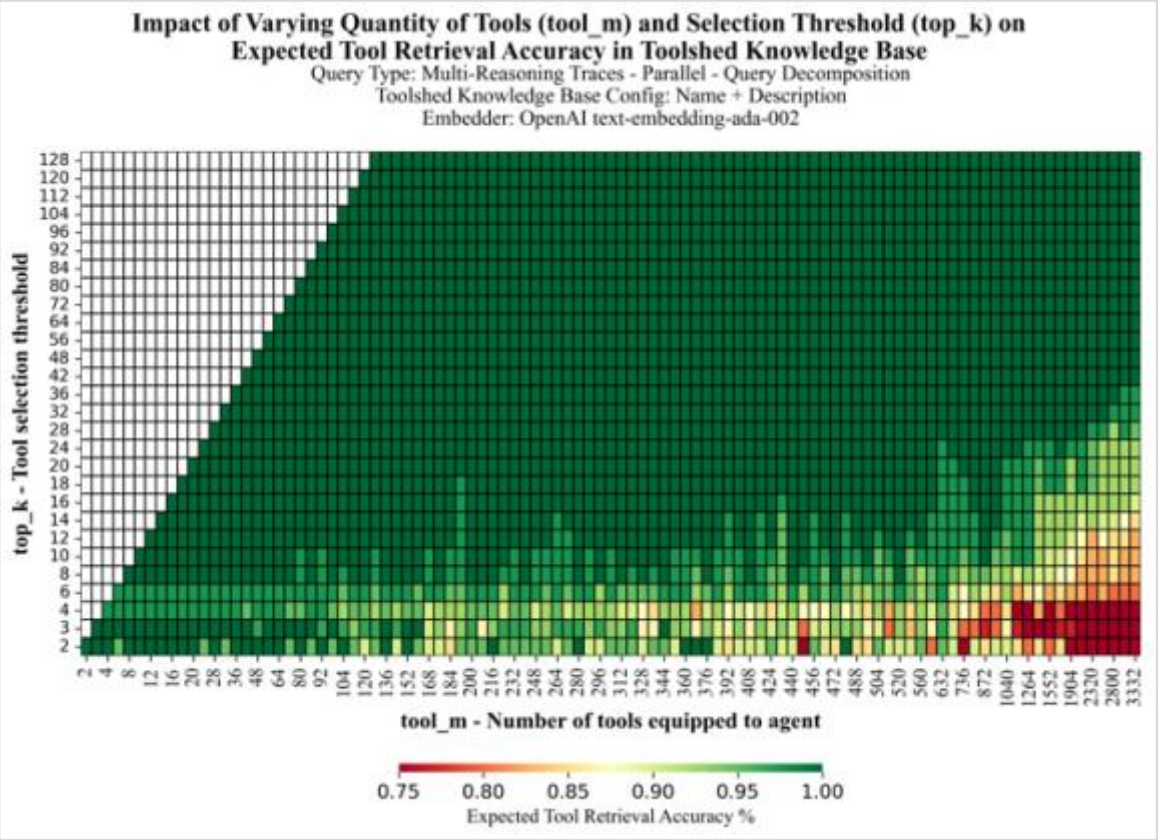

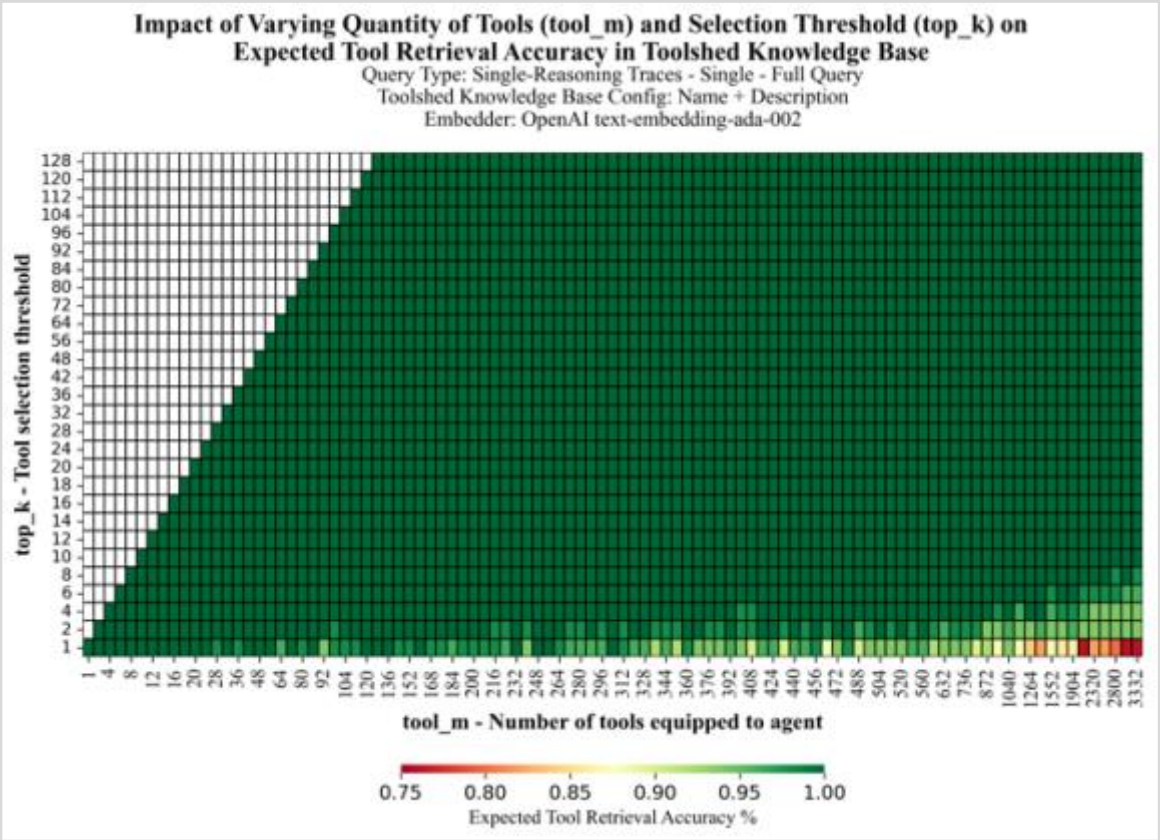

**Configuration 2A:** Embedder: OpenAI text-embedding-3-large
Toolshed Knowledge Base Configuration: Tool Name, Tool Description, Argument Schema

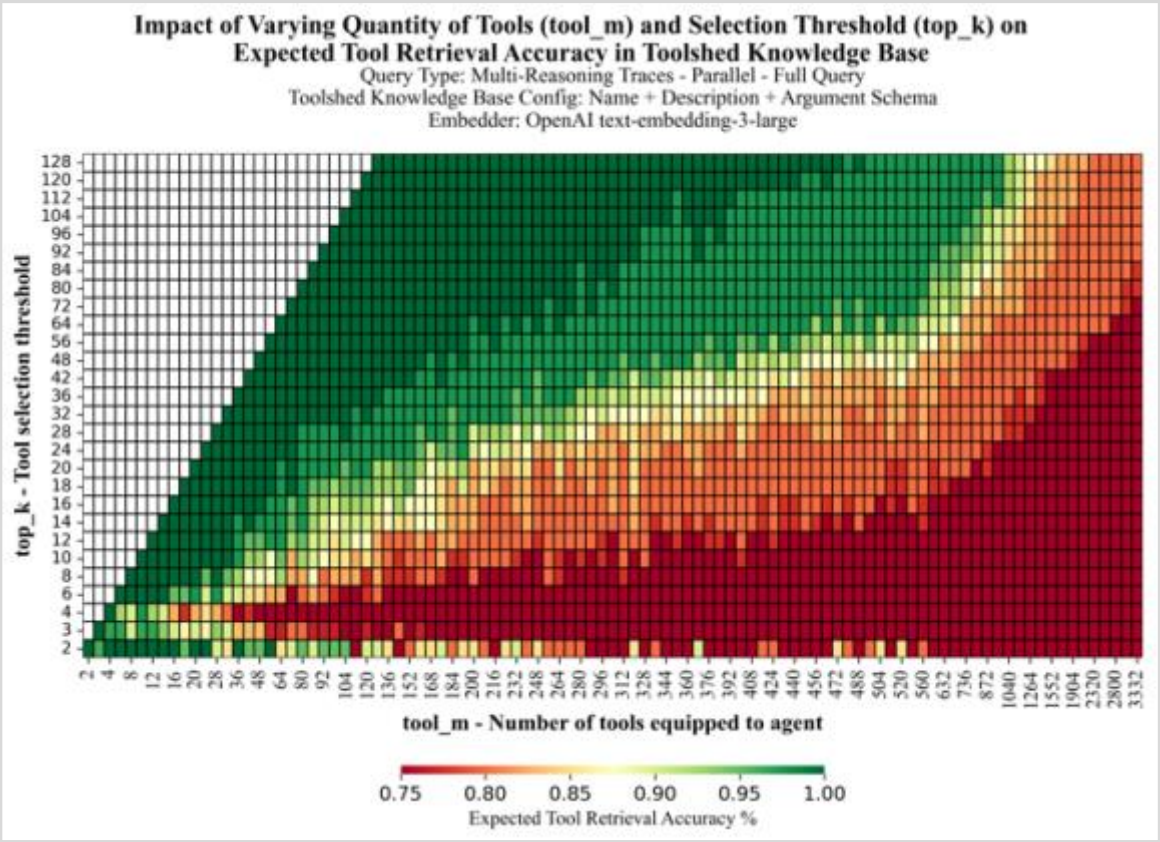

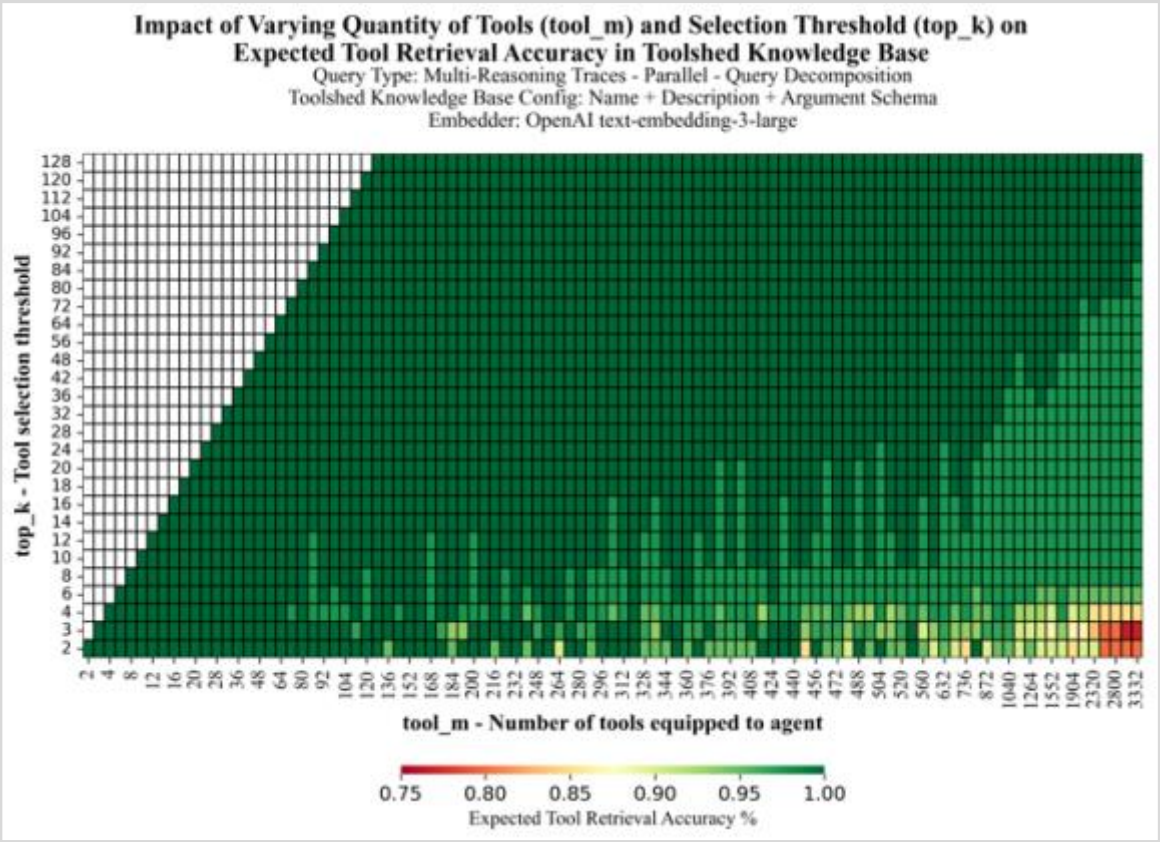

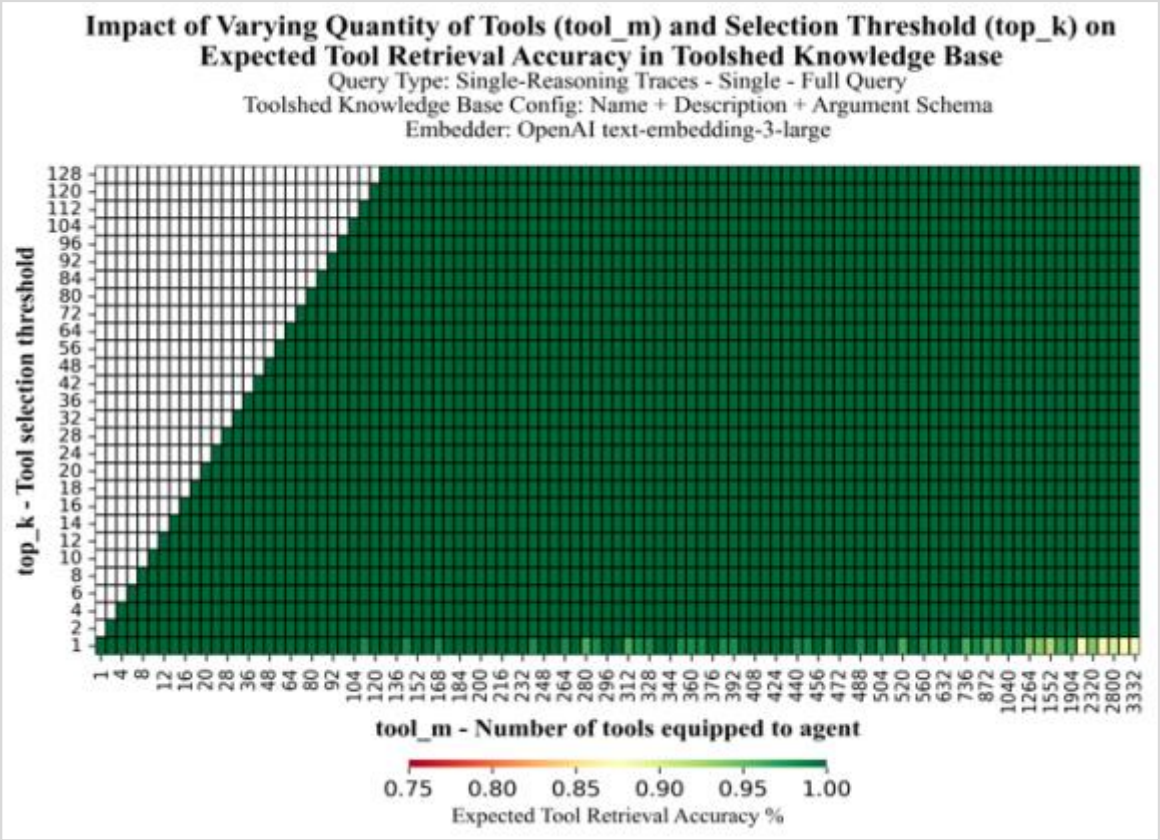

**Configuration 2B:** Embedder: OpenAI text-embedding-3-small
Toolshed Knowledge Base Configuration: Tool Name, Tool Description, Argument Schema

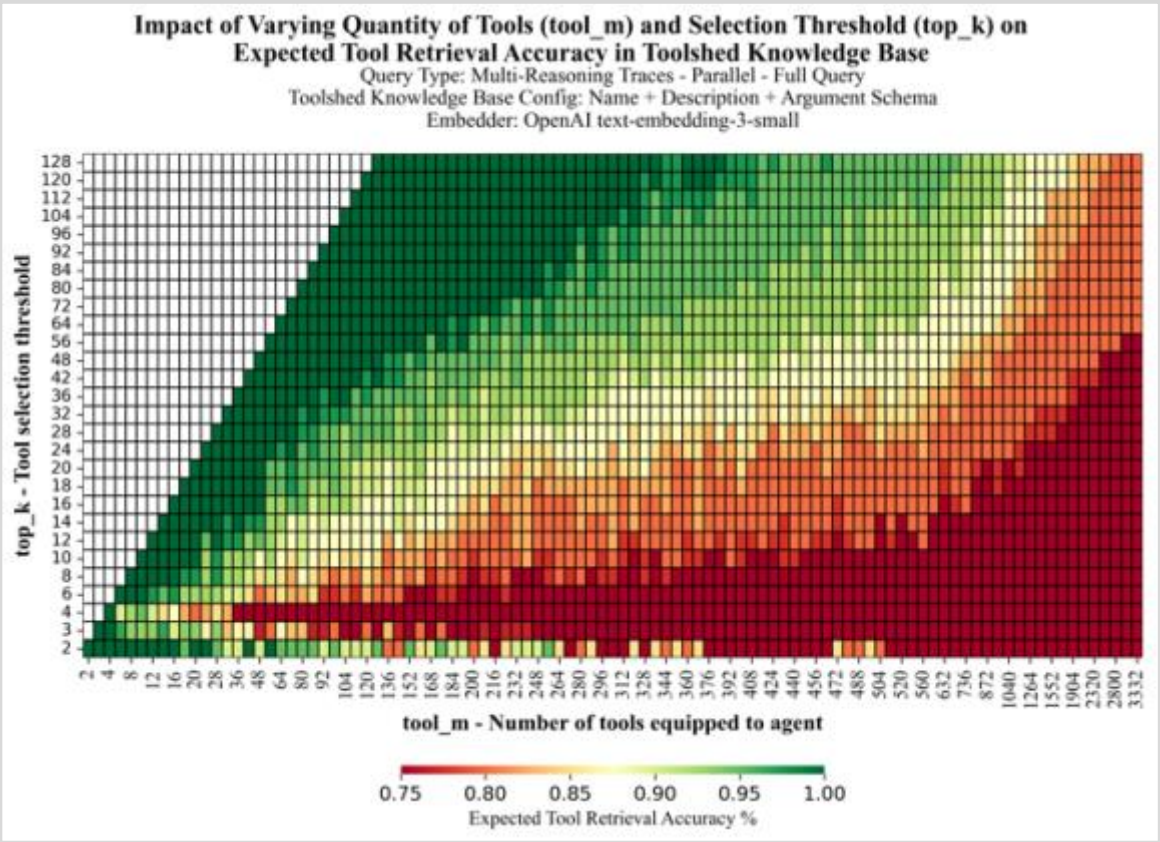

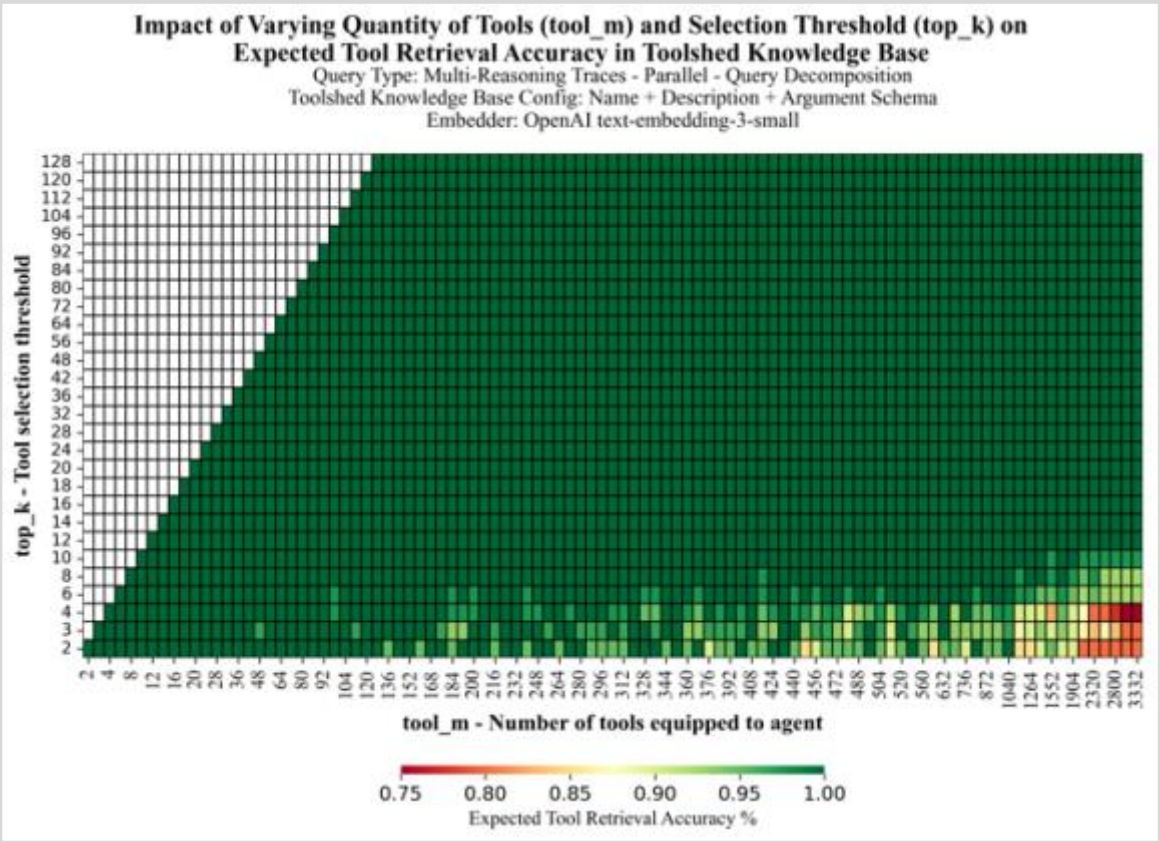

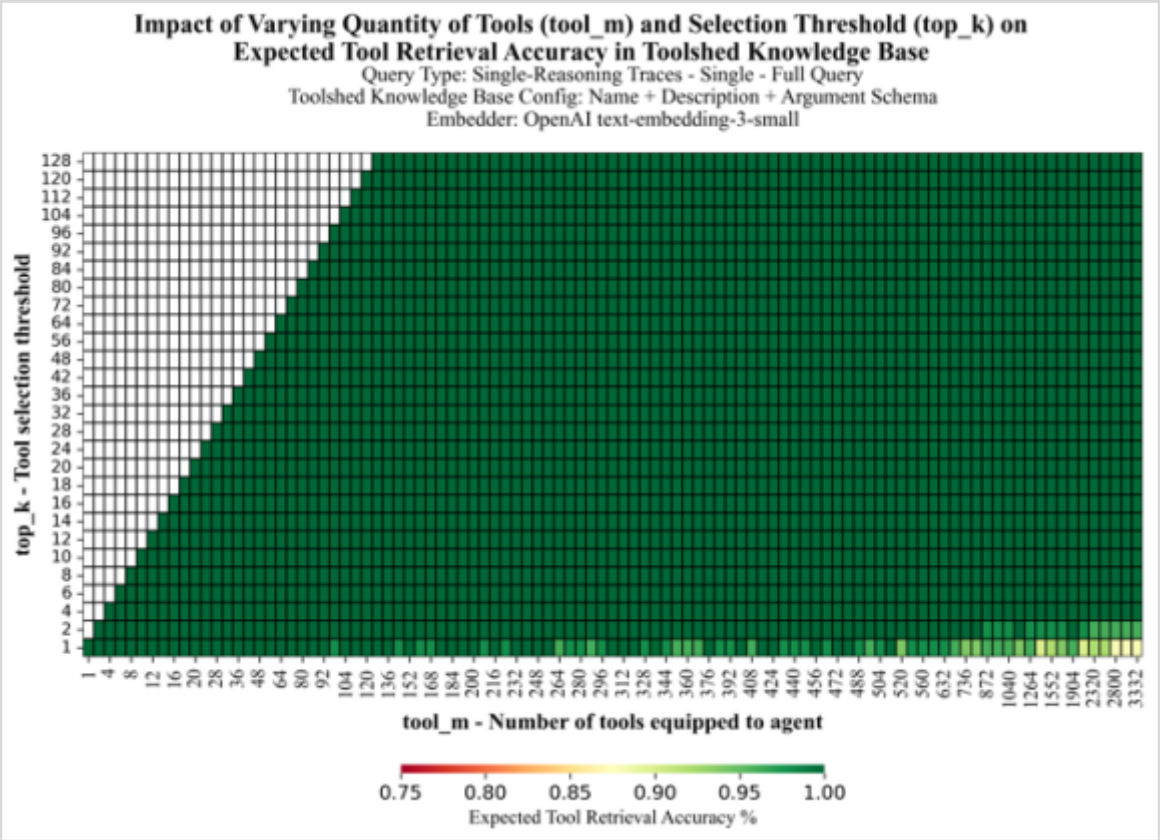

**Configuration 2C:** Embedder: OpenAI text-embedding-ada-002
Toolshed Knowledge Base Configuration: Tool Name, Tool Description, Argument Schema

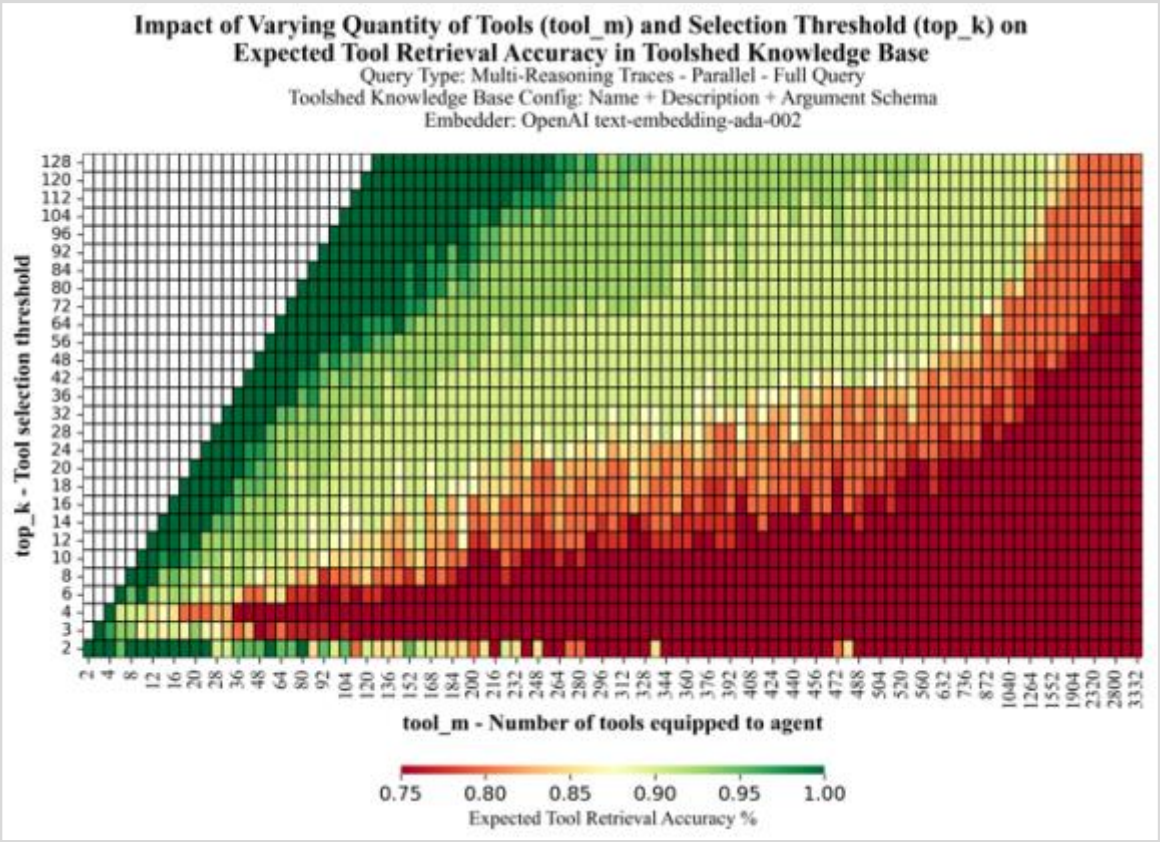

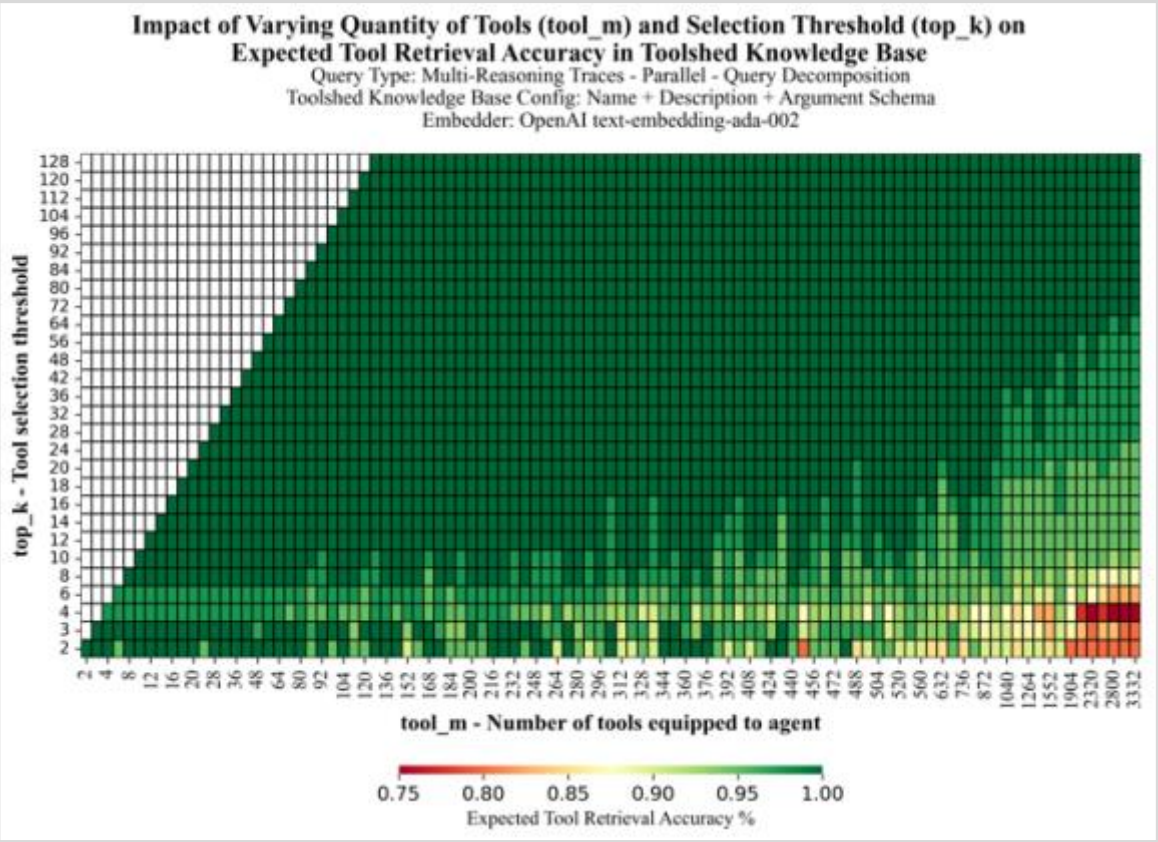

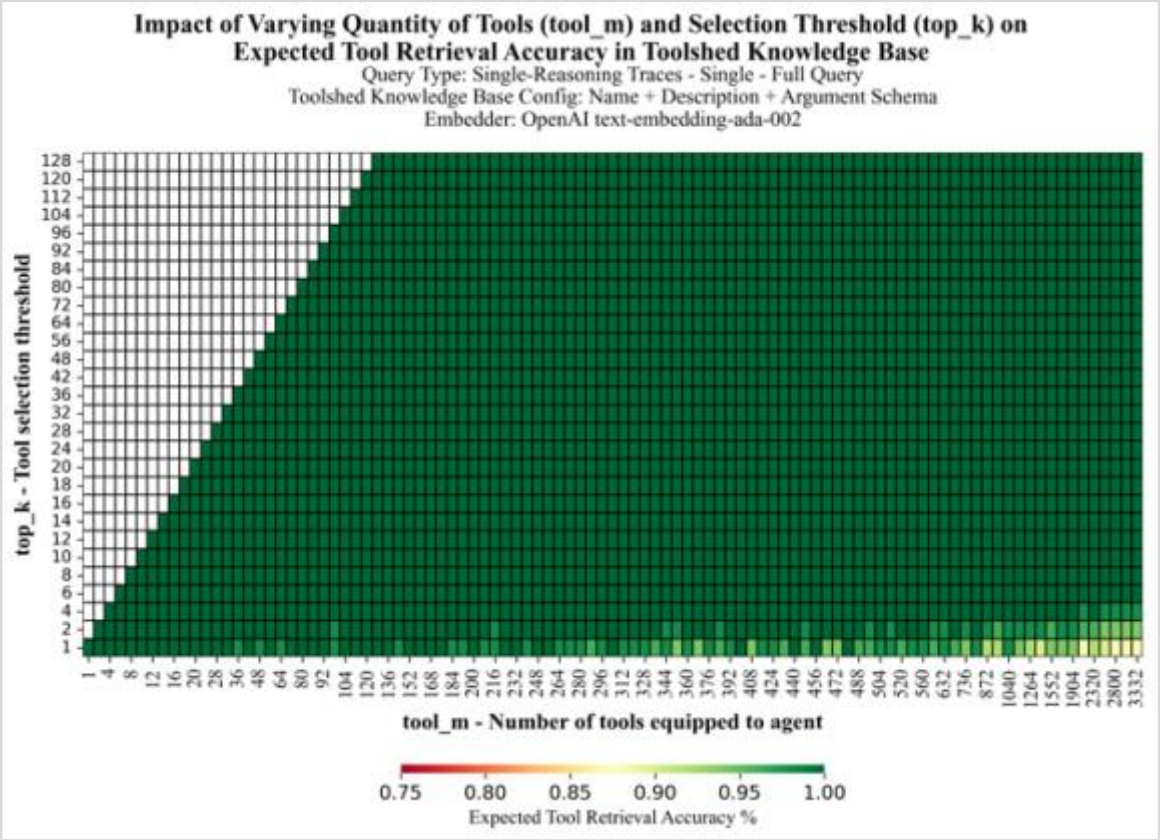

**Configuration 3A:** Embedder: OpenAI text-embedding-3-large
Toolshed Knowledge Base Configuration: Tool Name, Tool Description, 1 Hypothetical Query

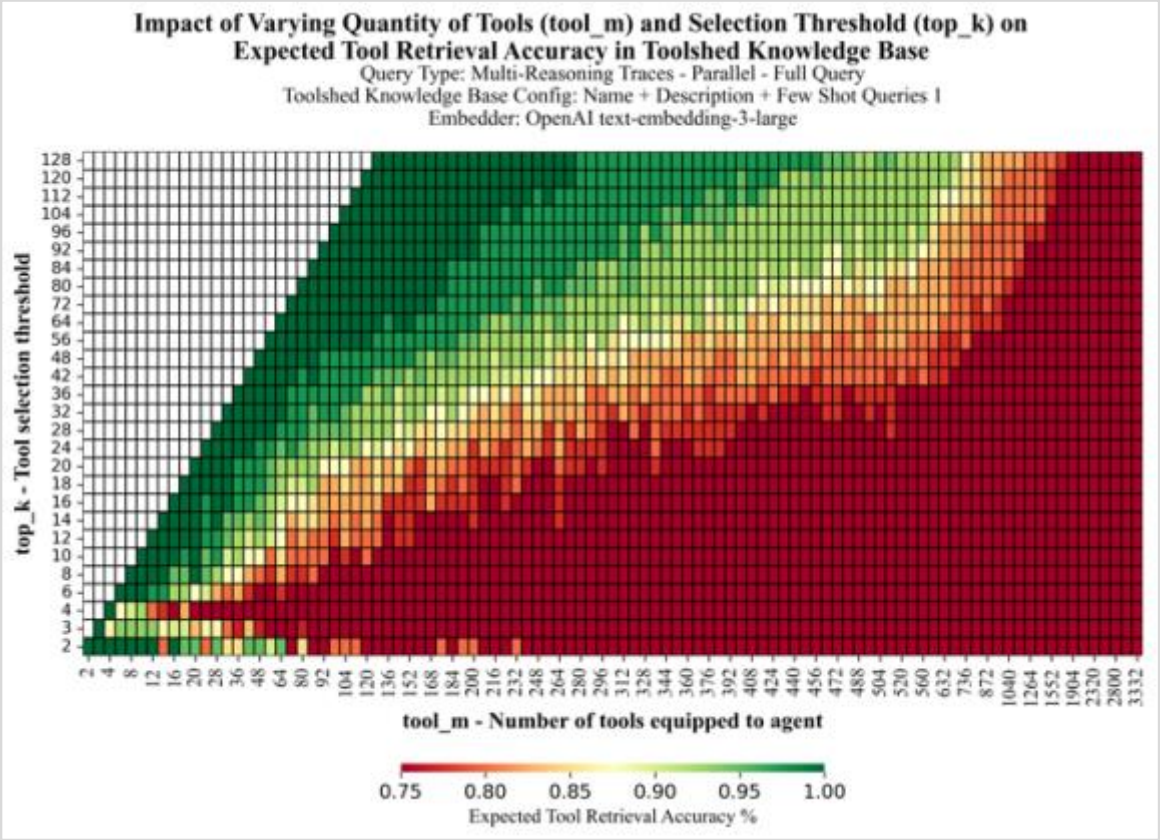

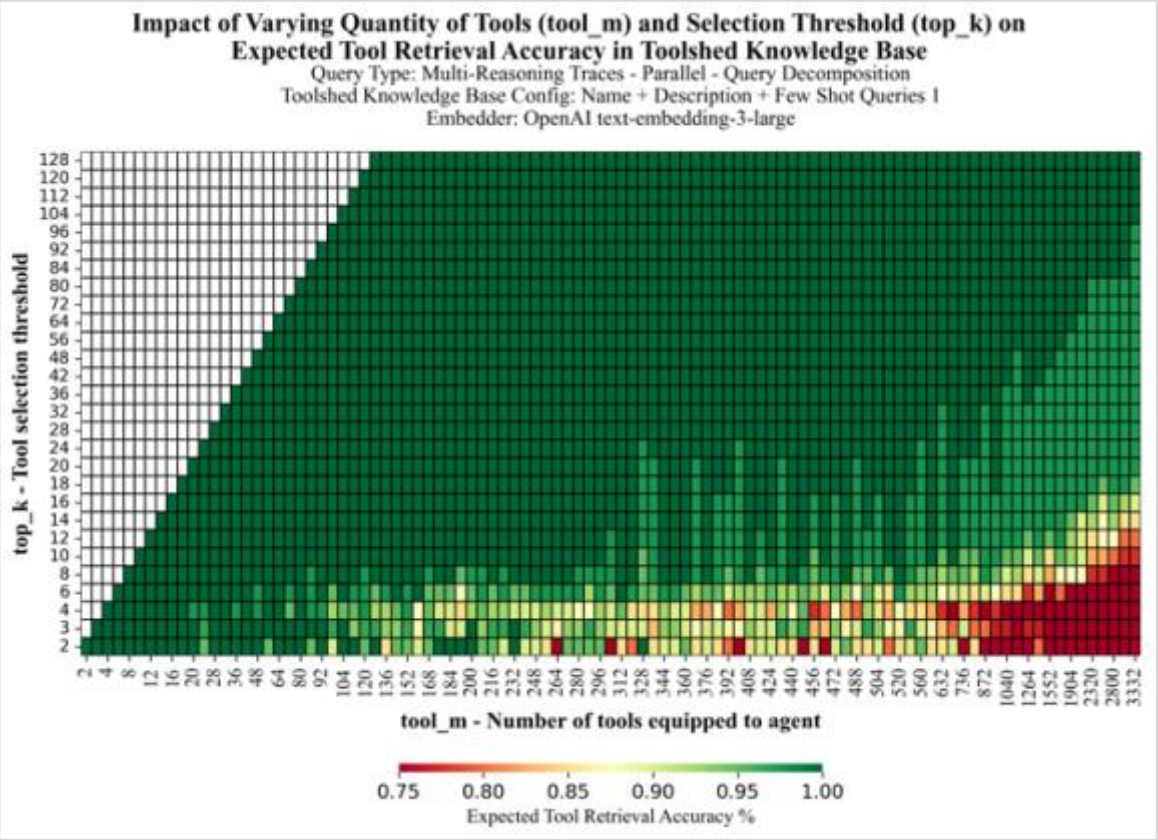

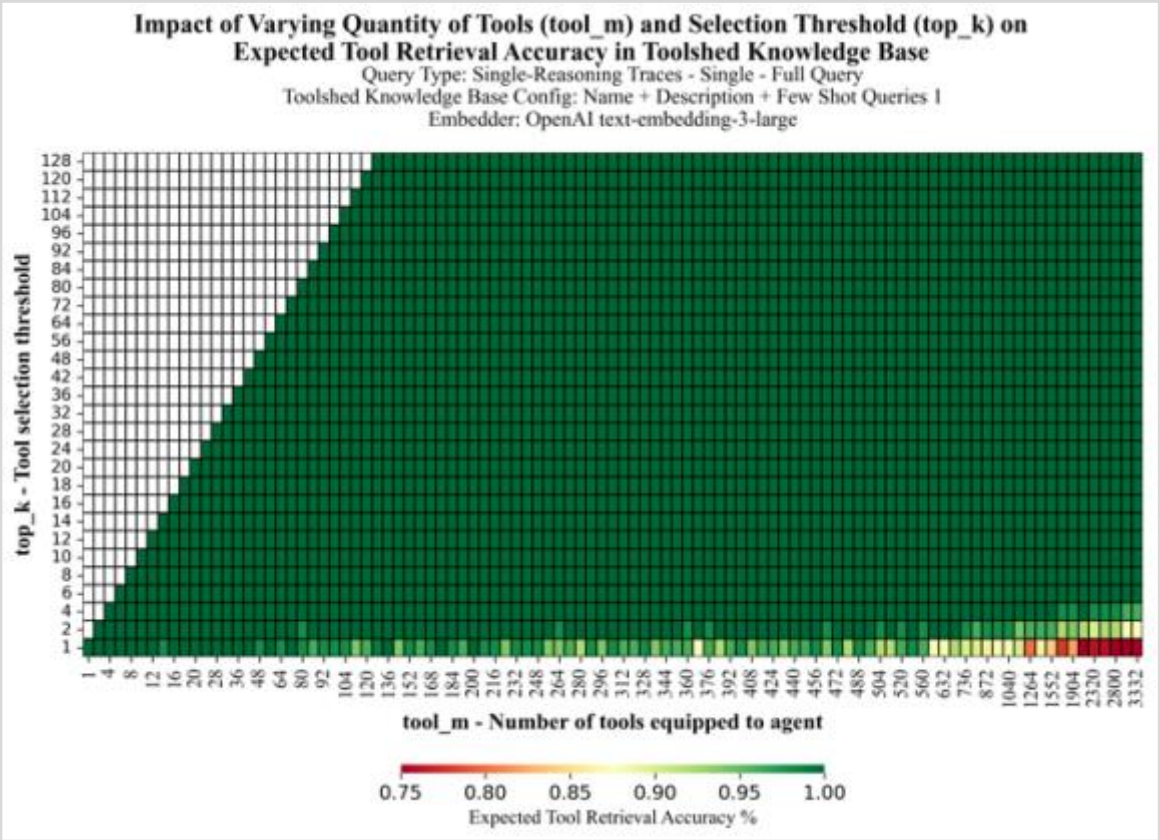

**Configuration 3B:** Embedder: OpenAI text-embedding-3-small
Toolshed Knowledge Base Configuration: Tool Name, Tool Description, 1 Hypothetical Query

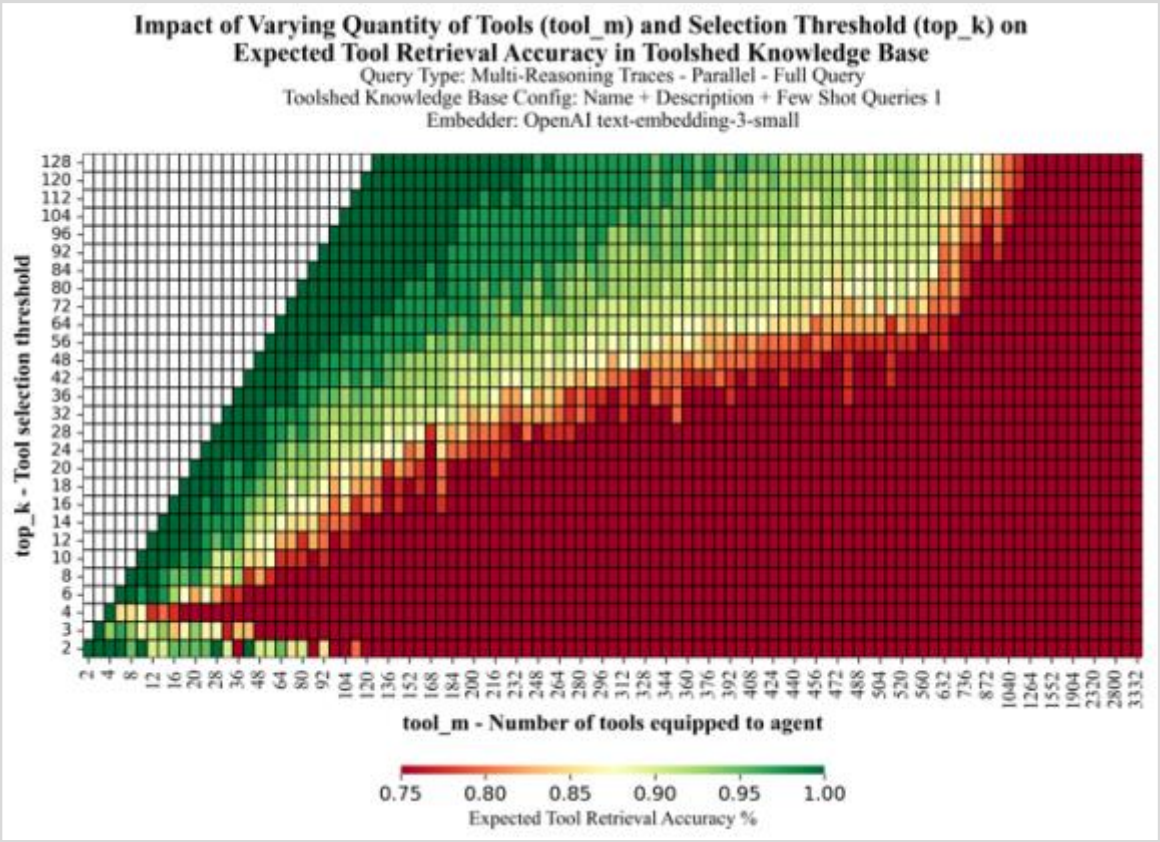

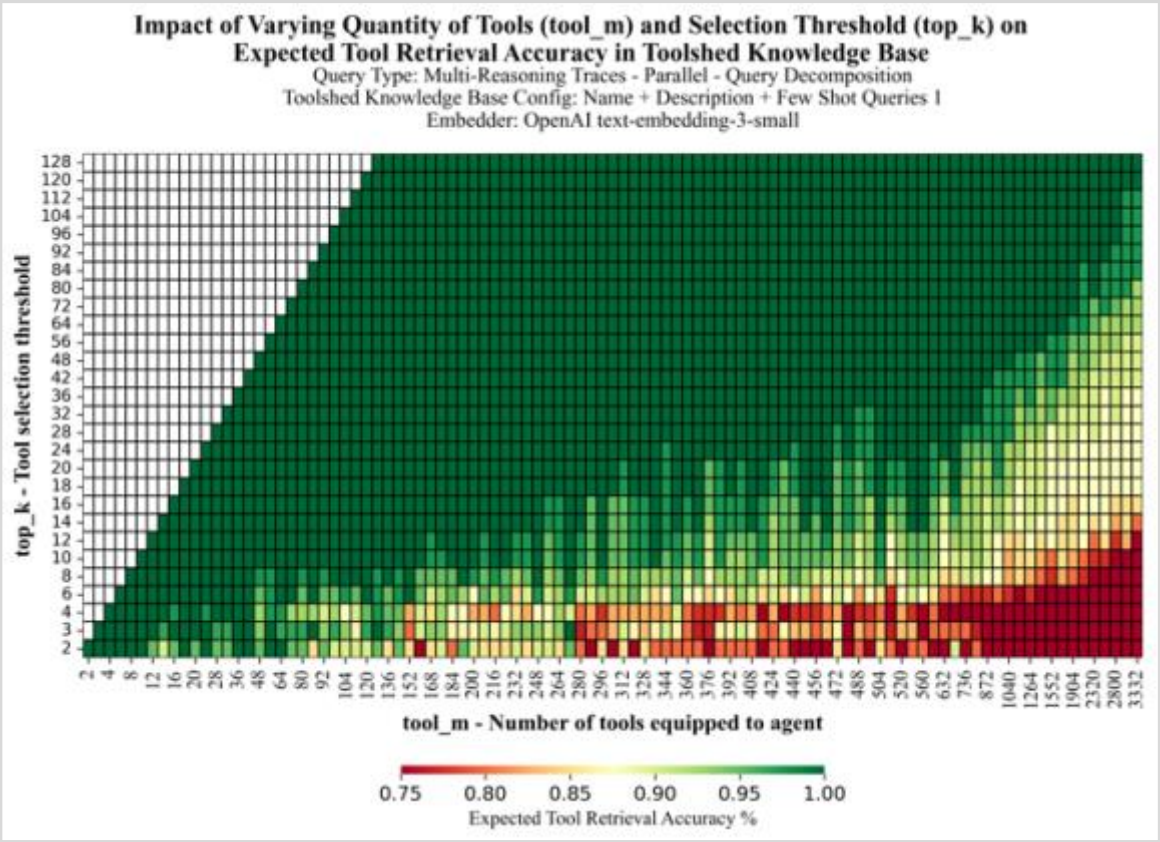

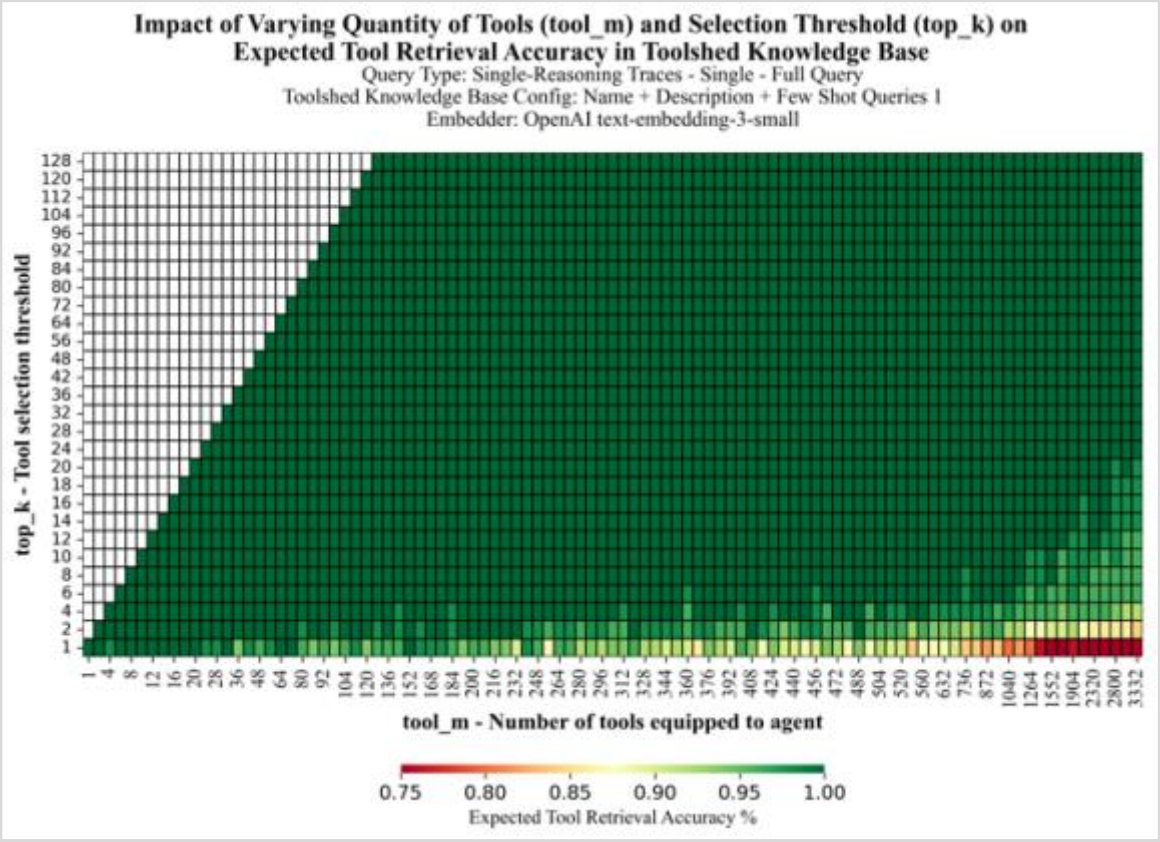

**Configuration 3C:** Embedder: OpenAI text-embedding-ada-002
Toolshed Knowledge Base Configuration: Tool Name, Tool Description, 1 Hypothetical Query

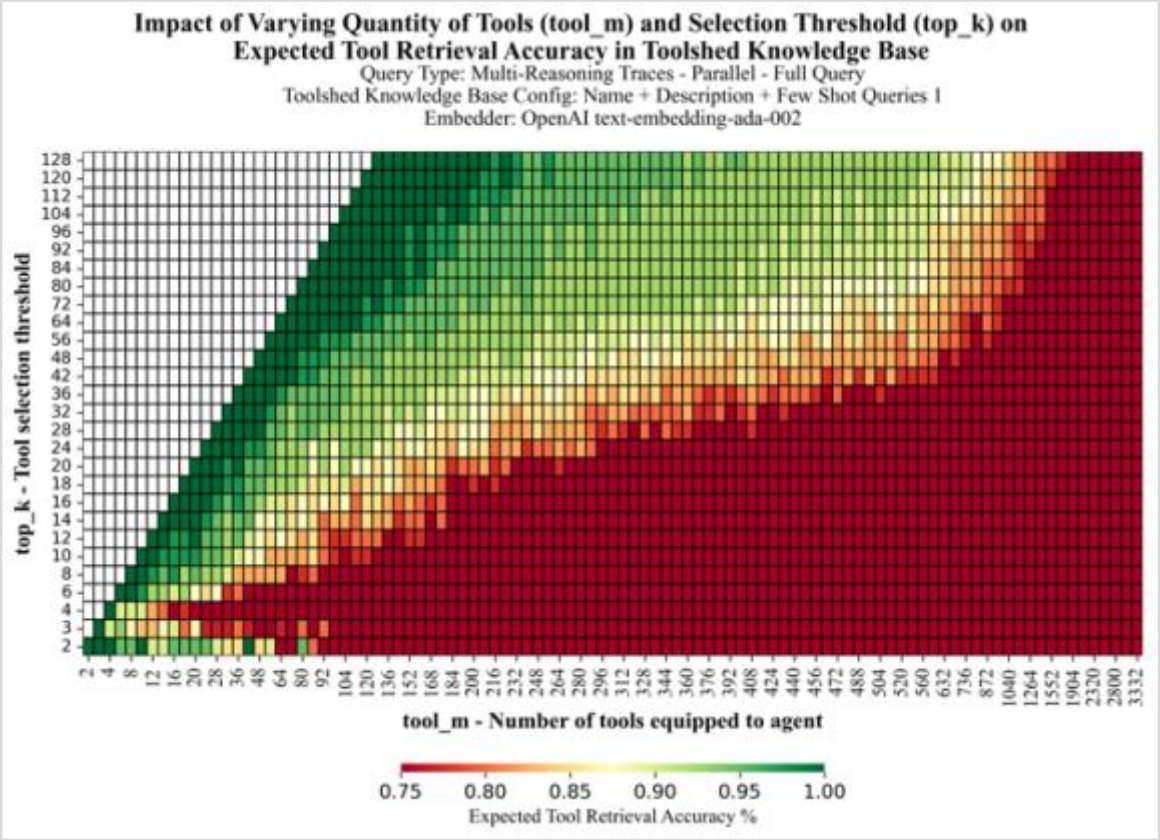

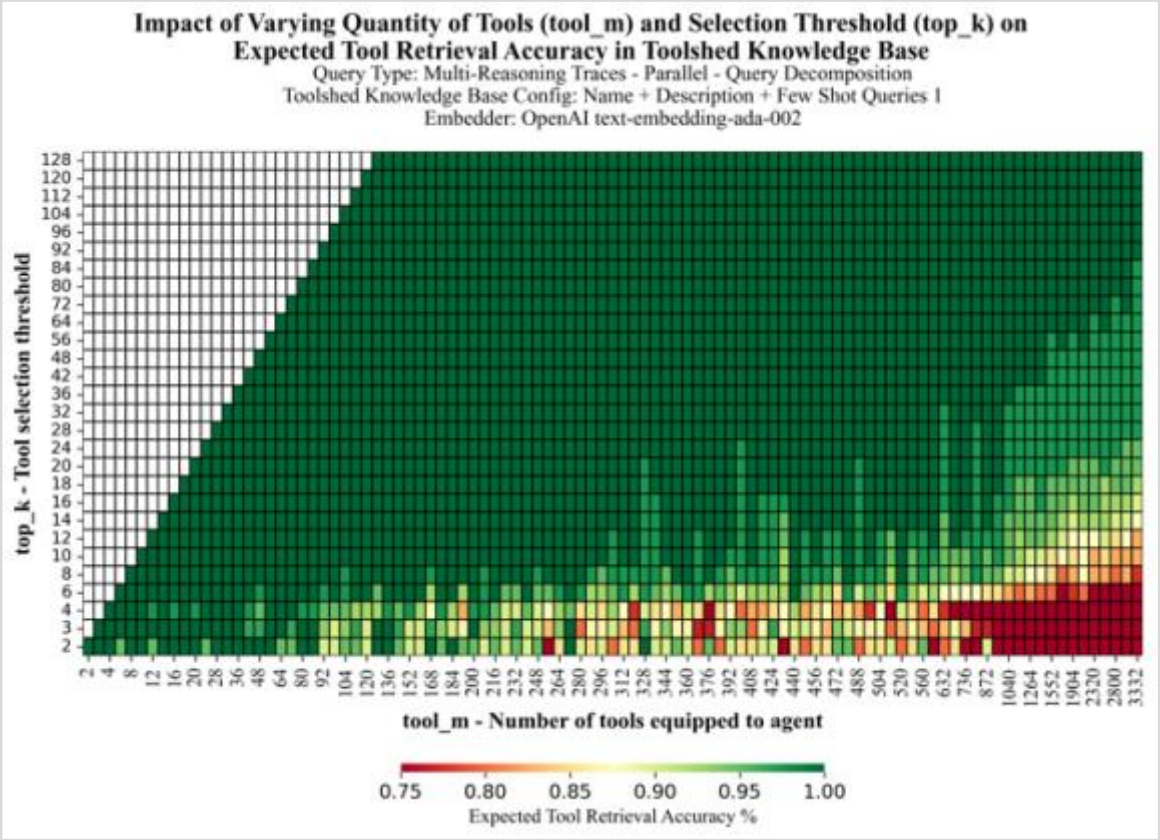

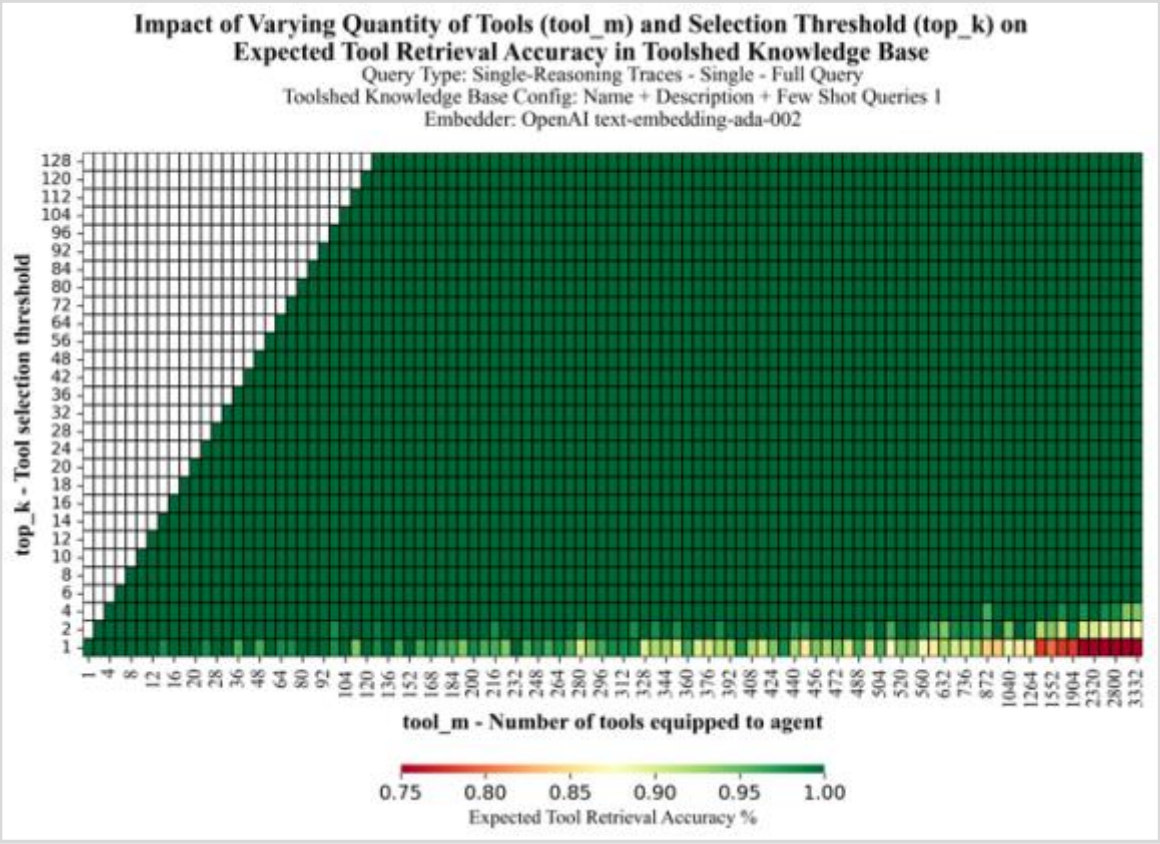

**Configuration 4A:** Embedder: OpenAI text-embedding-3-large
Toolshed Knowledge Base Configuration: Tool Name, Tool Description, Argument Schema, 1 Hypothetical Query

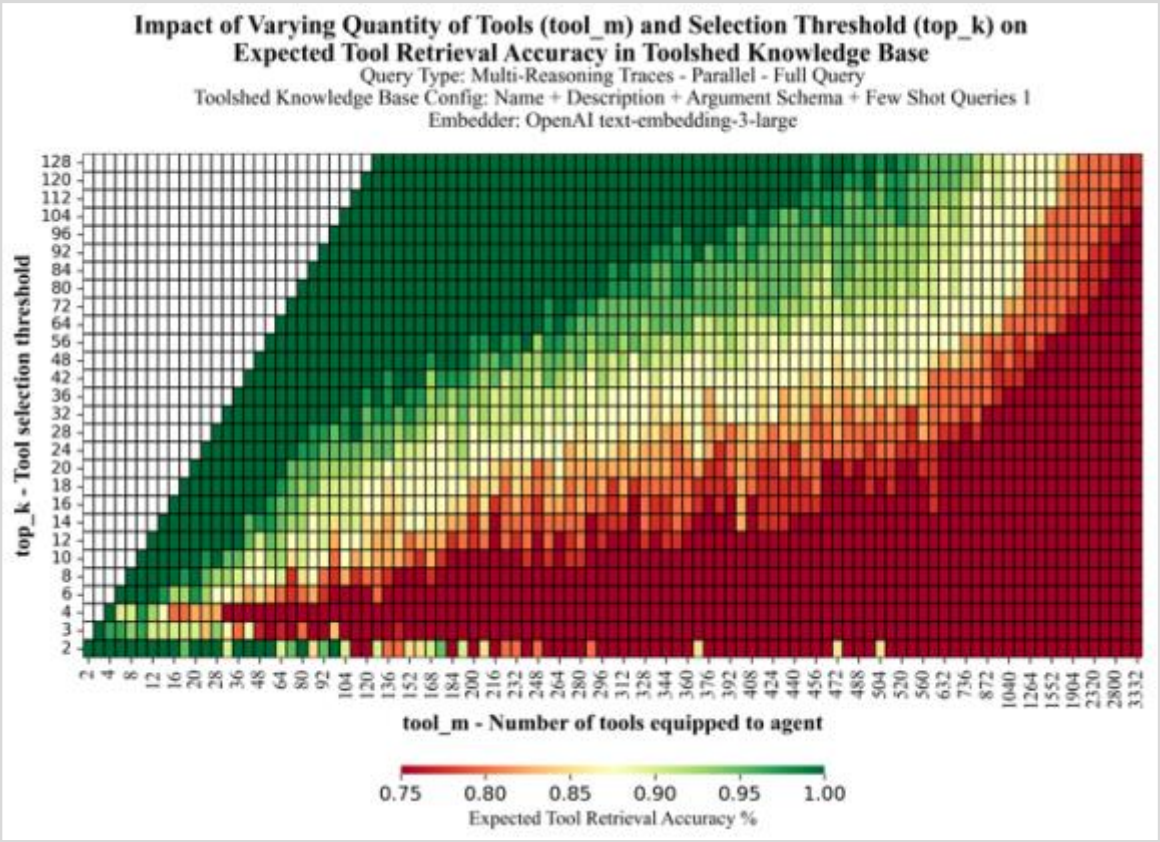

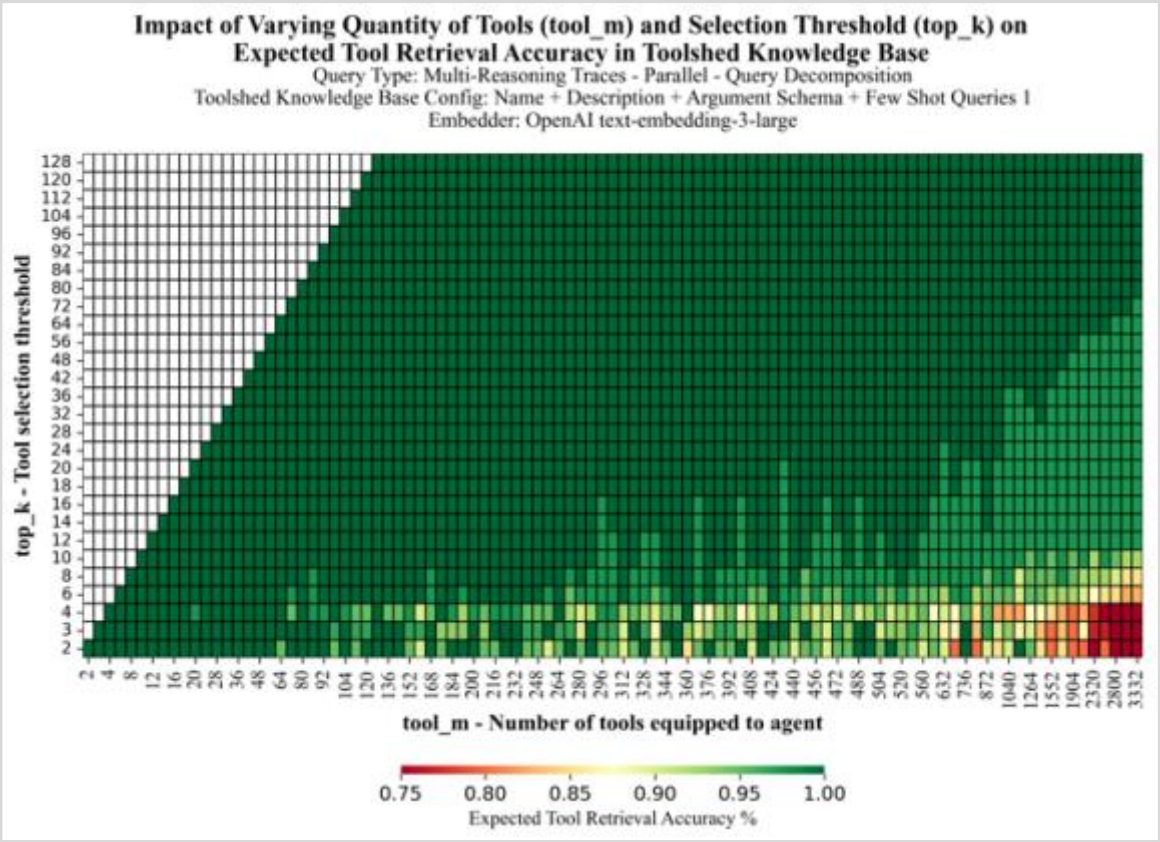

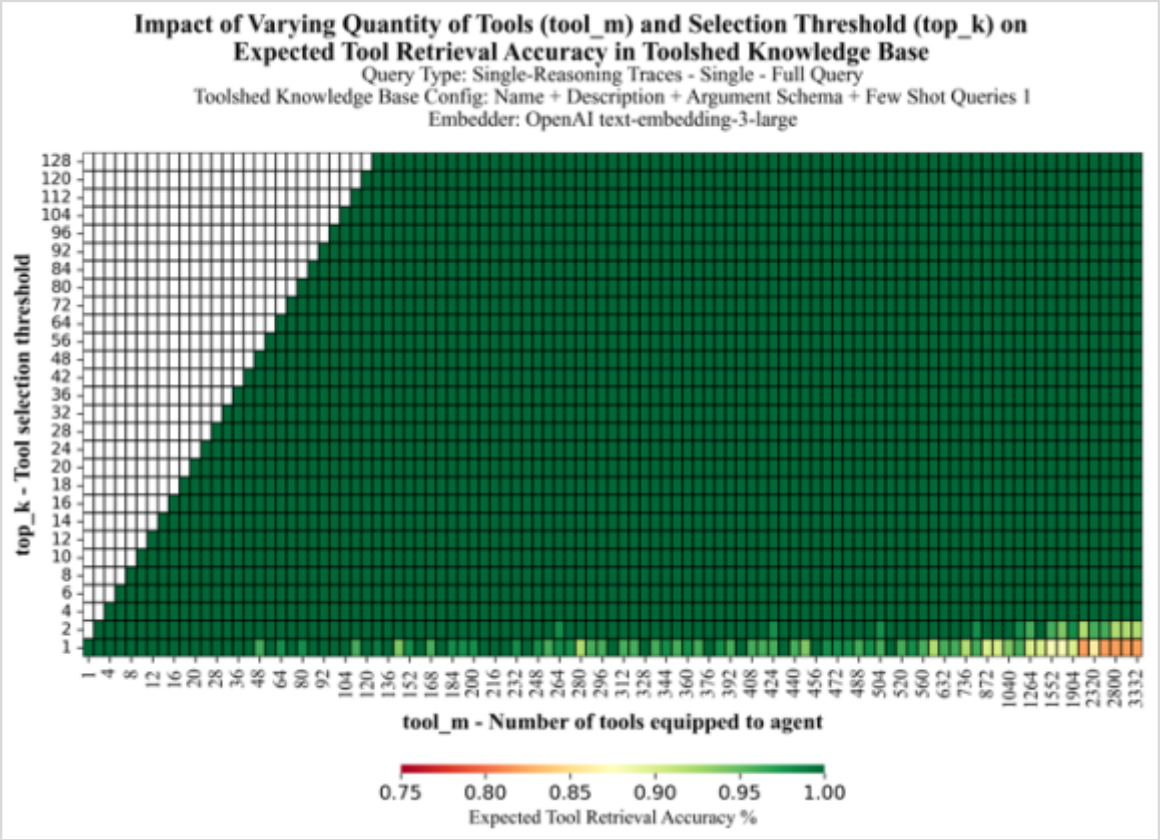

**Configuration 4B:** Embedder: OpenAI text-embedding-3-small
Toolshed Knowledge Base Configuration: Tool Name, Tool Description, Argument Schema, 1 Hypothetical Query

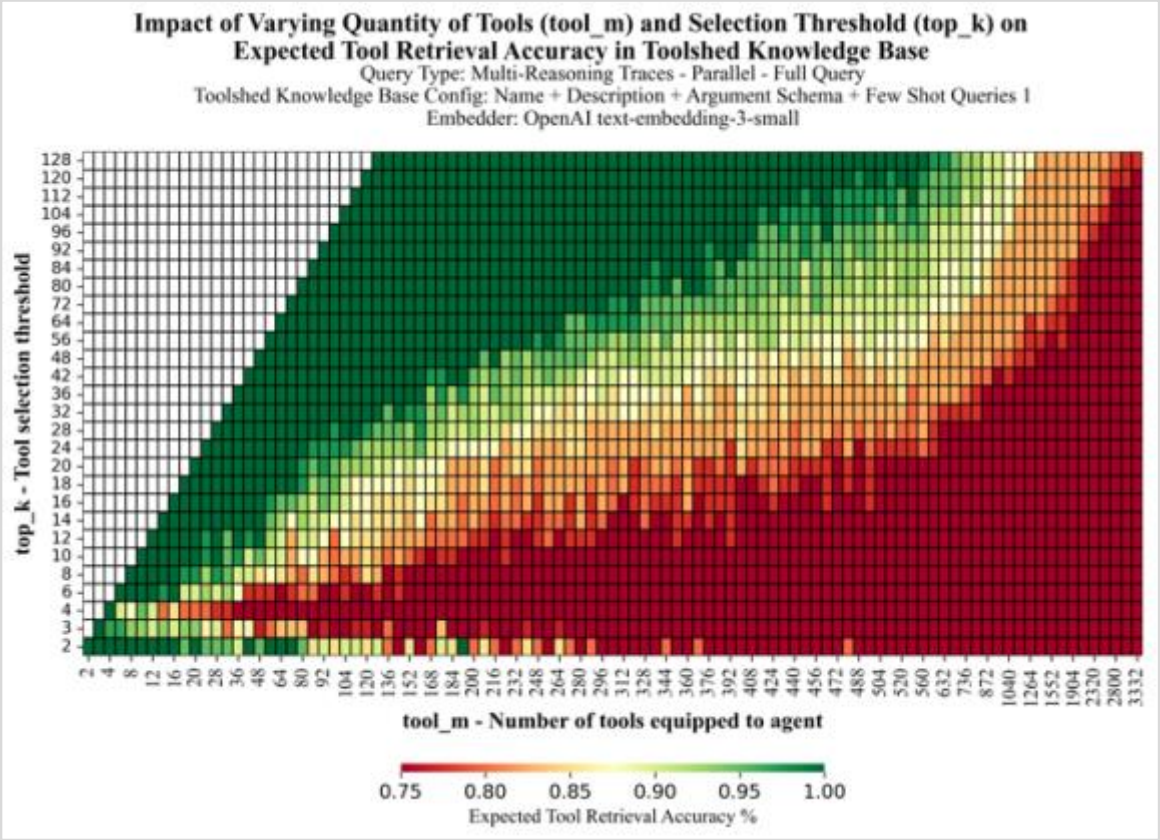

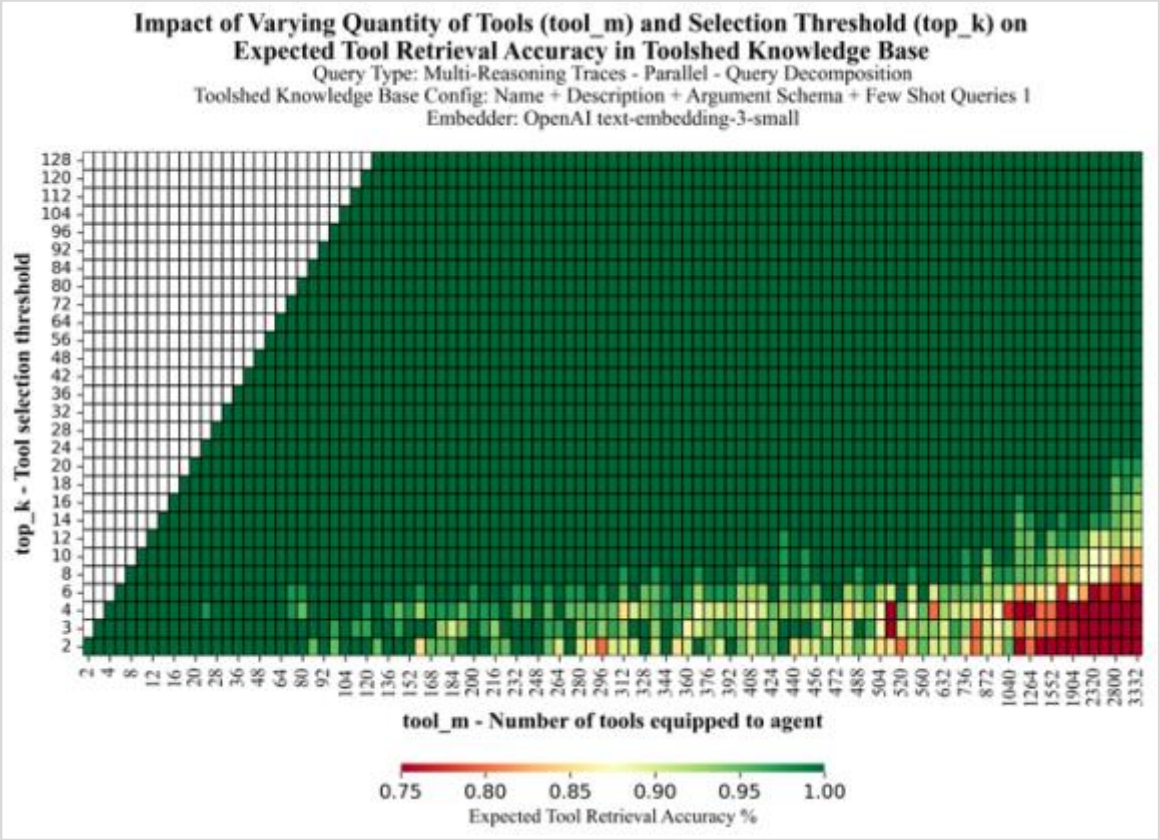

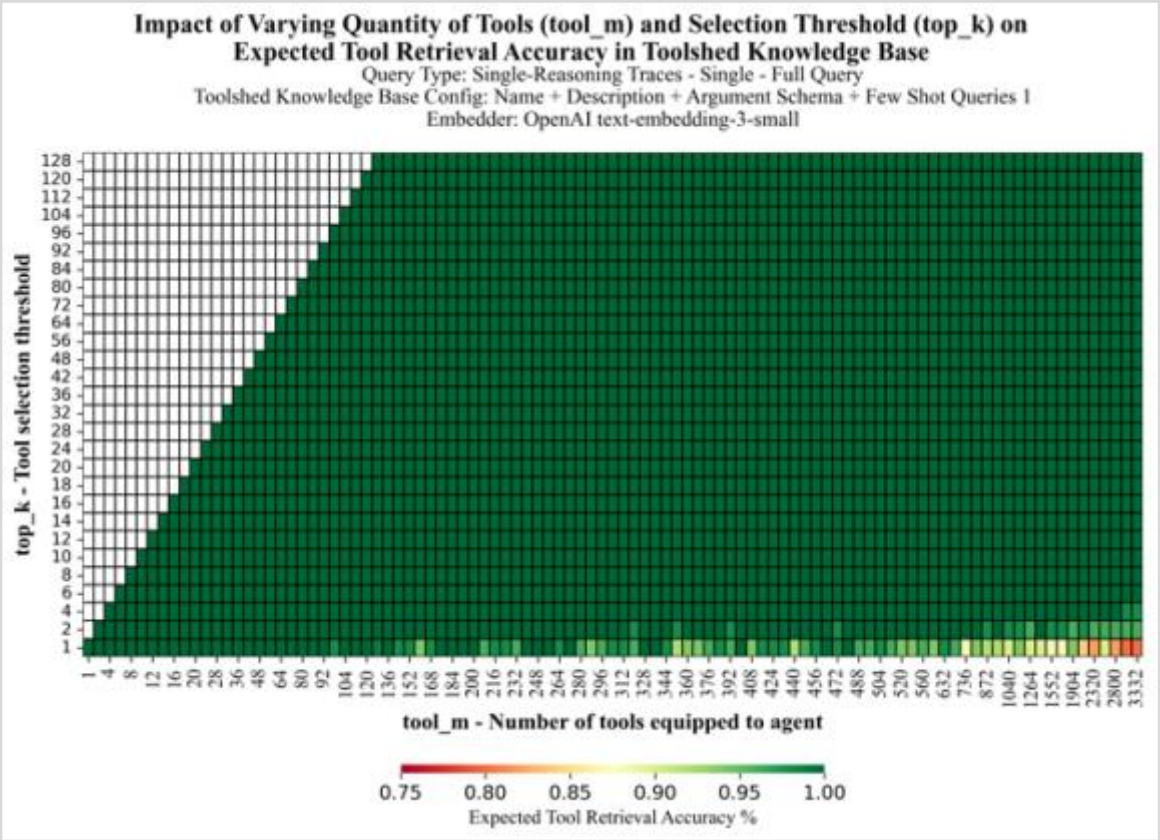

**Configuration 4C:** Embedder: OpenAI text-embedding-ada-002
Toolshed Knowledge Base Configuration: Tool Name, Tool Description, Argument Schema, 1 Hypothetical Query

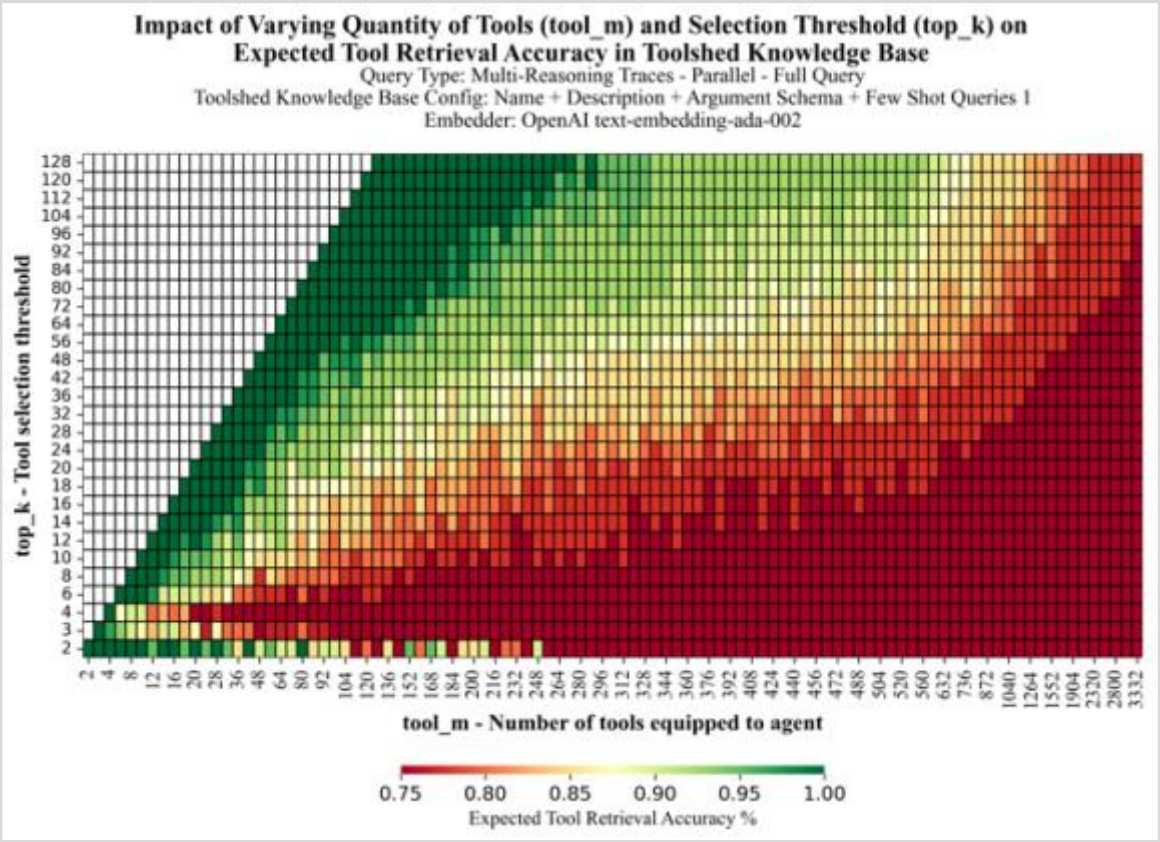

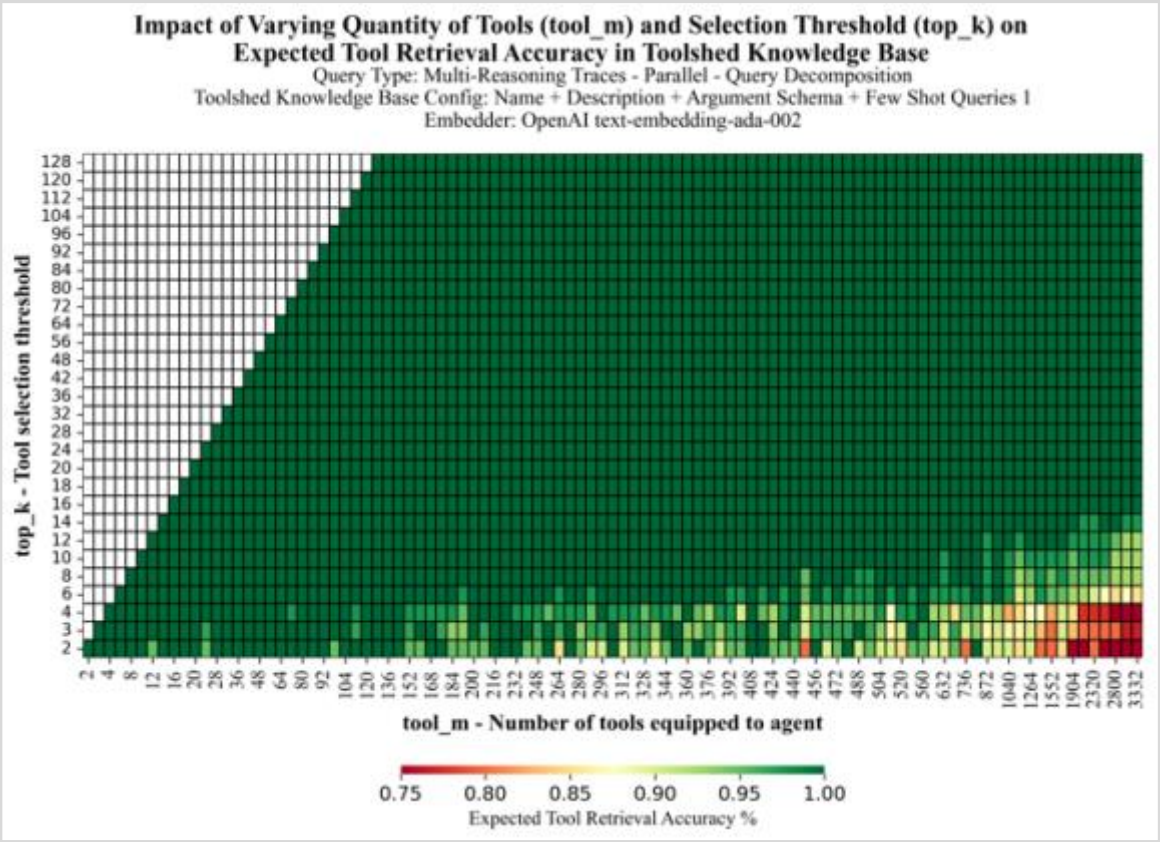

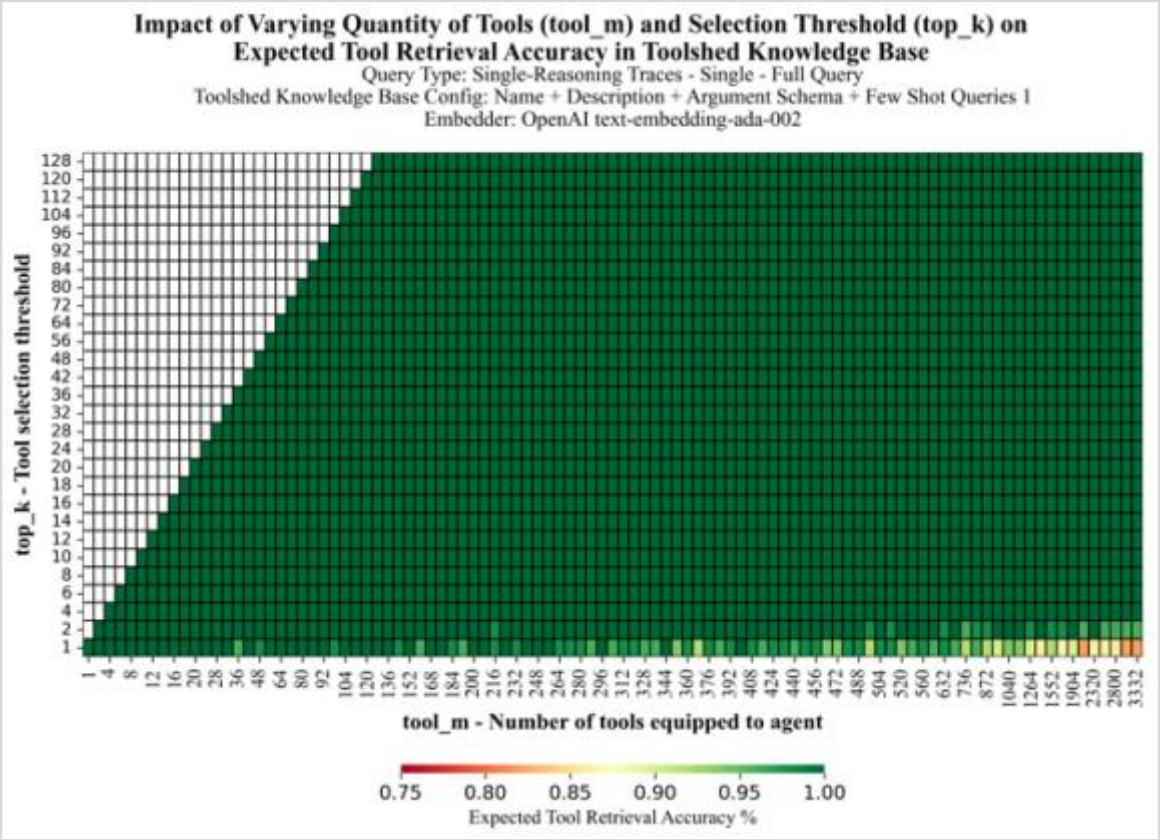

**Configuration 5A:** Embedder: OpenAI text-embedding-3-large
Toolshed Knowledge Base Configuration: Tool Name, Tool Description, 2 Hypothetical Queries

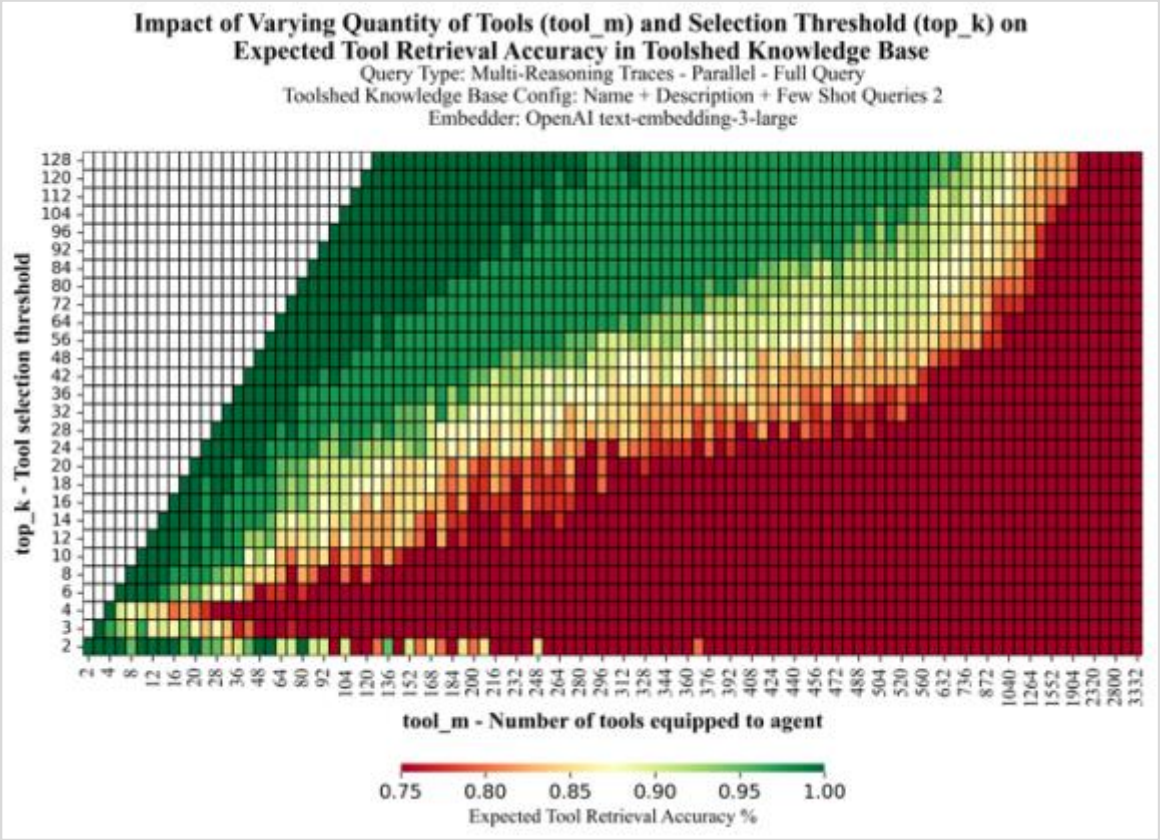

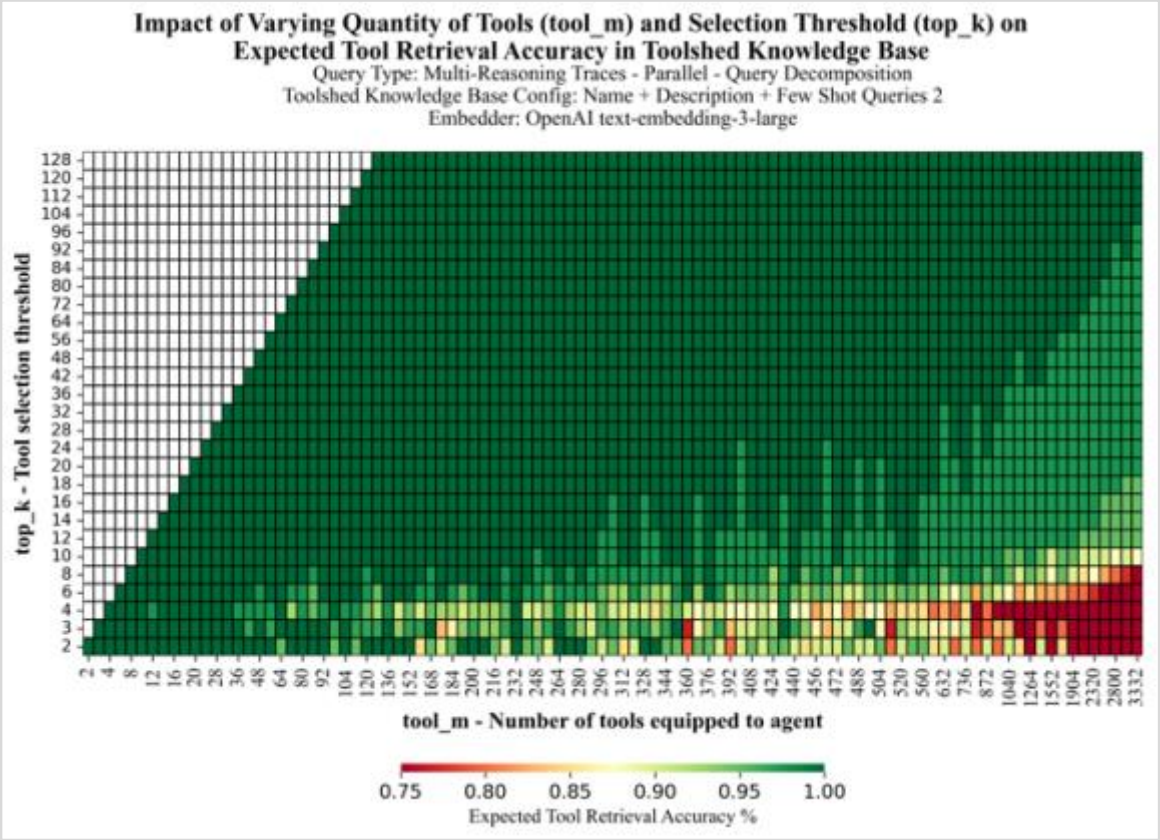

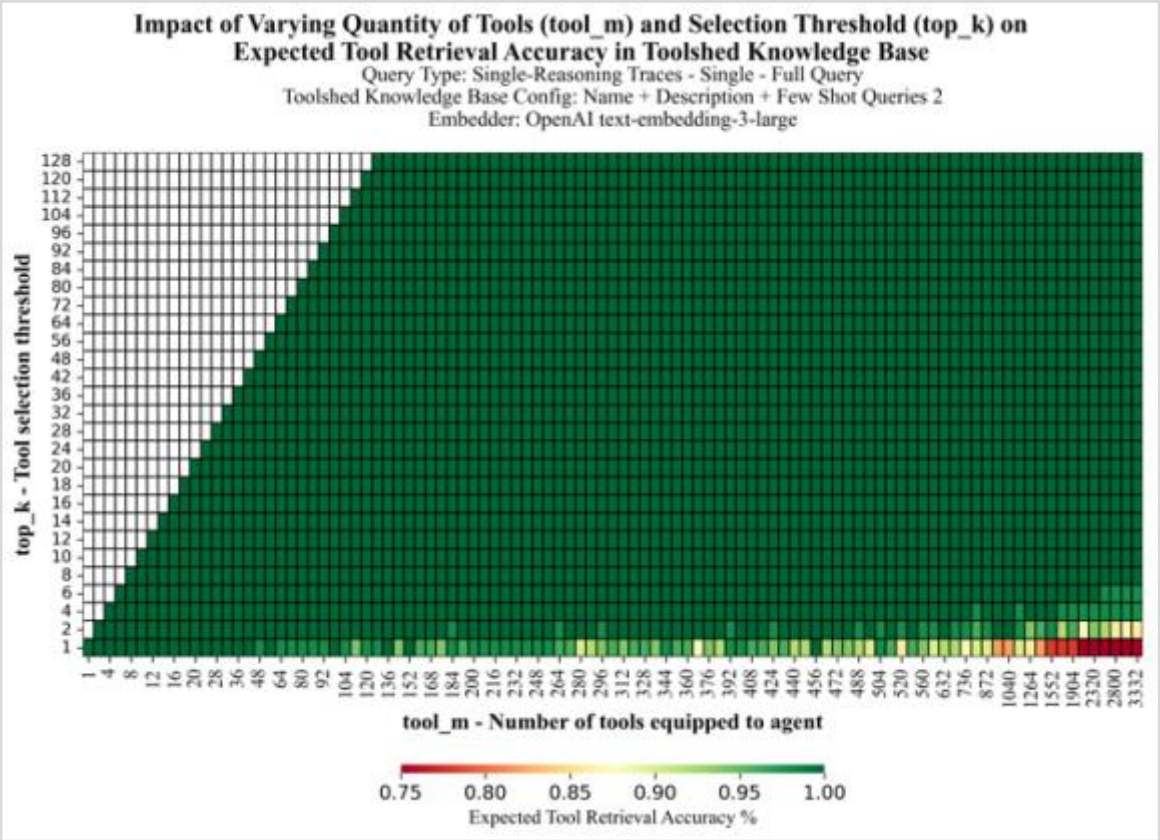

**Configuration 5B:** Embedder: OpenAI text-embedding-3-small
Toolshed Knowledge Base Configuration: Tool Name, Tool Description, 2 Hypothetical Queries

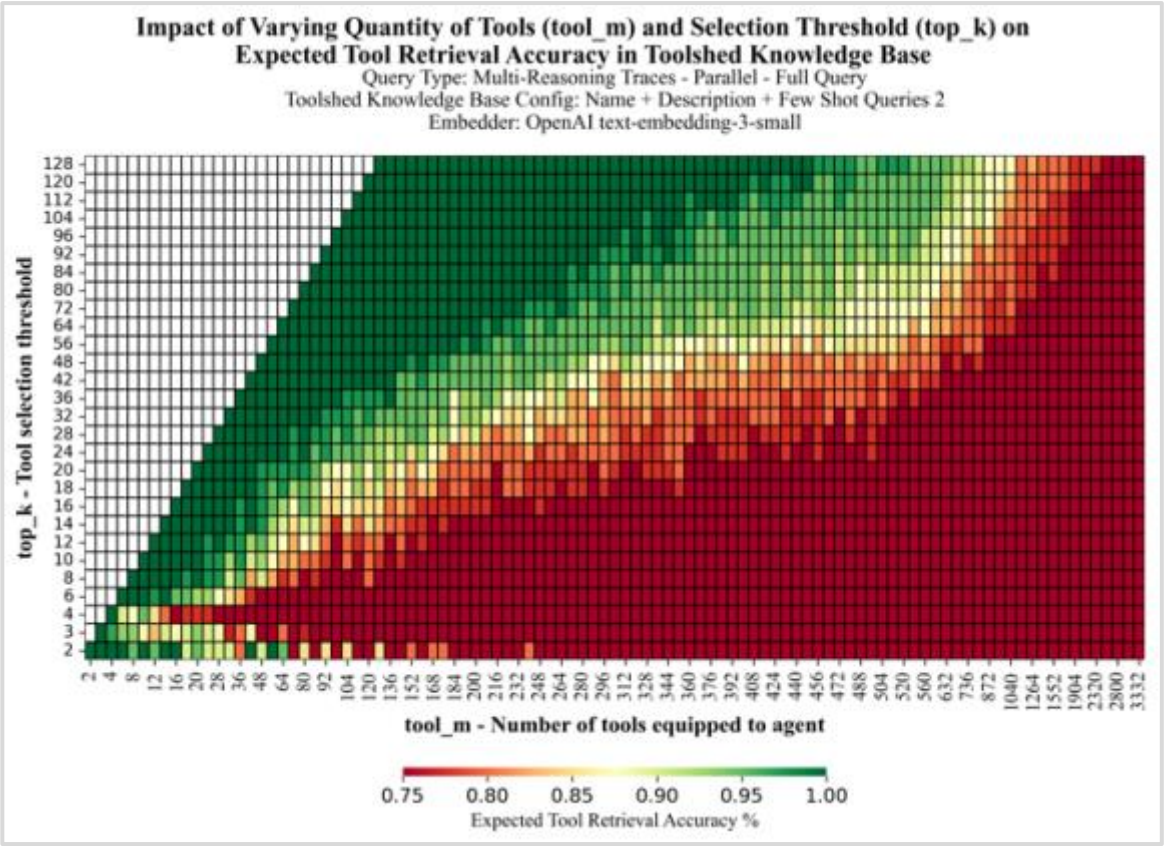

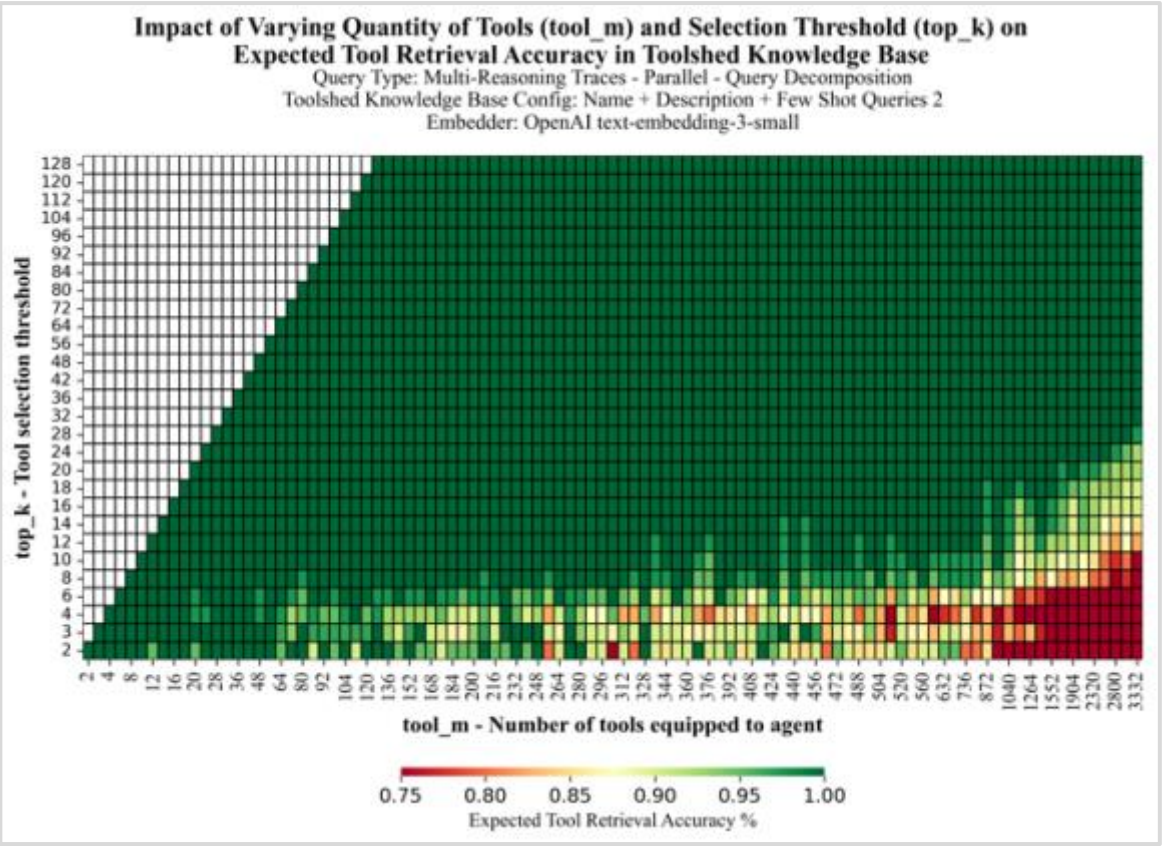

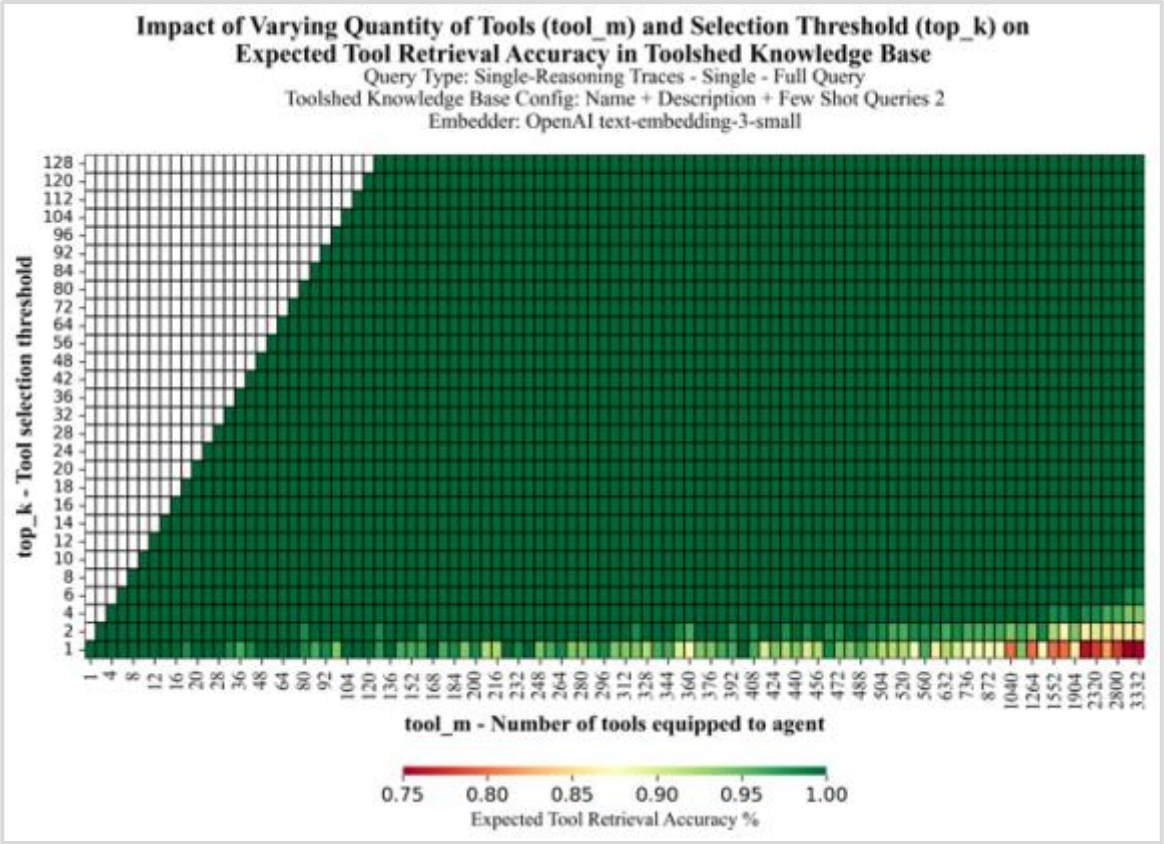

**Configuration 5C:** Embedder: OpenAI text-embedding-ada-002
Toolshed Knowledge Base Configuration: Tool Name, Tool Description, 2 Hypothetical Queries

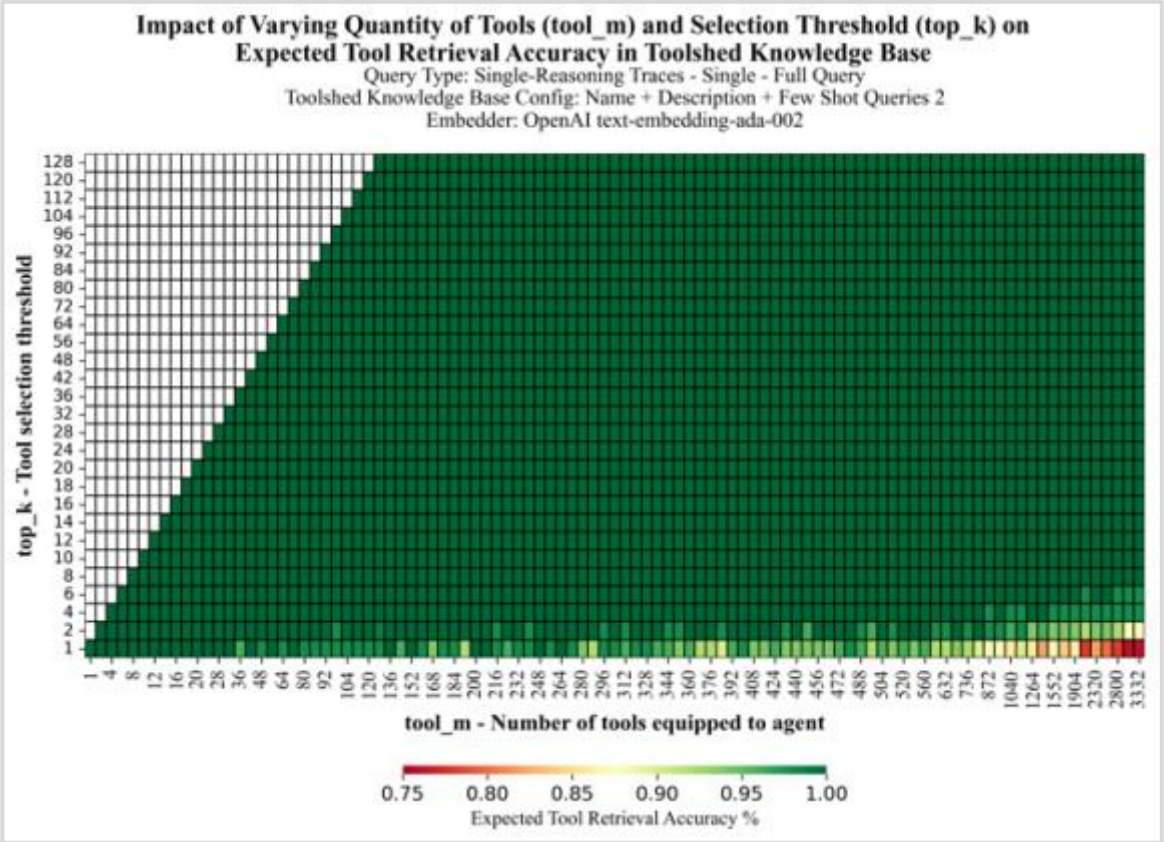

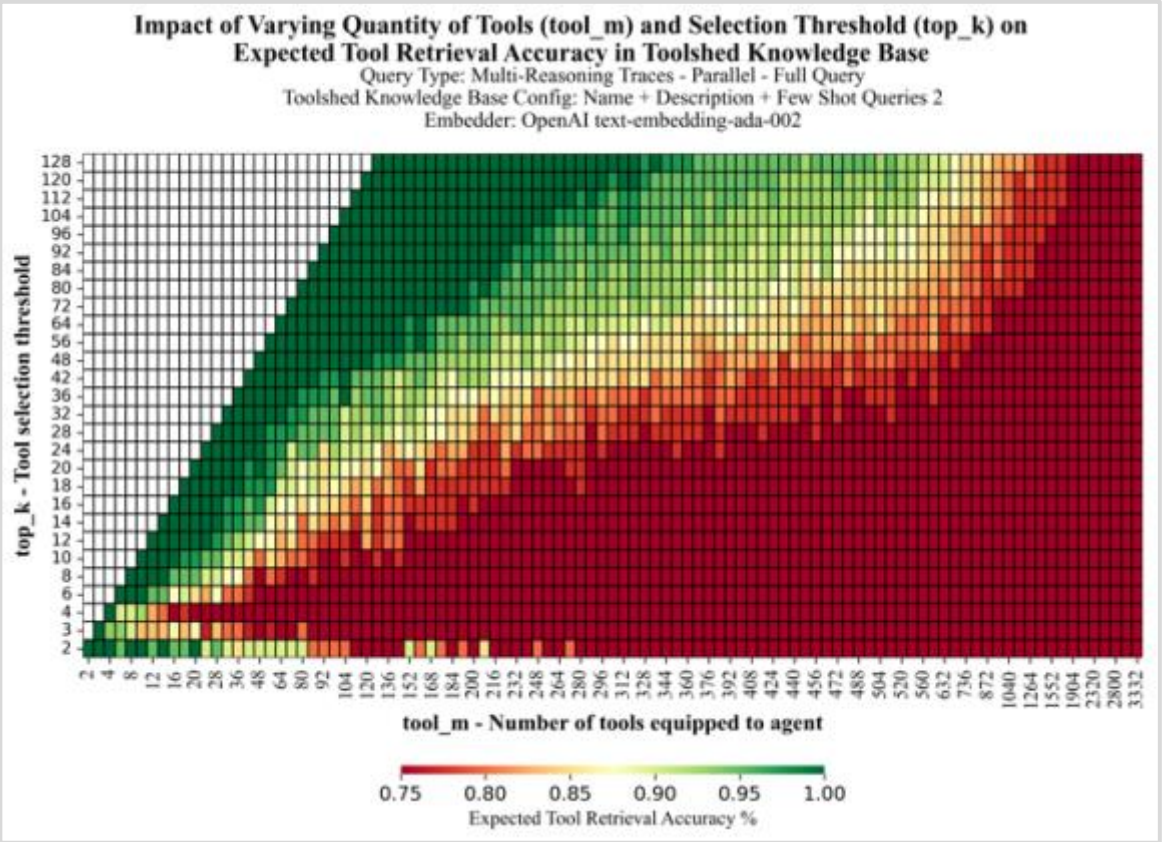

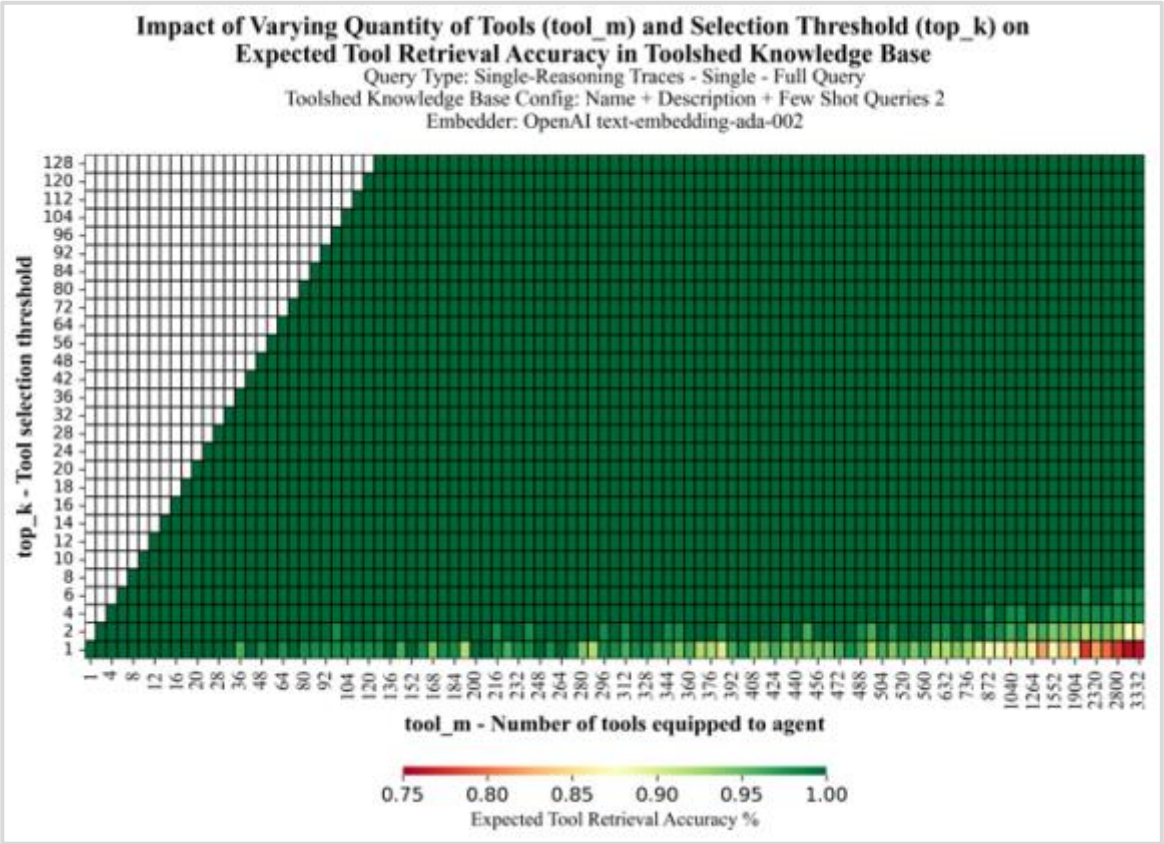

**Configuration 6A:** Embedder: OpenAI text-embedding-3-large
Toolshed Knowledge Base Configuration: Tool Name, Tool Description, Argument Schema, 2 Hypothetical Queries

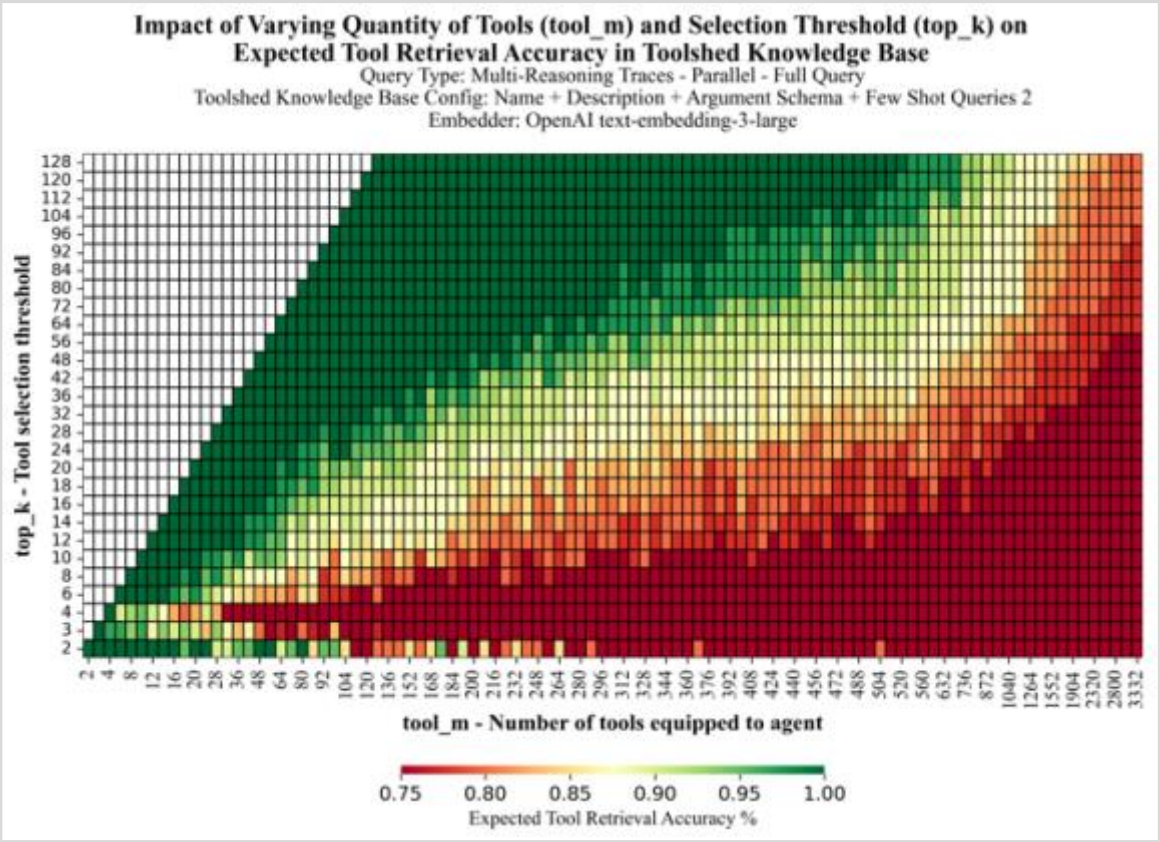

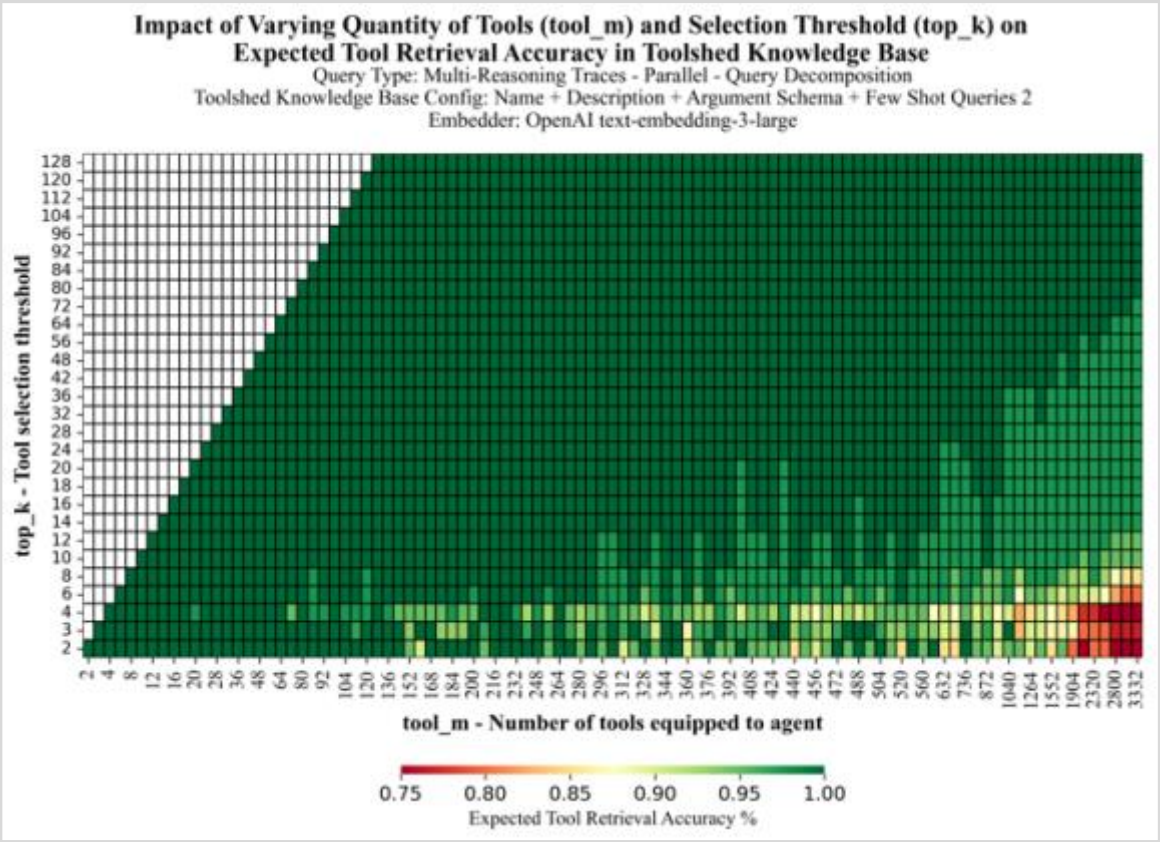

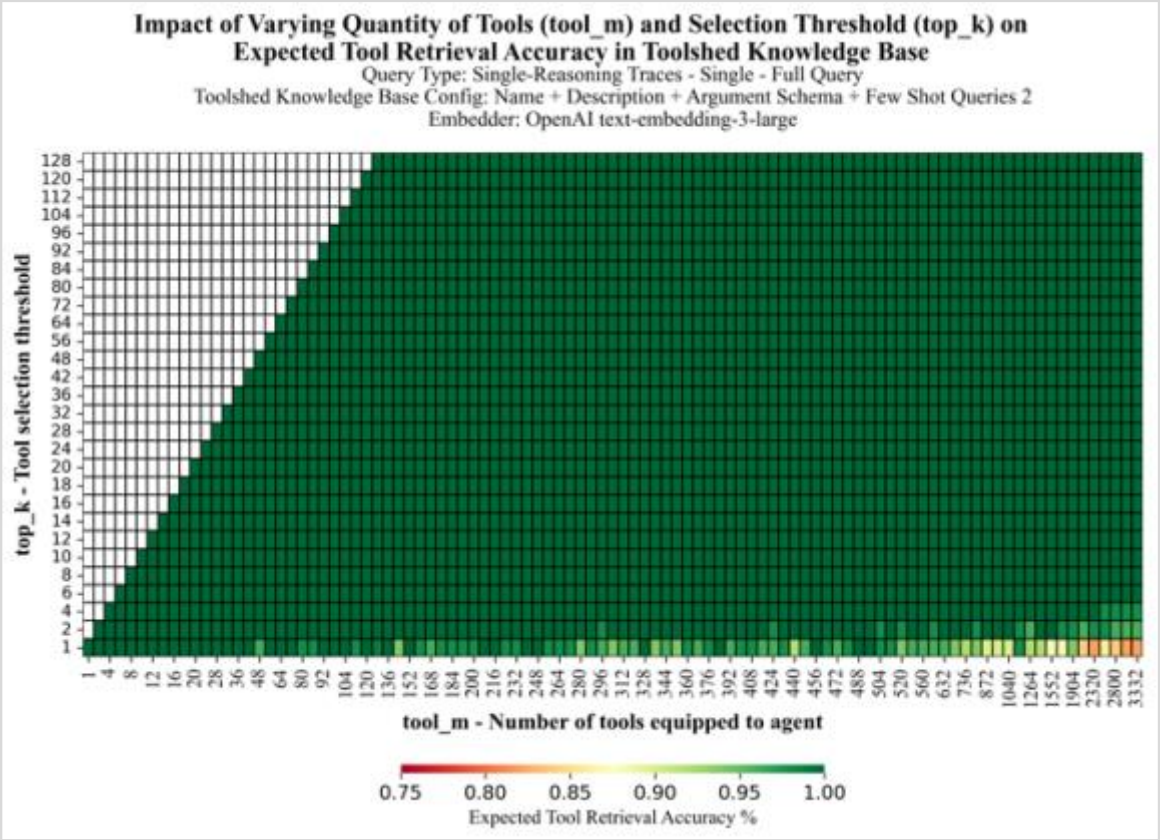

**Configuration 6B:** Embedder: OpenAI text-embedding-3-small
Toolshed Knowledge Base Configuration: Tool Name, Tool Description, Argument Schema, 2 Hypothetical Queries

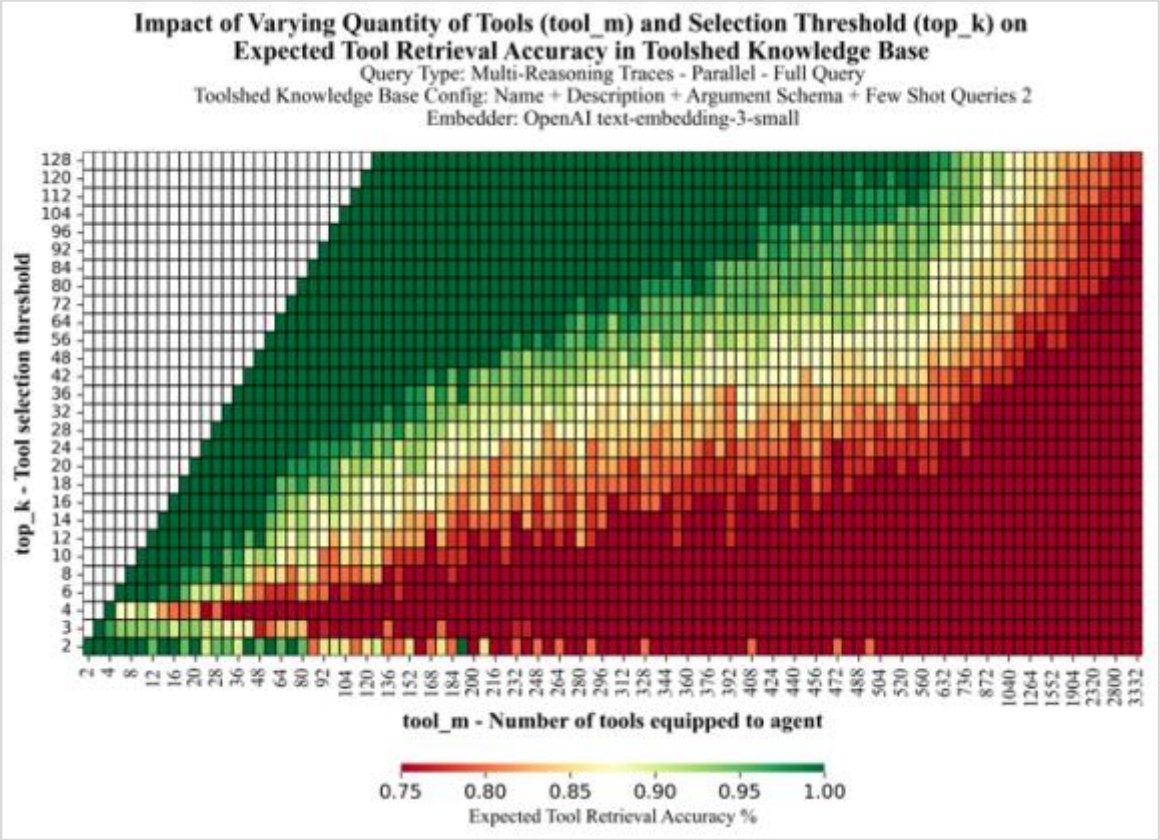

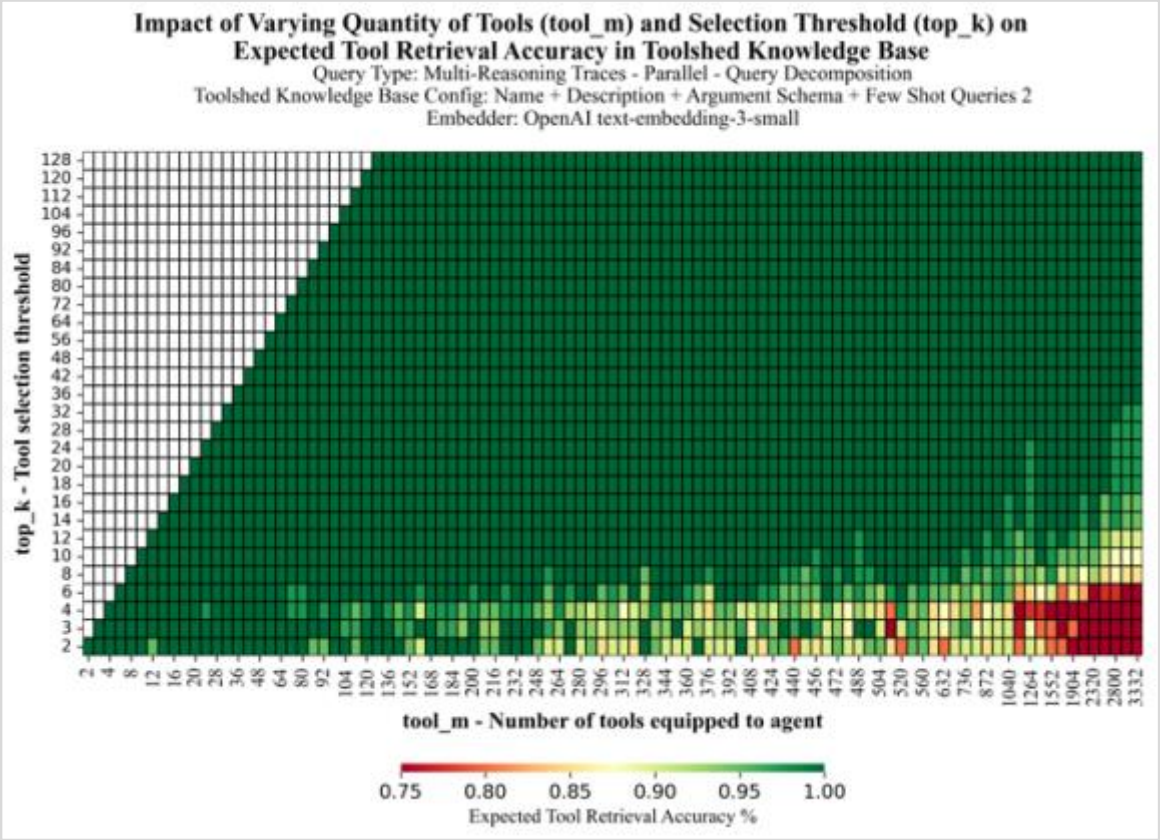

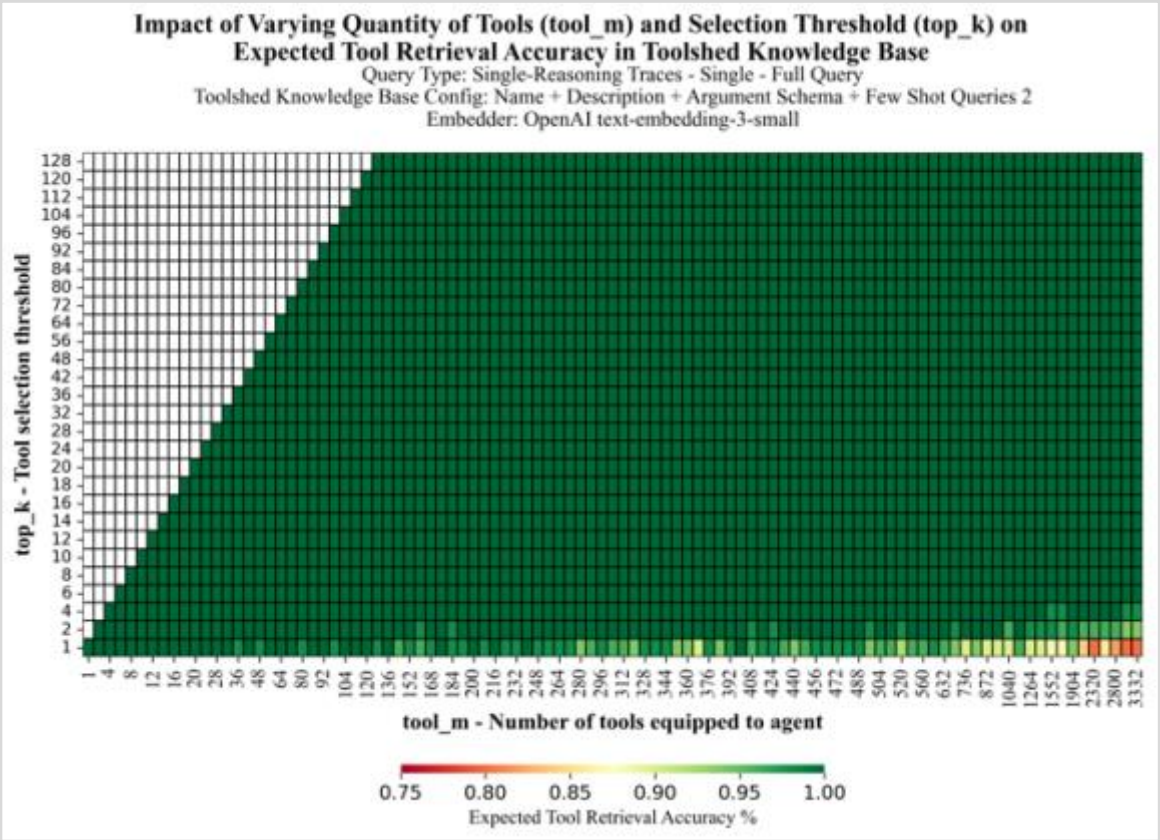

**Configuration 6C:** Embedder: OpenAI text-embedding-ada-002

Toolshed Knowledge Base Configuration: Tool Name, Tool Description, Argument Schema, 2 Hypothetical Queries

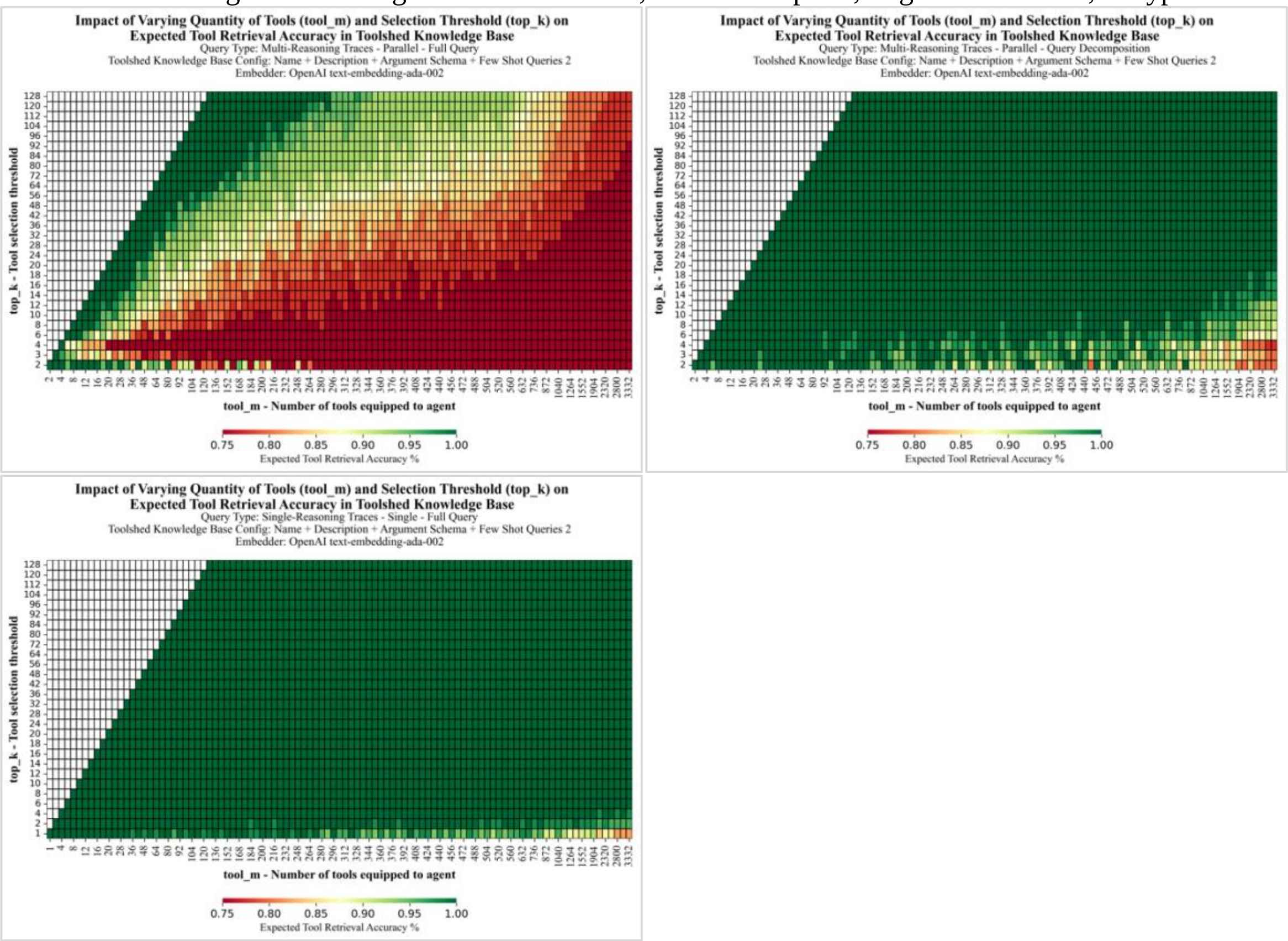

**Figure 25.** Impact of different Azure OpenAI embedders and *Toolshed Knowledge Base* configurations on *Advanced RAG-Tool Fusion* retrieval accuracy, varying the total number of tools (*tool-M*) and tool selection threshold (*top-k*). The configurations compare tool documents stored as: (1) tool name and description, (2) tool name, description, and argument schema, (3) tool name, description, and one few-shot query, (4) tool name, description, argument schema, and one few-shot query, (5) tool name, description, and two few-shot queries, and (6) tool name, description, argument schema, and two few-shot queries. The graphs also assess modular approaches with and without query decomposition, focusing on single reasoning traces and multi-reasoning traces (parallel). Results indicate that argument schema inclusion is critical for high retrieval accuracy, while hypothetical queries showed less impact in this dataset but could be more effective with other datasets, such as ToolE.